\pdfoutput=1
\documentclass[10pt,twocolumn,letterpaper]{article}

\usepackage[pagenumbers]{cvpr} 

\usepackage{graphicx}
\usepackage{amsmath}
\usepackage{amssymb}
\usepackage{booktabs}
\usepackage{CJKutf8}  
\usepackage{array}
\usepackage{tabularx}
\usepackage{multirow}
\usepackage{subcaption}
\usepackage[dvipsnames]{xcolor}
\usepackage{breakcites} 
\usepackage{comment}
\usepackage{url}
\usepackage{colortbl}
\usepackage{nccmath}
\usepackage{makecell}
\usepackage{diagbox}
\usepackage{hhline}
\usepackage{makecell} 

\usepackage{colortbl}
\usepackage{array}

\definecolor{lightgreen}{rgb}{0.88, 1.0, 0.88}
\definecolor{lightpink}{rgb}{1.0, 0.88, 0.88}
\definecolor{lightyellow}{rgb}{1.0, 1.0, 0.88}

\usepackage[dvipsnames]{xcolor}

\definecolor{cvprblue}{rgb}{0.21,0.49,0.74}
\usepackage[pagebackref,breaklinks,colorlinks,citecolor=cvprblue]{hyperref}

\usepackage[capitalize]{cleveref}
\crefname{section}{Sec.}{Secs.}
\Crefname{section}{Section}{Sections}
\Crefname{table}{Table}{Tables}

\newcolumntype{Y}{>{\centering\arraybackslash}X}

\begin{document}
\begin{CJK}{UTF8}{}
\CJKfamily{mj}

\title{
Spectral and Polarization Vision: \\Spectro-polarimetric Real-world Dataset
}

\author{Yujin Jeon$^{1,*}$~ ~ ~
Eunsue Choi$^{1,*}$ ~ ~ ~
Youngchan Kim$^{1}$ ~ ~ ~
Yunseong Moon$^{1}$ ~ ~ ~ \\
Khalid Omer$^{2}$ ~ ~ ~
Felix Heide$^{3}$ ~ ~ ~
Seung-Hwan Baek$^{1}$ \\[2mm]
$^1$ POSTECH   ~ ~ ~   $^2$ Meta   ~ ~ ~  $^3$ Princeton University\\
}
\maketitle
\def\thefootnote{*}\footnotetext{Equal contribution}\def\thefootnote{\arabic{footnote}}



\begin{abstract}

Image datasets are essential not only in validating existing methods in computer vision but also in developing new methods. 
Most existing image datasets focus on trichromatic intensity images to mimic human vision.
However, polarization and spectrum, the wave properties of light that animals in harsh environments and with limited brain capacity often rely on, remain underrepresented in existing datasets. Although spectro-polarimetric datasets exist, these datasets have insufficient object diversity, limited illumination conditions, linear-only polarization data, and inadequate image count. 
Here, we introduce two spectro-polarimetric datasets: trichromatic Stokes images and hyperspectral Stokes images. These novel datasets encompass both linear and circular polarization; they introduce multiple spectral channels; and they feature a broad selection of real-world scenes. With our dataset in hand, we analyze the spectro-polarimetric image statistics, develop efficient representations of such high-dimensional data, and evaluate spectral dependency of shape-from-polarization methods. As such, the proposed dataset promises a foundation for data-driven spectro-polarimetric imaging and vision research. Dataset and code will be publicly available.
\end{abstract}
\section{Introduction}
\label{sec:intro}
Recent progress in computer vision can be largely attributed to comprehensive studies of real-world image datasets, such as ImageNet~\cite{deng2009imagenet}. Foundation models~\cite{ramesh2021zeroshot, alayrac2022flamingo, yuan2021florence, kirillov2023segany} further underscore data significance. Most of these datasets comprise trichromatic intensity images, inspired by human visual perception, enabling machines to emulate human vision with trichromatic RGB cameras. As such, the datasets have facilitated the development of low-cost, camera-based autonomous agents capable of perceiving and interacting with our world, as we do.
However, the reliance on trichromatic intensity in existing image datasets also comes with inherent limitations for analyzing objects in depth. Examples include textureless surface, low-albedo objects, and transparent materials. 

Light possesses wave properties, including {polarization and spectrum~\cite{collett2005field}}, which are not faithfully captured by trichromatic intensity imaging. While these properties are invisible to human, animals like honeybees and ants leverage the polarization and spectrum for navigation and other tasks. Horvath and Varju~\cite{horvath2004polarized} provide diverse examples and mechanisms of spectral and polarimetric vision in animals. Partly drawing inspiration from nature, researchers have extended the analysis of spectrum and polarization to a variety of fields, including computer vision, robotics, and astronomy. This has spurred interest in polarimetric~\cite{baek2018simultaneous, baltaxe2023polarimetric, mu2019optimized, kurita2023sensor} and hyperspectral imaging~\cite{imamoglu2018hyperspectral, aloupogianni2022hyperspectral,cui2021snapshot}, and more recently, their integration into spectro-polarimetric imaging~\cite{huber1997spectro, zhao2009spectropolarimetric, shibata2019robust, manakov2013reconfigurable, fan2020hyperspectral, fan2022four, altaqui2021mantis, lv2020snapshot, garcia2017bio, mu2017snapshot}. Prior work using spectro-polarimetric images has shown potential for skin analysis~\cite{zhao2009spectropolarimetric}, vegetation classification~\cite{yao2022combination}, shape reconstruction~\cite{huynh2013shape}, object recognition~\cite{denes1998spectropolarimetric}, and segmentation~\cite{islam2020hybrid, tu2021synergetic}.

There are existing spectro-polarimetric datasets, summarized in Figure~\ref{fig:dataset}, that have been invaluable for these analysis~\cite{lapray2018data, qiu2021demos, kurita2023sensor, fan2023fullst} and training neural networks~\cite{lei2020refl, lyu2019refl, ba2019sfp, liang2022material, mei2022glass, dave2022pandora, gao2022ppp, kondo2020accurate, lei2022wild}. However, these datasets unfortunately do not capture the diversity of real-world spectro-polarimetric images as effectively as their trichromatic intensity counterparts do. They typically suffer from limited object, scene, and illumination diversity, contain primarily linear polarization information, and offer a small number of images. To advance the field, we propose a comprehensive spectro-polarimetric dataset that encompasses: (1) \emph{Full Stokes polarimetric data}, including both linear and circular polarization states, represented by Stokes vectors~\cite{collett2005field} for each pixel and wavelength. (2) \emph{A diverse range of spectral channels}, facilitating in-depth exploration of the interplay between spectrum and polarization.
(3) \emph{A broad array of real-world scenes}, crucial for extracting meaningful statistics and relationships within spectro-polarimetric images.
\noindent

To this end, we introduce two novel spectro-polarimetric datasets designed to cover real-world spectro-polarimetric scenes: a trichromatic Stokes dataset consisting of 2,022 images, and a hyperspectral Stokes dataset containing 311 images across 21 spectral channels. 
The trichromatic Stokes dataset covers a wider range of scenes thanks to its convenient capture setup and process.
The hyperspectral Stokes dataset provides richer spectral-polarimetric information than the trichromatic Stokes dataset.
Both datasets cover a variety of natural indoor and outdoor scenes. Each image in these datasets is annotated with four specific parameters: the type of environment (indoor or outdoor), the illumination conditions (clear/cloudy sunlight or white/yellow office light), the timestamp of capture, and the scene categorization (either object- or scene-oriented).

Utilizing these datasets, we systematically analyze the statistics of real-world spectro-polarimetric images. We focus on examining statistics of Stokes vectors, in addition to the gradients and polarimetric attributes associated with them. We also conduct an analysis of unpolarized and polarized images derived through polarimetric decomposition.
We then develop two efficient spatio-spectral-polarimetric representations using principal component analysis (PCA) and implicit neural representation (INR). These representations exhibit effective denoising capabilities and low memory footprints by exploiting the compressible structure of spectro-polarimetric images. 
We also analyze the impact of intensity denoising for spectro-polarimetric images, spectral dependency of shape-from-polarization methods, and environment dependency on the statistics of spectro-polarimetric images.

In summary, we make the following contributions.
\begin{itemize}
    \item 
    {We introduce a trichromatic Stokes dataset and a hyperspectral Stokes dataset, featuring 2,333 diverse annotated indoor and outdoor scenes under various illumination conditions, which encompasses full-Stokes  polarization data for linear and circular states.}
    \item We develop efficient spatio-spectral-polarimetric representations and analyze real-world spectro-polarimetric images, encompassing Stokes vectors and their gradients, unpolarized and polarized images, shape from polarization, denoising, and environment dependency.
\end{itemize}

\section{Related Work}
\label{sec:relwork}

\begin{figure*}[t]
    \begin{minipage}{0.48\linewidth}
        \centering
        \includegraphics[width=\textwidth]{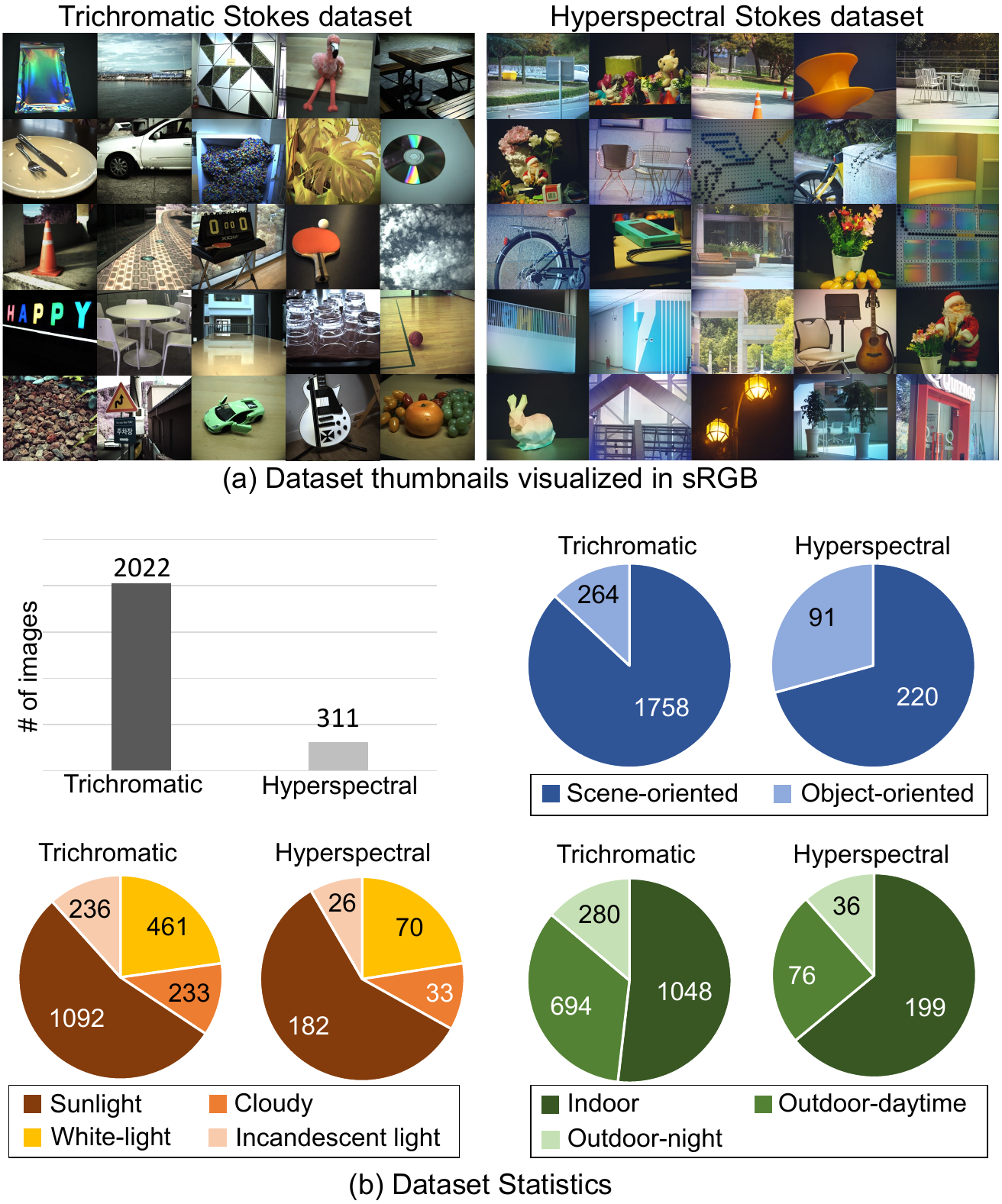}
    \end{minipage}
    \hfill
    \begin{minipage}{0.52\linewidth}
        \centering
        \setlength{\tabcolsep}{2pt}
        \begin{tabular}{|p{1.1cm}|p{1.1cm}|p{1.0cm}|p{1.1cm}|p{3.65cm}|}
            \hline
            Dataset & Polar. & \#\mbox{ of} bands & Scene count & Scene diversity \\ 
            \Xhline{2\arrayrulewidth}\Xhline{0.2pt}
\cite{lei2022wild} & \cellcolor{lightpink}LP & \cellcolor{lightpink}1 & \cellcolor{lightyellow}522 & \cellcolor{lightyellow}Outdoor scenes \\ 
\cite{ba2020deep} & \cellcolor{lightpink}LP & \cellcolor{lightpink}1 & \cellcolor{lightyellow}300 & \cellcolor{lightyellow}Indoor objs. \\ 
\cite{lapray2018data} & \cellcolor{lightpink}LP & \cellcolor{lightyellow}6 & \cellcolor{lightpink}10 & \cellcolor{lightyellow}Indoor objs. \\ 
\cite{qiu2021demos} & \cellcolor{lightpink}LP & \cellcolor{lightyellow}3 & \cellcolor{lightpink}40 & \cellcolor{lightyellow}Indoor objs. \\ 
\cite{dave2022pandora} & \cellcolor{lightpink}LP & \cellcolor{lightyellow}3 & \cellcolor{lightpink}3 & \cellcolor{lightyellow}Indoor multiview\\ 
\cite{dave2022pandora} & \cellcolor{lightpink}LP & \cellcolor{lightyellow}3 & \cellcolor{lightpink}2 & \cellcolor{lightyellow}Synthetic multiview\\ 
\cite{gao2022ppp} & \cellcolor{lightpink}LP & \cellcolor{lightyellow}3 & \cellcolor{lightpink}6 & \cellcolor{lightyellow}Indoor multiview\\ 
\cite{lei2020refl} & \cellcolor{lightpink}LP & \cellcolor{lightyellow}3 & \cellcolor{lightyellow}807 &  \cellcolor{lightyellow}Reflective objs.   \\ 
\cite{liang2022material} & \cellcolor{lightpink}LP & \cellcolor{lightyellow}3 & \cellcolor{lightyellow}500 & \cellcolor{lightyellow}Outdoor scenes \\ 
\cite{kondo2020accurate} & \cellcolor{lightpink}LP & \cellcolor{lightyellow}3 & \cellcolor{lightgreen}44,300 & \cellcolor{lightpink}Synthetic \\ 
\cite{lyu2019refl} & \cellcolor{lightpink}LP & \cellcolor{lightyellow}3 & \cellcolor{lightgreen}3,200 & \cellcolor{lightyellow}Reflective objs.   \\ 
\cite{mei2022glass} & \cellcolor{lightpink}LP & \cellcolor{lightyellow}3 & \cellcolor{lightgreen}4,500 &  \cellcolor{lightyellow}Transparent objs.  \\ 
\cite{kurita2023sensor} & \cellcolor{lightpink}LP & \cellcolor{lightyellow}3 & \cellcolor{lightgreen}2,000 & \cellcolor{lightgreen}Indoor/outdoor scenes \\ 
\cite{fan2023fullst} & \cellcolor{lightgreen}LP, CP & \cellcolor{lightgreen}18 & \cellcolor{lightpink}67 & \cellcolor{lightpink}Flat objs. \\ 
\cite{kim2023neural} & \cellcolor{lightgreen}LP, CP & \cellcolor{lightgreen}21 & \cellcolor{lightpink}4 & \cellcolor{lightpink}Synthetic multiview \\ 
\cite{kim2023neural} & \cellcolor{lightgreen}LP, CP & \cellcolor{lightgreen}21 & \cellcolor{lightpink}4 & \cellcolor{lightgreen}Indoor/outdoor multiview\\ 
\hline
\begin{tabular}[c]{@{}l@{}}\textbf{Ours}\\ \textbf{(RGB)}\end{tabular} & \cellcolor{lightgreen}LP, CP & \cellcolor{lightyellow}3 & \cellcolor{lightgreen} 2022  & \cellcolor{lightgreen}Indoor/outdoor scenes \\ 
\begin{tabular}[c]{@{}l@{}}\textbf{Ours}\\ \textbf{(HS)}\end{tabular} & \cellcolor{lightgreen}LP, CP & \cellcolor{lightgreen}21 & \cellcolor{lightyellow} 311 & \cellcolor{lightgreen}Indoor/outdoor scenes \\ \hline 
        \end{tabular}
    \end{minipage}
\vspace{-2mm}
\caption{\textbf{Spectro-polarimetric image datasets.} We present trichromatic and hyperspectral Stokes datasets of which thumbnails are shown in (a) and label statistics in (b). The table shown on the right compares our datasets with existing spectro-polarimetric datasets. Ours are the only datasets that encompass both linear (LP) and circular (CP) polarization over multiple of spectral bands for diverse real scenes.}

    \vspace{-5mm}
    \label{fig:dataset}
\end{figure*}

\paragraph{Spectro-polarimetric Image Dataset}
Several datasets have been introduced for analyzing polarization and spectral information. 
With the advent of trichromatic linear-polarization cameras, a line of work has attempted to acquire trichromatic linear-polarization images, ranging from a few objects and scenes~\cite{ba2020deep, dave2022pandora, liang2022material, qiu2021demos} to a large number of scenes for specific target applications such as reflection separation~\cite{lei2020refl, lyu2019refl} and glass segmentation~\cite{mei2022glass}.
Lapray et al.~\cite{lapray2018data} acquire linear-polarization images for the near-infrared spectral band, albeit only on 10 objects.
Fan et al.~\cite{fan2023fullst} acquire the first multispectral full-Stokes polarimetric dataset covering linear and circular states, while it only contains 64 flat objects captured in a lab environment. 
Our proposed datasets enables analyzing the statistics of real-world spectro-polarimetric images, which cannot be achieved by prior datasets.
See Figure~\ref{fig:dataset} for a comprehensive comparison.

\begin{figure}[t]
	\centering
		\includegraphics[width=\linewidth]{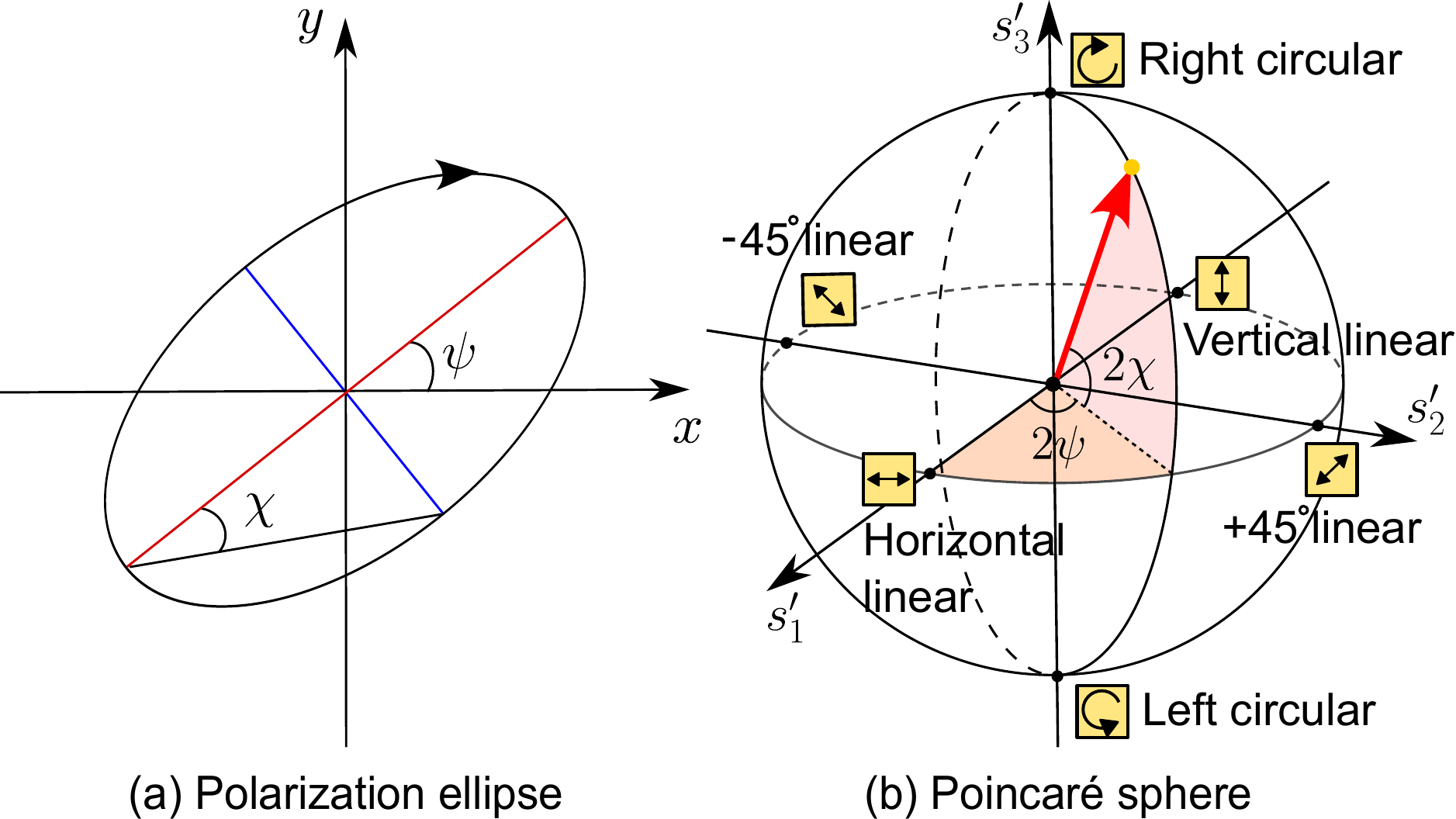}
            \vspace{-5mm}
		\caption{ \textbf{Two representative polarization visualizations.} (a) Polarization ellipse depicts the electric-field oscillation projected onto a plane tangent to the light propagation. (b) Poincaré sphere visualizes the polarization state of light on the normalized Stokes-vector axes $s'_1$, $s'_2$, $s'_3$. 
                }
		\label{fig:poincare_sphere}
  \vspace{-5mm}
\end{figure}

\paragraph{Applications of Spectro-polarimetric Imaging}
Spectro-polarimetric information has been investigated for diverse vision and imaging tasks.
Using linear-polarization images has found applications in shape reconstruction~\cite{kadambi2015polarized, lei2022wild, zou20203d, ba2019sfp, ba2020deep, baek2018simultaneous, ding2021polarimetric, fukao2021polarimetric}, appearance acquisition~\cite{deschaintre2021deep, kondo2020accurate}, removing reflections~\cite{nayar1997separation, kong2014refl, yang2016method, lyu2019refl, lei2020refl, wen2021polarization}, transparent-object segmentation~\cite{Kalra_2020_CVPR, mei2022glass}, seeing through scattering~\cite{fang2014dehaze, zhou2021learning, liu2015polarimetric}, and image enhancement \cite{zhou2023low}. 
Trichromatic Stokes images have been used for tone-mapping~\cite{del2019polarization} and seeing through scattering~\cite{baek2021polarimetric}. Expanding into multi-spectral domain, spectral-polarimetric analysis has been applied to object recognition~\cite{denes1998spectropolarimetric}, skin analysis~\cite{zhao2009spectropolarimetric}, dehazing~\cite{xia2016image}, specular reflection inpainting~\cite{islam2021specular}, background segmentation~\cite{islam2020hybrid} and tensor representation~\cite{zhang2017joint}. 
In addition to vision tasks, spectro-polarimetric imaging has been used for various biological applications, such as marsh vegetation classification~\cite{yao2022combination}, coastal wetland classification~\cite{tu2021synergetic} and leaf nitrogen determination~\cite{liu2021combining}. 
While the aforementioned studies demonstrate the benefits of using {spectro-polarimetric} data, we believe that the full potential of spectro-polarimetric images is still locked by the absence of real-world spectro-polarimetric datasets.

\section{Background on Polarization}
\label{sec:background}
\noindent
Polarization, the oscillation pattern of the electric field, can be represented using a Stokes vector, $\mathbf{s}= [s_0, s_1, s_2, s_3]^\intercal$. $s_0$ denotes the total intensity, $s_1$ and $s_2$ describe the differences in the intensity of linearly-polarized components at orientations of 0$^\circ$/90$^\circ$ and 45$^\circ$/-45$^\circ$, respectively. $s_3$ is the difference in intensity between right- and left-circularly polarized components. 

{Figure~\ref{fig:poincare_sphere} shows two visualization methods for polarization, the polarization ellipse and Poincaré sphere. Polarization ellipse can be described in terms of the orientation angle $\psi$ and ellipticity $\chi$ with respect to the projected Stokes vector x and y axes~\cite{collett2005field}. The Poincaré sphere visualizes polarization in a three-dimensional space, using the normalized Stokes-vector elements relative to the total intensity: $s'_1={s_1/s_0}, \, s'_2={s_2/s_0}, \, s'_3={s_3/s_0}$.} 
To effectively analyze a Stokes vector, one can compute the degree of polarization (DoP) denoted as $\rho$, the angle of linear polarization (AoLP) represented by $\psi$, and the ellipticity angle given by $\chi$, that is
\begin{equation}
\label{eq:polarimetric_parameter}
    \rho = \frac{P}{s_0}, \, \psi = \frac{1}{2}\arctan\left(\frac{s_2}{s_1}\right), \, \chi = \frac{1}{2}\arctan\left(\frac{s_3}{L}\right),
\end{equation}
where $P = \sqrt{s_1^2 + s_2^2 + s_3^2}$ and $L = \sqrt{s_1^2 + s_2^2}$. 

{We also use the polarimetric visualization method proposed by Wilkie et al.~\cite{wilkie2010standardised} using DoP, AoLP, and chirality of polarization (CoP).} 
CoP describes the left- or right-handedness of the circularly polarized component, which is related to $\chi$. 
Finally, the Mueller matrix $\mathbf{M}\in \mathbb{R}^{4\times4}$ describes the change of a Stokes vector: $\mathbf{s}_\text{out}=\mathbf{M}\mathbf{s}_\text{in}$,  where $\mathbf{s}_\text{in}$ and $\mathbf{s}_\text{out}$ are the input/output Stokes vectors. 
For more details on polarization, we refer to the book by Collett~\cite{collett2005field}.

\begin{figure*}[t]
	\centering
		\includegraphics[width=\linewidth]{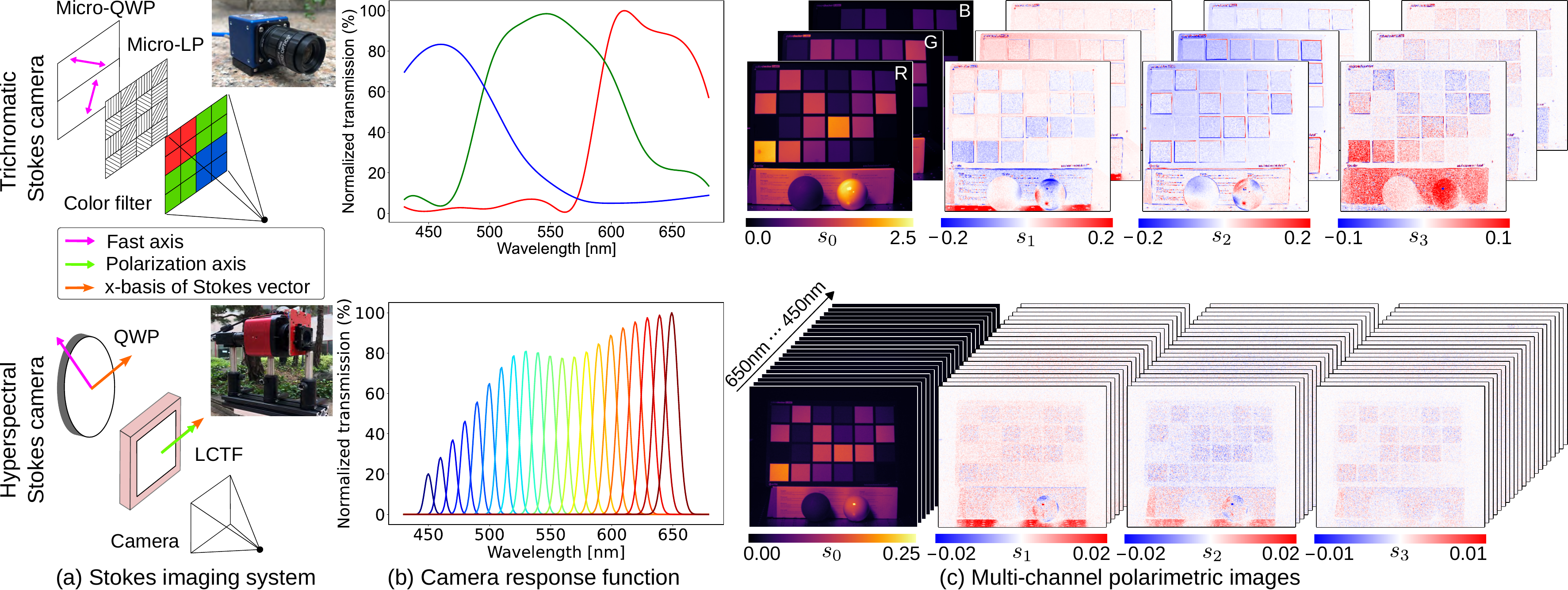}
            \vspace{-5mm}
		\caption{\textbf{Acquisition of spectro-polarimetric images.} We capture spectro-polarimetric images using (a) trichromatic and hyperspectral Stokes cameras~\cite{tu2017division,kim2023neural}. (b) Camera response functions. (c) Reconstructed raw Stokes images per each spectral channel. }
		\label{fig:image_formation}
  \vspace{-5mm}
\end{figure*}

\section{Spectro-polarimetric Dataset}
\label{sec:dataset}
\noindent
We introduce a trichromatic Stokes dataset comprising {2022} Stokes images and a hyperspectral Stokes dataset with {311} Stokes images at 21 spectral channels. Both datasets cover natural real-world indoor and outdoor scenes. Each Stokes image is accompanied by four labels detailing: (1) the environment (indoor or outdoor), (2) the illumination condition, including clear or cloudy sunlight and white or incandescent light, (3) the time of image capture, (4) the scene type, distinguishing between object-oriented and scene-oriented.
Figure~\ref{fig:dataset} shows thumbnails, label statistics, and comparison of our datasets to existing ones.
Prior datasets suffer from a narrow range of scenes, restricted illumination conditions, linear polarization only, and fewer images. 

\paragraph{Acquisition}
We acquire the datasets using two imaging systems depicted in Figure~\ref{fig:image_formation}(a), proposed and developed by previous studies~\cite{tu2020divi,kim2023neural}. 
First, the trichromatic Stokes camera by Tu et al. \cite{tu2020divi} incorporates on-sensor quarter-wave plates (QWPs) and linear polarizers (LPs)~\cite{collett2005field}. This allows for single-shot capture of trichromatic Stokes images, enabling convenient data collection on diverse scenes. 
The resolution of a trichromatic Stokes image is $2100 \text{ (width)} \times1920 \text{ (height)}\times3\text{ 
 (RGB)}\times4\text{ (Stokes elements)}$.
Second, the hyperspectral Stokes camera from Kim et al. \cite{kim2023neural} captures images by sequentially scanning 21 spectral channels from 450\,nm to 650\,nm in 10\,nm increments with a LCTF which functions as a LP. For each spectral channel, we capture images by rotating a QWP with a fixed LP. This system enables a detailed analysis of the interplay between wavelength and polarization. 
The resolution of a hyperspectral Stokes image is $612 \text{ (width)} \times 512 \text{ (height)}\times 21\text{ 
 (wavelengths)}\times4\text{ (Stokes elements)}$.

\paragraph{Spectro-polarimetric Image Formation}
Using the two imaging systems, we capture raw images from which a per-pixel Stokes vector for each spectral channel is reconstructed.
We introduce an unified image formation model that can be applied to both cameras. 
Suppose a light ray with a Stokes vector $\mathbf{s}_\lambda$ at wavelength $\lambda$ impinges on a Stokes camera. As the light passes through polarization-modulating optical filters modeled by the Mueller matrix $\mathbf{M}(\Theta)$, its Stokes vector transforms. $\Theta$ denotes the polarization-filter configuration. The camera sensor then captures light intensity, represented by the first element of the Stokes vector.
The recorded intensity, ${I}_{c}(\Theta)$, at a spectral channel $c$ and polarimetric filter configuration $\Theta$, is described by
\begin{align}
\label{eq:image_formation}
    {I}_{c}(\Theta)
    &= \left[\int{\Omega_{{c}, \lambda} \mathbf{M}_c(\Theta) \mathbf{s}_\lambda} d\lambda \right]_0 \nonumber \\
    &= \left[\mathbf{M}_c(\Theta) \int{\Omega_{{c}, \lambda}  \mathbf{s}_\lambda} d\lambda \right]_0 \nonumber \\
    &= \left[\mathbf{M}_c(\Theta)\mathbf{s}_c \right]_0,
\end{align}
where $\Omega_{{c}, \lambda}$ is the spectral transmission per channel at wavelength $\lambda$ shown in Figure~\ref{fig:image_formation}(b). $[x]_0$ denotes the first-element of the Stokes vector $x$, which is the total intensity. For a spectral channel $c$, $\mathbf{M}_c$ is the Mueller matrix of the polarization-modulating optics, and $\mathbf{s}_c$ is the Stokes vector. 

For polarization modulation, both cameras utilize a QWP and a LP, yielding the Mueller matrix
\begin{align}
\label{eq:polar_modulation}
\mathbf{M}_c(\Theta) = \mathbf{C}_c\mathbf{Q}_c(\theta_1)\mathbf{P}_c(\theta_2),
\end{align}
where $\mathbf{C}_c$ is the error-compensating calibration matrix~\cite{tu2017division,kim2023neural}. $\mathbf{Q}_c$ and $\mathbf{P}_c$ are the QWP and LP Mueller matrices~\cite{collett2005field}, respectively. The set $\Theta=\{\theta_1, \theta_2\}$ denotes the corresponding angles of the QWP fast axis and the LP polarization axis, {which is set for accurate Stokes vector reconstruction~\cite{tu2017division,kim2023neural}}.
Lastly, we determine the per-channel Stokes vector $\mathbf{s}_c$ by solving the least-squares problem
\begin{align}
\label{eq:stokes_vector_estimation}
    \underset{\mathbf{s}_c}{\mathrm{argmin}}\sum_{i=1}^{|\Theta|}{({I}_{c}(\Theta_i) - \left[\mathbf{M}(\Theta_i) \mathbf{s}_c \right]_0)}^2.
\end{align} 
For the hyperspectral Stokes camera, we use four configurations with the rotating QWP.
For the trichromatic Stokes camera, the fixed micro-filter setup shown in Figure~\ref{fig:image_formation}(a) gives four/eight configurations for the (red, blue) channels and the green channel, respectively.

Figure~\ref{fig:image_formation}(c) shows the reconstructed Stokes images. 
A Stokes vector is physically-valid if DoP meets the following inequality: $0 \leq \rho \leq 1$. 
99\% of the reconstructed Stokes vectors in our datasets satisfy the DoP condition. 
For the following dataset analysis, we filter out Stokes vectors violating the DoP condition and the unstable Stokes vectors reconstructed from saturated {and underexposed} pixel intensity values.

\begin{figure*}[t]
    \centering
    \includegraphics[width=0.95\linewidth]{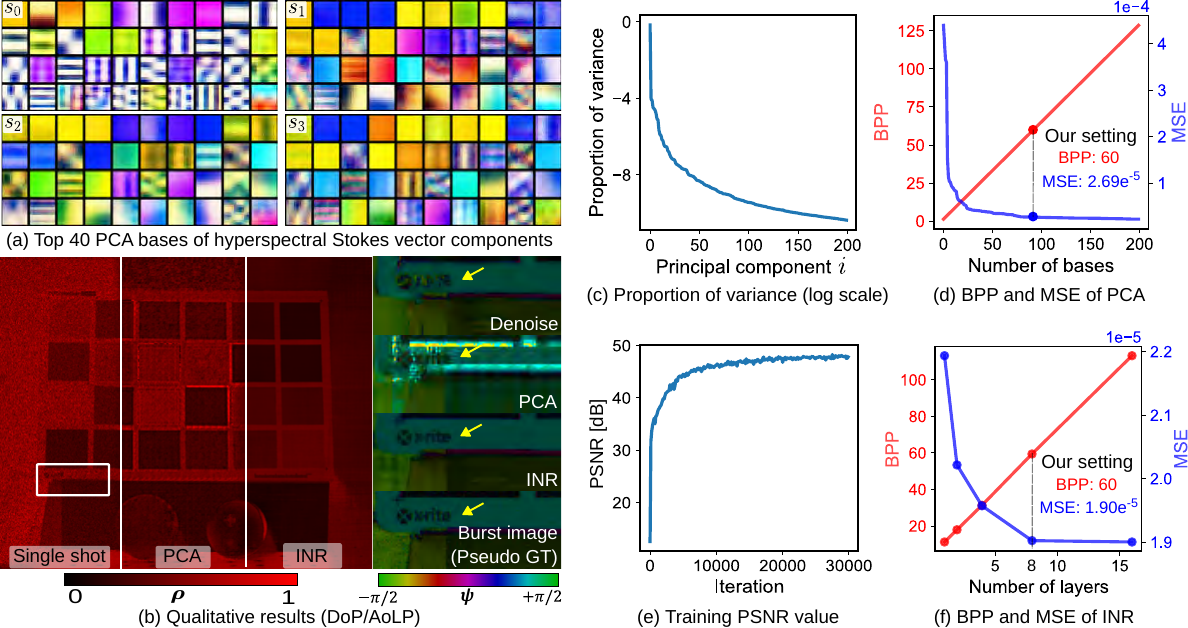}
    \vspace{-2mm}
    \caption{
    \textbf{Efficient spatio-spectral-polarimetric representations.} 
    (a) PCA basis of the hyperspectral Stokes dataset in sRGB. (b) Qualitative results of PCA and INR compared to single-shot denoising~\cite{zhang2023kbnet} at 550\,nm. (c) Proportion of variance with respect to each PCA basis in order, $\log(\sigma_i^2/\sum_n\sigma_n^2)$, where $\sigma_i$ denotes standard deviation of the $i$-th basis. (d) BPP and MSE analysis of PCA with respect to the number of PCA bases. (e) Training PSNR curve of INR.  (f) BPP and MSE value of INR with respect to the number of MLP layers.
    }
    \vspace{-5mm}
    \label{fig:representations}
\end{figure*}
\begin{figure}[t]
	\centering
		\includegraphics[width=\columnwidth]{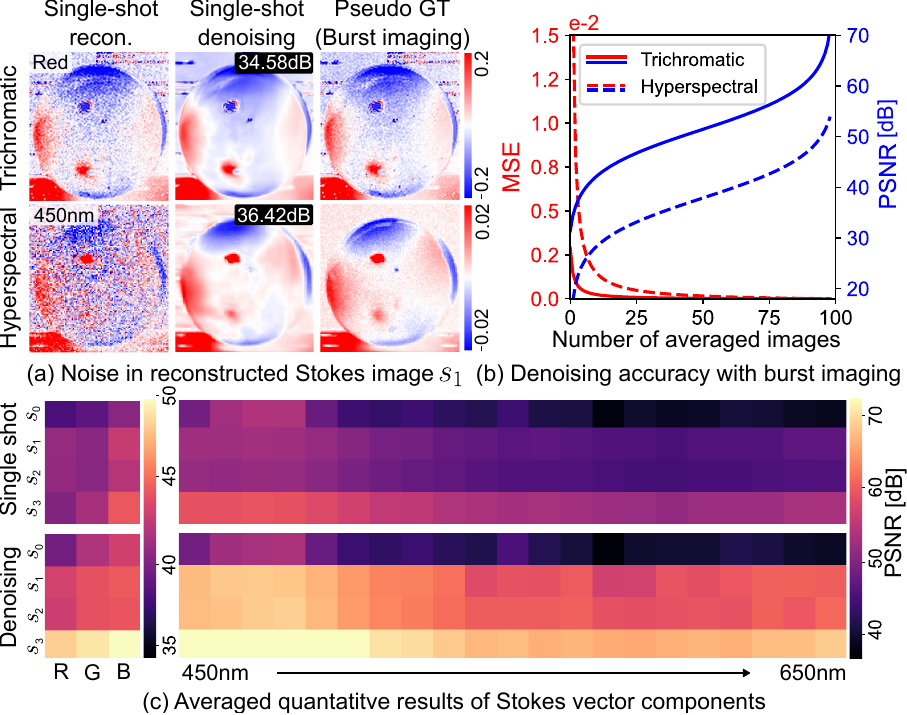}
            \vspace{-3mm}
		\caption{\textbf{Noise in Stokes images and intensity denoising.} (a) Stokes vector $s_1$ reconstructed from a single-shot, single-shot with a learned intensity denoiser~\cite{zhang2023kbnet}, and burst imaging (Pseudo GT) averaged over 100 shots. (b) Reconstruction accuracy of a Stokes image with varying number of averaged images. (c) Reconstruction accuracy of Stokes elements with and without intensity denoising. 
        }
        \vspace{-5mm}
		\label{fig:denoising}
\end{figure}

\section{Dataset Analysis}
\label{sec:analysis}
\noindent
We provide our analysis of the acquired Stokes datasets.

\paragraph{Noise and Intensity Denoising}
Spectro-polarimetric images are susceptible to noise due to the low-light throughput of spectral and polarimetric filters. Our datasets are not exempt from these issues.
To assess noise in Stokes images, we capture and average 100 images of a scene shown in Figure~\ref{fig:image_formation}(c) for each polarization configuration $\Theta$. From these averaged images, we reconstruct pseudo ground-truth Stokes image, shown in Figure~\ref{fig:denoising}(a).
Figure~\ref{fig:denoising}(b) reports the MSE and PSNR of reconstructed Stokes images with respect to the number of averaged images. To achieve a PSNR exceeding 35\,dB, we need to average over 4/25 shots for the trichromatic/hyperspectral Stokes cameras, indicating lower SNR of the hyperspectral Stokes dataset.
{We find that state-of-the-art learning-based intensity denoising methods, such as KBNet~\cite{zhang2023kbnet}, can effectively reduce noise for each polarization configuration, leading to accurate Stokes-vector reconstruction, despite lack of polarization images during training.} 
For the denoised single-shot capture, we achieve a PSNR of 34.5\,dB, demonstrating the potential of using pretrained intensity restoration networks for Stokes imaging. 
Figure~\ref{fig:denoising}(c) shows PSNRs of reconstructed Stokes images per each spectral channel and Stokes element. With the intensity denoising, we find that the PSNR significantly increases for $s_1$, $s_2$, and $s_3$.

\paragraph{Efficient Spatio-spectral-polarimetric Representations} 
Each pixel in a hyperspectral Stokes image contains a Stokes vector for every spectral channel, leading to a total of $21\times4\times32$ bits using single-precision floating format. This results in a bit-per-pixel (BPP) value of 2,688, equating to 100\,MB for storing a single hyperspectral Stokes image of $512\times612$ pixels. Given the substantial memory required to store a spectro-polarimetric image, we investigate efficient representations of real-world spatio-spectral-polarimetric data, for which we explore two methods: a PCA-based model and implicit neural model.

First, we conduct PCA on non-overlapping hyperspectral Stokes patches. Figure~\ref{fig:representations}(a) shows the 40 most significant PCA bases, revealing varied spatial and spectral features for each Stokes element: $s_0$, $s_1$, $s_2$, $s_3$. Notably, spatial structures are more evident in $s_0$, while $s_1$, $s_2$, $s_3$ shows spectral features, suggesting a stronger correlation between spectrum and polarization than spatial features. To visualize hyperspectral intensity, we convert it to sRGB, which means that the same sRGB color may originate from different spectra. Figure~\ref{fig:representations}(c) shows the variance of the coefficients for the top 200 PCA bases, indicating that spatio-spectral-polarimetric data can indeed be compressed. This is further evidenced by Figure~\ref{fig:representations}(d), which shows the reconstruction error and BPP when varying number of PCA bases used to recreate a hyperspectral Stokes image. Using 2.22\,MB coefficients adequately represents a 100\,MB hyperspectral Stokes image  as shown in Figure~\ref{fig:representations}(b), exhibiting a high compression rate {with the reconstruction error of $2.69\times10^{-5}$}. 
See the Supplemental Document for further details on PCA analysis. 

Second, we devise an INR for hyperspectral Stokes images by modifying the recently-proposed network architecture, NeSpoF~\cite{kim2023neural}. The original NeSpoF architecture models a volumetric hyperspectral Stokes field.
Here, instead, we aim to represent a hyperspectral Stokes image. Specifically, our INR, modeled by an MLP $F_\gamma$, outputs the Stokes vector $\mathbf{s}$ for a given pixel position $p_x, p_y$ and spectral channel index $c$, that is
\begin{equation}
{\mathbf{s} = F_\gamma(p_x, p_y, c),}
\end{equation}
where $\gamma$ is the network parameters. 
We fit the MLP to a hyperspectral Stokes image by minimizing the reconstruction loss between the network output and the hyperspectral Stokes image. The training curve is shown in Figure~\ref{fig:representations}(e). Figure~\ref{fig:representations}(f) shows the reconstruction error and BPP of our INR with respect to varying number of the MLP layers. With just 8 layers corresponding to a BPP of 60, we achieve a converged reconstruction error of $1.90\times10^{-5}$, resulting in just 2.22\,MB of network parameters to represent a 100\,MB hyperspectral Stokes image.
See Supplemental Document for architecture details. 

Both PCA and INR experiments validate that a natural spectro-polarimetric image is compressible. PCA provides PCA basis vectors that can be applied to any instance, however with a lower reconstruction accuracy than INR.
INR is overfitted to a single instance, while higher reconstruction accuracy can be achieved. These representations are also beneficial for denoising spectral-polarimetric images, as shown in Figure~\ref{fig:representations}(b), which even outperforms the learning-based intensity denoiser~\cite{zhang2023kbnet}.

\begin{figure}[t]
	\centering
		\includegraphics[width=\linewidth]{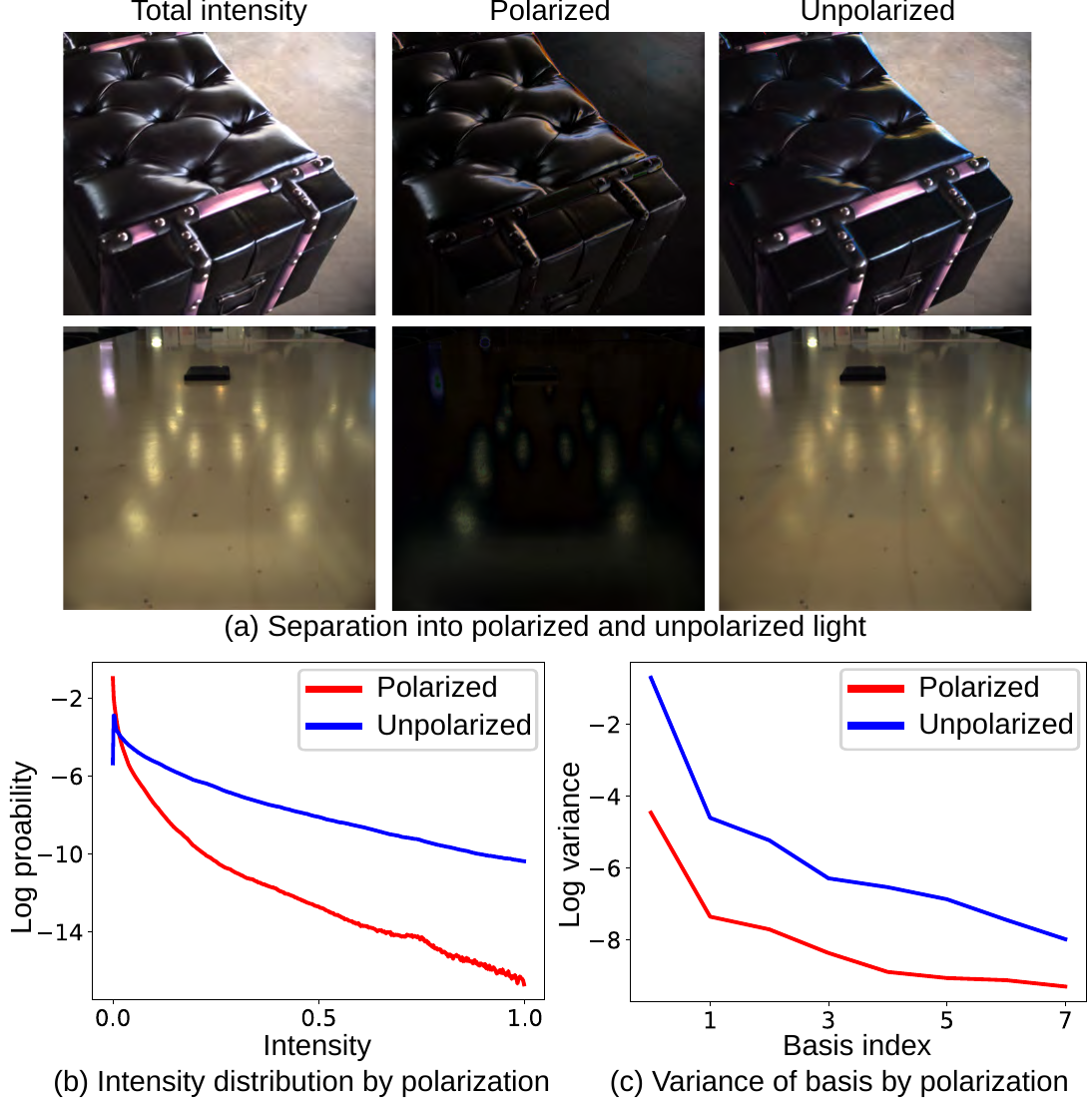}
		\caption{\textbf{Polarized and unpolarized light distributions.} (a) Separation into polarized and unpolarized light. (b) Intensity distributions for polarized and unpolarized components. (c) Variance of PCA basis of polarized and unpolarized intensity across spectral channel.}
		\label{fig:specular_diffuse}
  \vspace{-5mm}
\end{figure}

\paragraph{Polarized and Unpolarized Intensity} 
We decompose hyperspectral Stokes images into the polarized images $P=\sqrt{s_{1}^2 + s_{2}^2 + s_{3}^2}$ and unpolarized images $U=s_{0} - P$ per each spectral channel. 
Figure~\ref{fig:specular_diffuse}(a) shows specular reflections such as the glow of leather sofa separated into polarized light. Note that the polarized image typically encodes the illumination colors for dielectric surfaces. 
Figure~\ref{fig:specular_diffuse}(b) reveals that the intensity distributions of polarized light, obtained from the entire hyperspectral Stokes dataset, is skewed towards low and high-intensity values compared to the unpolarized light.
This is because polarized images mostly contain specular reflections, which is sparsely distributed and has high intensity values. 
We then compute the variance of the PCA bases for polarized intensity along the spectral channel.
Figure~\ref{fig:specular_diffuse}(c) highlights that the spectral variance for polarized intensity is lower than that of unpolarized intensity. We speculate that the color of polarized light lies in a lower dimensional space than {that of} unpolarized light, since diffuse reflection with diverse spectral variations is mostly captured by unpolarized light, making the spectral variation of unpolarized light more pronounced.

\paragraph{Stokes Vector Distributions in Natural Stokes Images}

\noindent
\begin{figure}[t]
	\centering
		\includegraphics[width=\linewidth]{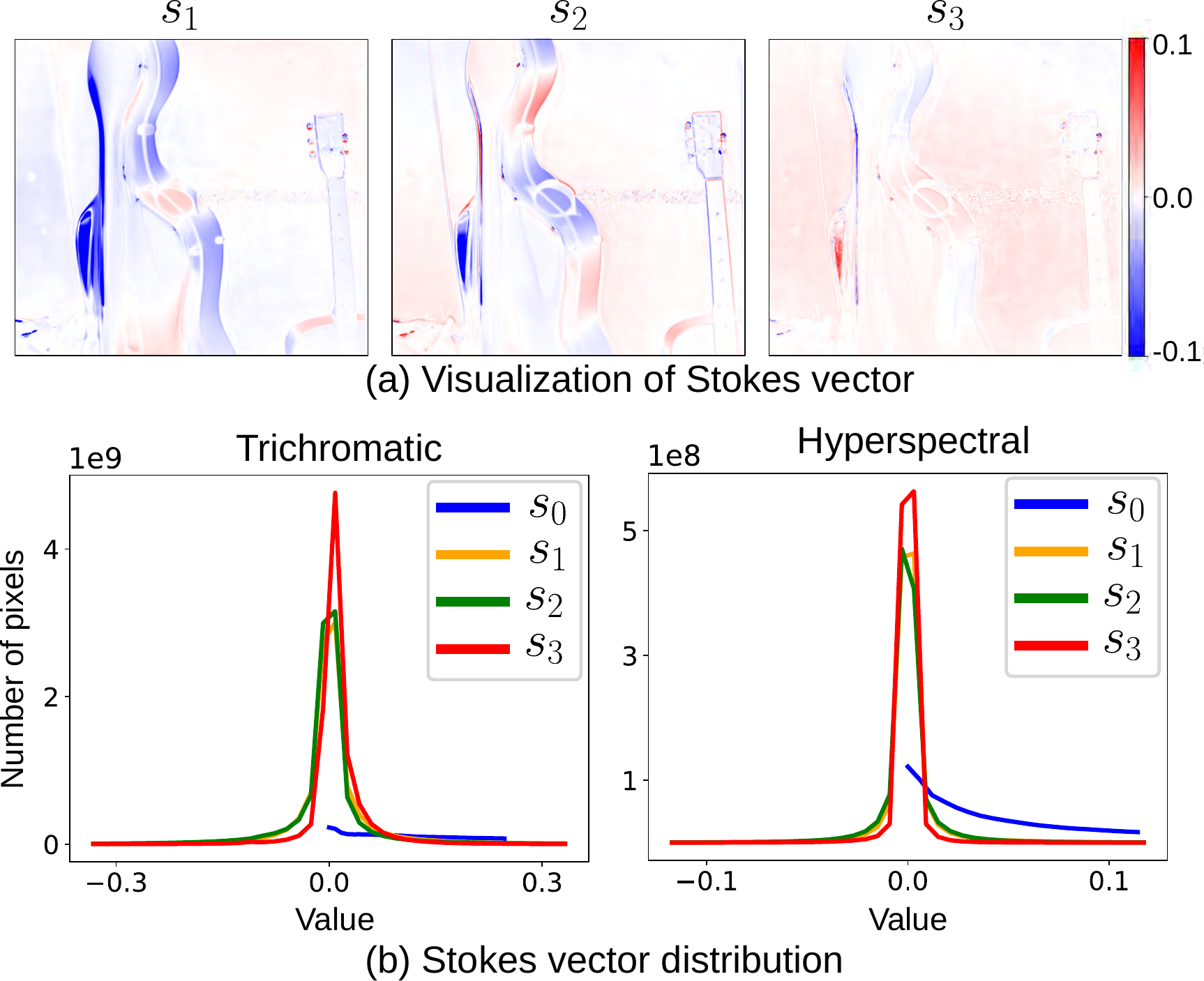}
		\caption{\textbf{Stokes-vector distributions.} (a) Stokes images of $s_1$, $s_2$, and $s_3$ at the green channel. (b) Stokes-vector distributions of $s_0$, $s_1$, $s_2$ and $s_3$ for trichromatic and hyperspectral datasets.
  }
  \vspace{-4mm}
		\label{fig:stokes_distribution}
\end{figure}
Next, we analyze the distribution of all Stokes vectors in our Stokes dataset. Figure~\ref{fig:stokes_distribution} reports the histograms of Stokes elements $s_0$, $s_1$, $s_2$, $s_3$ across all spectral channel. We find that the distributions of Stokes elements ($s_1$, $s_2$, $s_3$) have symmetric shapes of positive and negative sides. 
Stokes elements of $s_1$ and $s_2$ have similar shapes meaning that the directions of linearly-polarized light are equally distributed in natural images.
The circular component $s_3$ is more condensed near zero than the linear elements, resulting in a higher peak both in trichromatic and hyperspectral datasets.
This indicates that pixels are often more linearly polarized than circularly polarized. 
Refer to the Supplemental Document for further analysis.

\begin{figure}[t]
	\centering
		\includegraphics[width=\linewidth]{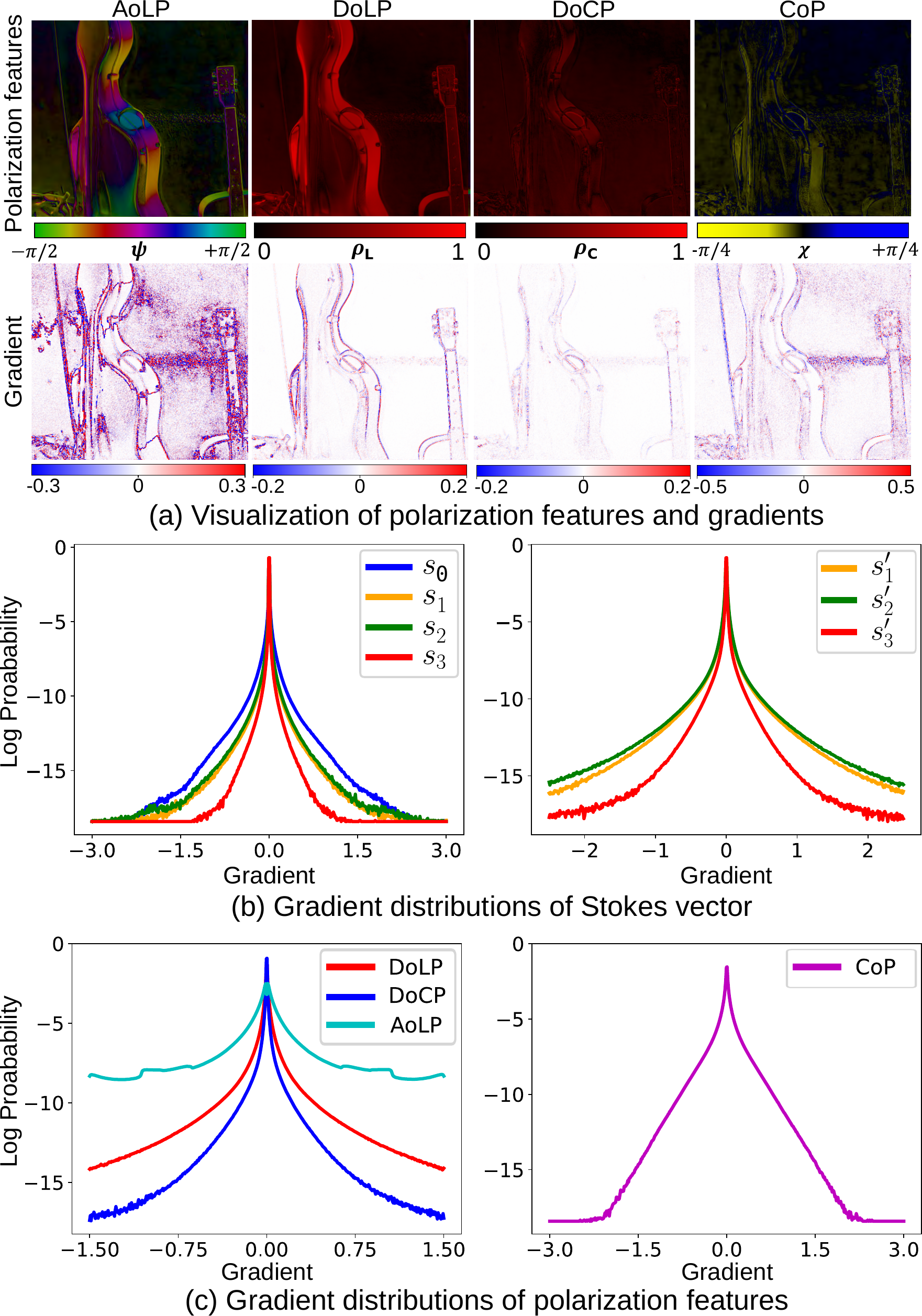}
		\caption{\textbf{Gradient analysis of Stokes images.} (a) Visualization of AoLP, DoLP, DoCP and CoP values and their gradients. Examining the log probability of the gradient for (b) Stokes vectors $s_0$, $s_1$, $s_2$, $s_3$ and normalized Stokes vectors $s'_1$, $s'_2$, and $s'_3$, and (c) polarization features including DoLP, DoCP, AoLP and CoP.}
		\label{fig:gradient}
  \vspace{-4mm}
\end{figure}

\paragraph{Gradient Analysis of Stokes Images}
Gradient distribution of images has been often used as priors for image-based applications including image restoration, understanding, and editing. 
Here, we perform gradient analysis of Stokes and polarization-feature images. Figure~\ref{fig:gradient}(b) shows that the gradient of Stokes and normalized Stokes vectors exhibits a similarity to Hyper-Laplacian priors, commonly used to describe the gradient of natural intensity-images. 
An interesting finding is that total intensity $s_0$ has more high-gradient values than the linear components of $s_1$ and $s_2$, and the circular component $s_3$ has the lowest-value distribution.

We then analyze the gradient distributions of polarization features, including AoLP, degree of linear polarization (DoLP), degree of circular polarization (DoCP), and CoP.
DoLP and DoCP are computed as $\mathrm{DoLP}=\sqrt{s_1^2 + s_2^2}/s_0$ and $\mathrm{DoCP}=|s_3|/s_0$ respectively.
To compute the gradient of AoLP images, we consider the angular wrapping property. 
That is, AoLP has a range from $-\frac{\pi}{2}$ to $\frac{\pi}{2}$ and the AoLPs of $-\frac{\pi}{2}$ and $\frac{\pi}{2}$ are identical. Thus, if the gradient exceeds $\frac{\pi}{2}$, we estimate the gradient  as $\nabla \text{AoLP} -\pi \times \texttt{sign}(\nabla \text{AoLP})$, where $\nabla$ is the gradient operator and $\texttt{sign}$ is the sign operator that returns $1$ if positive, otherwise $-1$. Figure~\ref{fig:gradient}(c) shows that the gradients of AoLP are generally higher than DoLP and DoCP.
This implies that sparsity in the measurement gradient is milder for AoLP than DoLP and DoCP.
DoLP and DoCP have shapes similar to Hyper-Laplacian priors while DoCP is sharper than DoLP. The gradient of CoP shows symmetric distributions for right-circular and left-circular directions. Unlike AoLP, the probability decreases as the gradient approaches extreme values.  The difference in tendency between linear-polarization and circular-polarization features, as well as their distributions, means that we need distinct priors for each polarization feature, emphasizing the importance of a full Stokes dataset that measure not only linear but also circular polarization.

\begin{figure}[t]
	\centering
		\includegraphics[width=\columnwidth]{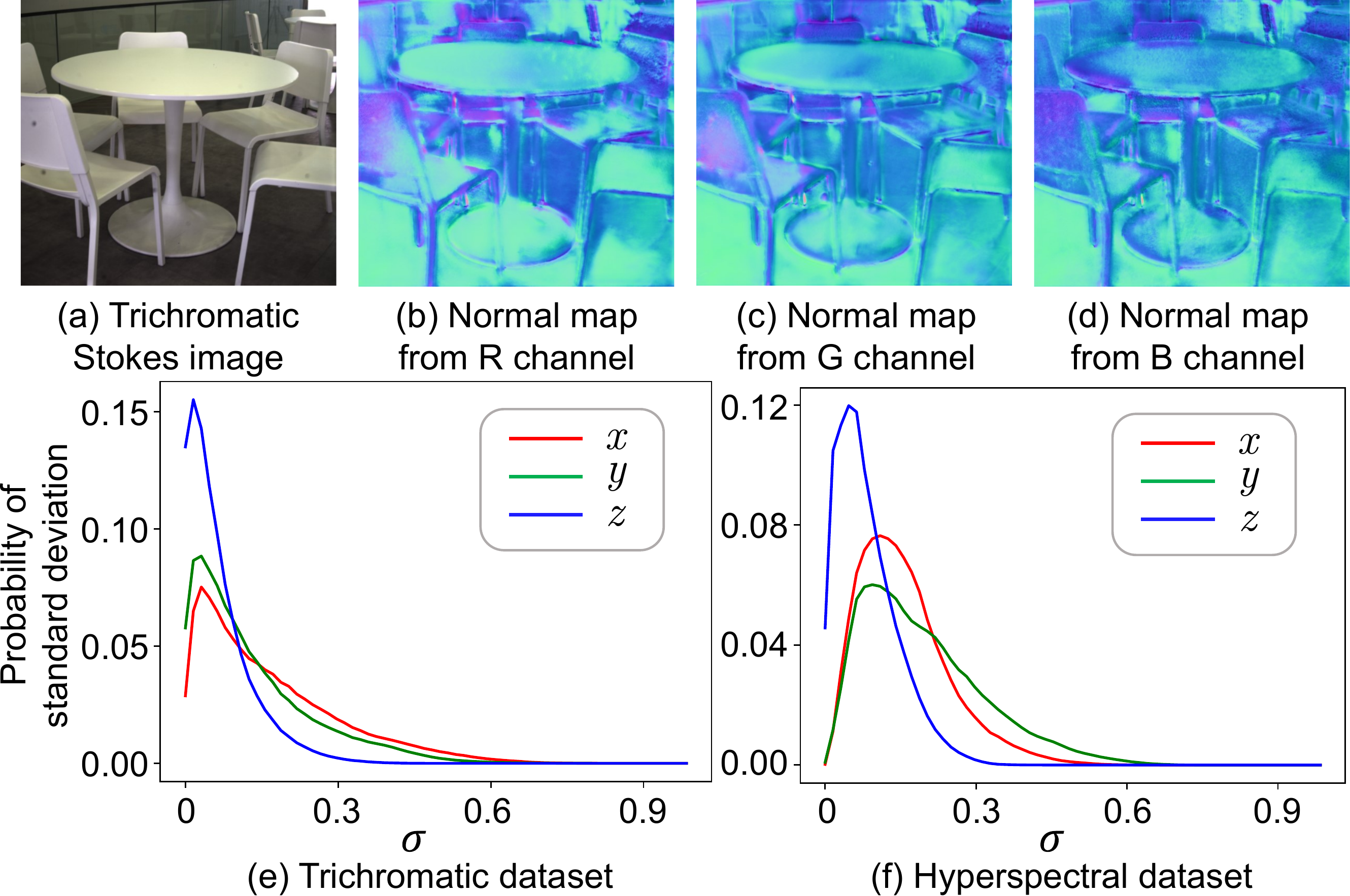}
		\caption{\textbf{Spectral dependency of conventional SFP method.} (a) Trichromatic Stokes image and estimated surface normals~\cite{lei2022wild} for each red, green, and blue spectral channels shown in (b), (c), and (d). Graphs (e) and (f) show the probability distributions of standard deviation of normal component $x$, $y$, and $z$ along the spectral channels for both datasets.}
		\label{fig:sfp}
  \vspace{-4mm}
\end{figure}

\paragraph{Shape from Polarization and Spectral Channels} 
Methods that recover shape from polarization, SfP, have made rapid progress in the last decade. SfP aims to extract normals by analyzing the normal-dependent polarization change of reflected light. Specifically, SfP analyzes the DoP and AoLP based on Fresnel theory~\cite{collett2005field}, which describes the polarization change of light upon reflection and transmission at a smooth surface~\cite{kadambi2017depth,lei2022wild}. Here, we analyze an overlooked problem in SfP: the spectral dependency of estimated normals. Surface normals, as a geometric surface property, should be consistent regardless of the input spectral channels used for SfP. In Figure~\ref{fig:sfp}, we test the state-of-the-art SfP method by Lei et al.~\cite{lei2022wild}, designed for in-the-wild scenes. The evaluation results on our Stokes dataset clearly reveal that normal maps reconstructed from different spectral channels exhibit variations. We compute the standard deviations of spectral variations for each $x$, $y$, and $z$ component of the estimated normals. Figures~\ref{fig:sfp}(e) and (f) show the probability distributions of the standard deviation, highlighting the large variations in the estimated normals for both hyperspectral and trichromatic datasets. Interestingly, the $x$ and $y$ components of normals show larger standard deviations than the $z$ component. This implies that the spectral variation of DoP, which determines the $z$ component, has less impact on the distribution than that of AoLP, which governs the $x$ and $y$ components. Our analysis underscores the development of a spectrum-aware SfP method as an interesting future research.

\begin{figure}[t]
	\centering
		\includegraphics[width=\columnwidth]{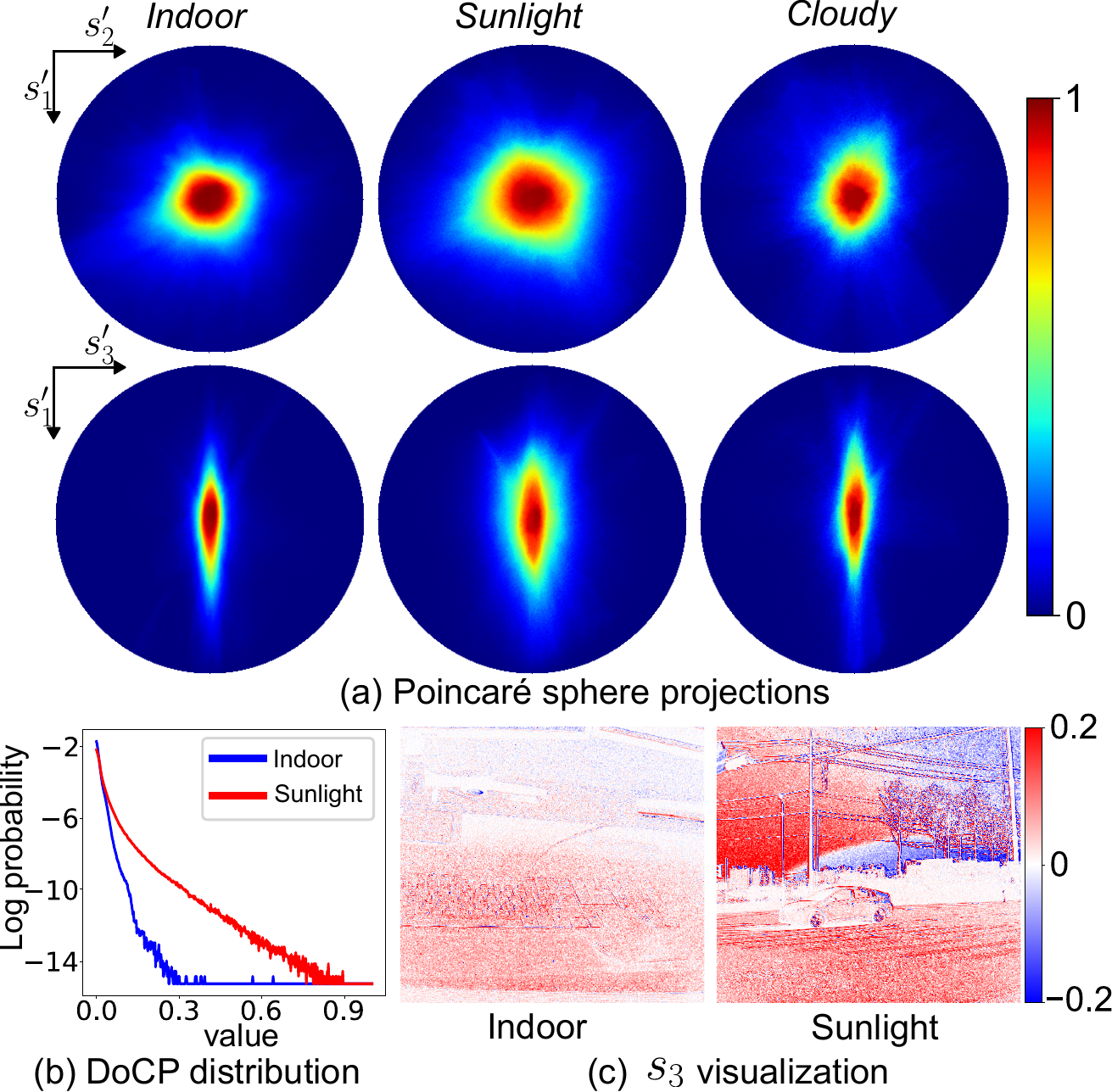}
		\caption{\textbf{Environment dependency.} (a) Projected Poincaré spheres onto the $s_1'-s_2'$ and $s_1'-s_3'$ planes with repsect to dataset labels. Colorbar is based on the normalized pixel count. (b) DoCP distributions for indoor and sunlight cateogries. (c) Example $s_3$ images.}
		\label{fig:environment_dependency}
  \vspace{-4mm}
\end{figure}

\paragraph{Environment Dependency} 
Figure~\ref{fig:environment_dependency} shows the Poincaré spheres projected onto the $s_1'-s_2'$ and $s_1'-s_3'$ planes for the three data labels: \textit{Indoor}, \textit{Sunlight} and \textit{Cloudy}. 
Sunlight is known to contain more circularly polarized light compared to other artificial lighting~\cite{horvath2004polarized}. As shown in Figure ~\ref{fig:environment_dependency}(a), Stokes vectors are distributed more widely across $s_3'$ axis under sunlight compared with Indoor scene.
In addition, we find that DoCP is distributed at higher values for the sunlight compared to the indoor: pixels with DoCP over $0.5$ are rarely observed in indoor scenes.
This is also confirmed in the example $s_3$ images for indoor and sunlight scenes. 

Another interesting finding is that cloudy or sunny illumination result in different polarization statistics.
Figure~\ref{fig:environment_dependency}(a) shows that Stokes vectors of cloudy scenes are more concentrated near the origin, meaing that light is more depolarized compared to light under clear sunlight. 
This is aligned with previous studies~\cite{vanderLaan:15,Hegedus:07} that report the impact of cloud-particle scatterings on light depolarization. 

\section{Conclusion}
\noindent
In this work, we have introduced a novel trichromatic and hyperspectral Stokes image dataset that encompasses diverse natural scenes and various illumination conditions, totaling more than 2,333 scenes. We analyze the empirical distribution of the Stokes vectors of natural spectro-polarimetric images. To efficiently represent spatio-spectral-polarimetric data, we devise a PCA-based model and an implicit neural representation. We further provide detailed analysis on Stokes gradient distributions, denoising characteristics, spectral dependency of SfP, and environment dependency.  As such, our work provides a foundation for future research on spectral-polarimetric imaging and vision.

{\small
\bibliographystyle{template/ieee_fullname}
\bibliography{references}
}

\end{CJK}
\end{document}


\begin{CJK}{UTF8}{}
\CJKfamily{mj}

\title{
Spectral and Polarization Vision: Spectro-polarimetric Real-world Dataset\\
(Supplementary Information)}

\author{Yujin Jeon$^{1,*}$~ ~ ~
Eunsue Choi$^{1,*}$ ~ ~ ~
Youngchan Kim$^{1}$ ~ ~ ~
Yunseong Moon$^{1}$ ~ ~ ~ \\
Khalid Omer$^{2}$ ~ ~ ~
Felix Heide$^{3}$ ~ ~ ~
Seung-Hwan Baek$^{1}$ \\[2mm]
$^1$ POSTECH   ~ ~ ~   $^2$ Meta   ~ ~ ~  $^3$ Princeton University\\
}

\maketitle
\def\thefootnote{*}\footnotetext{Equal contribution}\def\thefootnote{\arabic{footnote}}

\hypersetup{linkcolor=black}
\noindent
In this supplemental document, we provide additional analysis and details in support of the findings in the main manuscript. 

\tableofcontents

\newpage

\section{Additional Comparison to Existing Datasets}

We further highlight here the difference in dataset diversity between our datasets and prior works. The common datasets used in the field have been limited to polarization states, the number of spectral bands, scene count, and scene diversity, whereas our study expands the scope to include factors mentioned above. 

Table~\ref{tab:dataset_comparison} provides a comprehensive comparison between previous studies and our current research at a glance. 
In the "Polar." column, cells colored red indicate datasets encompassing only linear polarization states, whereas green cells denote datasets that include both linear and circular polarization states. Assessing the "\# of bands" column, datasets providing grayscale polarization data are marked in red. Those offering spectral bands of 10 or fewer are colored yellow, while green cells indicate datasets that provide Stokes vectors for more than 10 spectra. In terms of scene count, works featuring 1 to 100 scenes are highlighted in red, those with 101 to 1000 scenes are in yellow, and green cells represent datasets with more than 1001 scenes. For scene diversity, datasets colored red consist of very limited scenes, such as synthetic scenes or objects in consistent conditions. Yellow cells represent limited scene types, like diverse objects or scenes within restricted environments, whereas green cells indicate natural scenes encompassing diverse environments.

 \begin{table*}[h]
        \centering
        \begin{tabular}{|c|c|c|c|c|}
            \hline
            Dataset & Polar. & \#\mbox{ of} bands & Scene count & Scene diversity \\ 
            \Xhline{2\arrayrulewidth}\Xhline{0.2pt}
\cite{lei2022wild} & \cellcolor{lightpink}LP & \cellcolor{lightpink}1 & \cellcolor{lightyellow}522 & \cellcolor{lightyellow}Outdoor scenes \\ 
\cite{ba2020deep} & \cellcolor{lightpink}LP & \cellcolor{lightpink}1 & \cellcolor{lightyellow}300 & \cellcolor{lightyellow}Indoor objs. \\ 
\cite{lapray2018data} & \cellcolor{lightpink}LP & \cellcolor{lightyellow}6 & \cellcolor{lightpink}10 & \cellcolor{lightyellow}Indoor objs. \\ 
\cite{qiu2021demos} & \cellcolor{lightpink}LP & \cellcolor{lightyellow}3 & \cellcolor{lightpink}40 & \cellcolor{lightyellow}Indoor objs. \\ 
\cite{dave2022pandora} & \cellcolor{lightpink}LP & \cellcolor{lightyellow}3 & \cellcolor{lightpink}3 & \cellcolor{lightyellow}Indoor multiview\\ 
\cite{dave2022pandora} & \cellcolor{lightpink}LP & \cellcolor{lightyellow}3 & \cellcolor{lightpink}2 & \cellcolor{lightyellow}Synthetic multiview\\ 
\cite{gao2022ppp} & \cellcolor{lightpink}LP & \cellcolor{lightyellow}3 & \cellcolor{lightpink}6 & \cellcolor{lightyellow}Indoor multiview\\ 
\cite{lei2020refl} & \cellcolor{lightpink}LP & \cellcolor{lightyellow}3 & \cellcolor{lightyellow}807 &  \cellcolor{lightyellow}Reflective objs.   \\ 
\cite{liang2022material} & \cellcolor{lightpink}LP & \cellcolor{lightyellow}3 & \cellcolor{lightyellow}500 & \cellcolor{lightyellow}Outdoor scenes \\ 
\cite{kondo2020accurate} & \cellcolor{lightpink}LP & \cellcolor{lightyellow}3 & \cellcolor{lightgreen}44,300 & \cellcolor{lightpink}Synthetic \\ 
\cite{lyu2019refl} & \cellcolor{lightpink}LP & \cellcolor{lightyellow}3 & \cellcolor{lightgreen}3,200 & \cellcolor{lightyellow}Reflective objs.   \\ 
\cite{mei2022glass} & \cellcolor{lightpink}LP & \cellcolor{lightyellow}3 & \cellcolor{lightgreen}4,500 &  \cellcolor{lightyellow}Transparent objs.  \\ 
\cite{kurita2023sensor} & \cellcolor{lightpink}LP & \cellcolor{lightyellow}3 & \cellcolor{lightgreen}2,000 & \cellcolor{lightgreen}Indoor/outdoor scenes \\ 
\cite{fan2023fullst} & \cellcolor{lightgreen}LP, CP & \cellcolor{lightgreen}18 & \cellcolor{lightpink}67 & \cellcolor{lightpink}Flat objs. \\ 
\cite{kim2023neural} & \cellcolor{lightgreen}LP, CP & \cellcolor{lightgreen}21 & \cellcolor{lightpink}4 & \cellcolor{lightpink}Synthetic multiview \\ 
\cite{kim2023neural} & \cellcolor{lightgreen}LP, CP & \cellcolor{lightgreen}21 & \cellcolor{lightpink}4 & \cellcolor{lightgreen}Indoor/outdoor multiview\\ 
\hline
\textbf{Ours} \textbf{(RGB)} & \cellcolor{lightgreen}LP, CP & \cellcolor{lightyellow}3 & \cellcolor{lightgreen} 2022  & \cellcolor{lightgreen}Indoor/outdoor scenes \\ 
\textbf{Ours} \textbf{(hyperspectral)} & \cellcolor{lightgreen}LP, CP & \cellcolor{lightgreen}21 & \cellcolor{lightyellow} 311 & \cellcolor{lightgreen}Indoor/outdoor scenes \\ \hline 
        \end{tabular}
\caption{\textbf{Summary of relevant existing spectro-polarimetric image datasets.}}
    \label{tab:dataset_comparison}
    \end{table*}

\begin{itemize}

\item Lei et al. \cite{lei2022wild} (first row) offers grayscale linear polarization data confined to outdoor scenes.

\item The work of \cite{ba2020deep} presents a dataset analogous to that of \cite{lei2022wild}, both designed for shape-from-polarization (SfP) studies. The distinctive aspect of \cite{ba2020deep} is its focus on indoor objects, contrasting with the broader scene diversity of \cite{lei2022wild}.

\item \cite{lapray2018data} provides linear polarization data across 6 spectral bands, though with a limited range of captured objects.

\item The study by \cite{qiu2021demos} introduces a novel demosaicing technique for linear polarization data, showcasing a limited set of indoor object examples.

\item \cite{dave2022pandora} offers two varieties of linear polarization trichromatic data, including a few multiview datasets of indoor and synthetic scenes.

\item \cite{gao2022ppp} provides indoor scenes with vast number of views for trichromatic linear polarization state.

\item \cite{lei2020refl} supplies linear polarization trichromatic data with a focus on reflected objects, encompassing a collection of under 1000 instances.

\item \cite{liang2022material} provides outdoor linear polarization trichromatic data, supplemented with NIR and LiDAR information.

\item \cite{kondo2020accurate} produces synthetic linear polarization trichromatic data for the purpose of polarization scene rendering.

\item Both \cite{lyu2019refl} and \cite{mei2022glass} generate linear polarization trichromatic data with a considerable number of examples, though specifically limited to reflective and transparent scenes, respectively.

\item \cite{kurita2023sensor} delivers a diverse indoor and outdoor scenes, but only in linear polarization across three wavelength channels.

\item The work by \cite{fan2023fullst} includes both linear and circular polarization states over 18 spectral bands. However, the dataset primarily comprises flat objects captured in a laboratory environment, with the majority being refrigerator magnets, and the remainder consisting of 3D-printed and glued objects.

\item \cite{kim2023neural} offers two types of datasets, each featuring linear and circular polarization data across 21 spectral channels. or each dataset type, they provide four examples: one comprising synthetic multiview data and the other containing multiview datasets of indoor and outdoor scenes.

\end{itemize}

Our dataset comprises two distinct types: one with extensive trichromatic Stokes data and the other with hyperspectral Stokes data encompassing 21 spectral bands, ranging from 450\,nm to 650\,nm at 10\,nm increments. Both datasets provide linear and circular polarization information for a variety of indoor and outdoor natural scenes.


\section{Acquisition Setup}

\begin{figure*}[h]
    \centering
    \includegraphics[width=\linewidth]{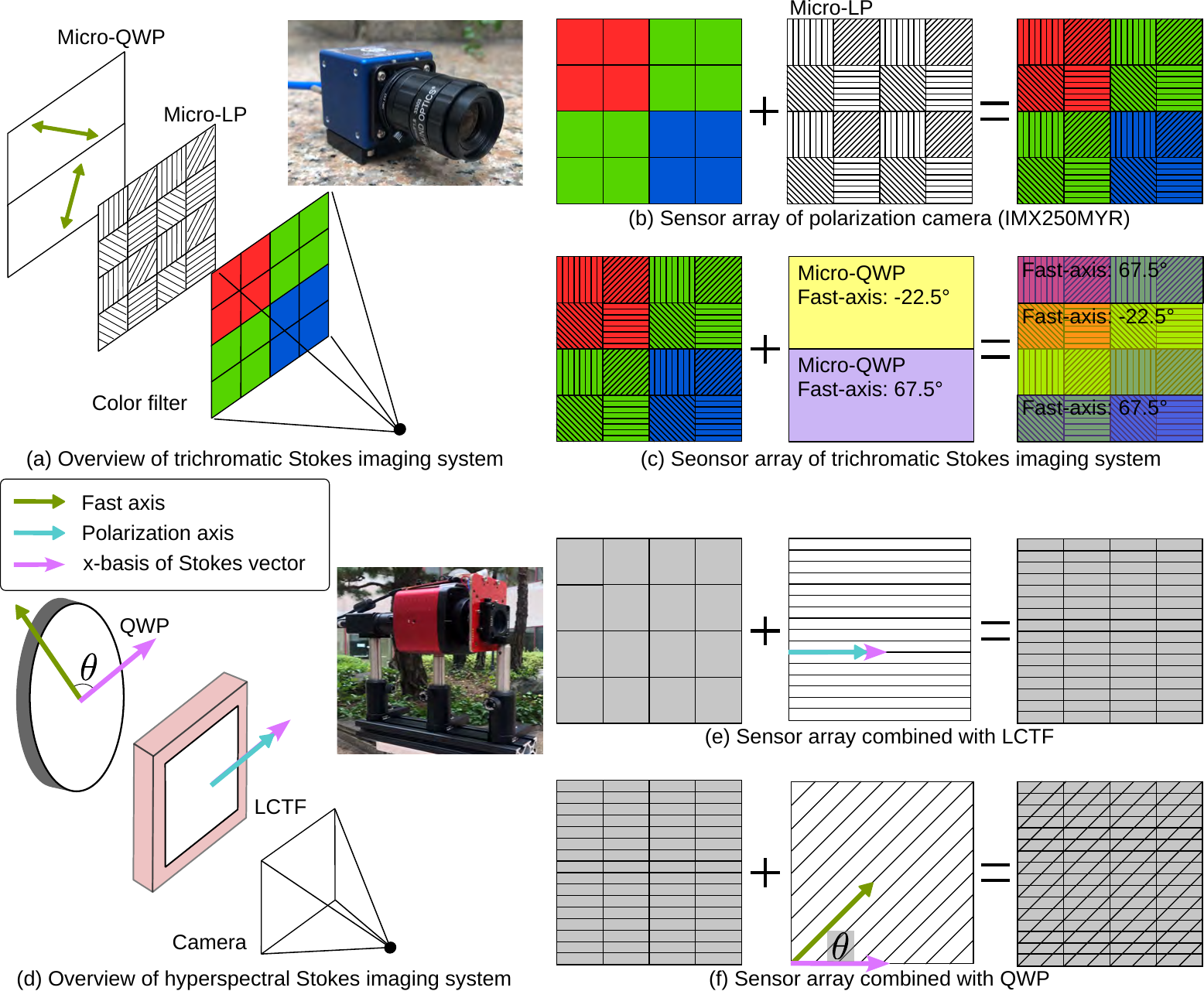}
    \caption{\textbf{Stokes Imaging Acquisition System.} (a) shows overview of acquisition system of trichromatic Stokes dataset. (b) illustrates structure of sensor array, SONY IMX250MYR, used widely in general polarization cameras. (c) represents the polarization sensor combined with micro-retarders which functions like quarter wave plate (QWP). (d) shows overview of hyperspectral Stokes imaging system. (e) and (f) illustrate sensor array of the hyperspectral system, equipped with LCTF and QWP.}
    \label{fig:setup}
\end{figure*}

To acquire full Stokes parameters, including both linear and circular polarization states, commercial polarization cameras typically require modifications, as they are originally designed to capture only linear polarization states. 
In the trichromatic Stokes imaging system developed by Tu et al. \cite{tu2017division}, as depicted in Figure~\ref{fig:setup}(a) to (c), the foundational element is a standard polarization camera. This camera is fitted with a Bayer color filter array, over which four types of wire-grid linear polarizers are placed, corresponding to the color filters. This arrangement facilitates the capture of linear polarization characteristics. In addition, micro-retarders with a retardance of 45$\,^{\circ}$ and two distinct fast axes are alternately attached to the system. These micro-retarders behave like quarter wave plate (QWP) converting the incoming light into circularly polarized light. Consequently, for each pixel, the system captures 4 $\times$ 4 RAW intensities, equating to four polarized intensities for every set of four Bayer color filter pixels.

The hyperspectral Stokes imaging system designed by Kim et al. \cite{kim2023neural} utilizes monochromatic sensor to capture grayscale intensity of light from various spectral bands. The light sequentially passes through a QWP oriented at one of four distinct angles, $\theta \in \{30\,^{\circ}, -45\,^{\circ}, 60\,^{\circ}, -90\,^{\circ}\}$, during separate exposures. Subsequently, it traverses an LCTF, which functions as a linear polarizer and selectively capture light at desired wavelengths, before reaching the camera sensor.

\section{Spectro-polarimetric Image Formation}

After acquiring the RAW intensity data of a scene, several post-processing steps are undertaken to accurately reconstruct the Stokes parameters. For the trichromatic camera, the captured scene is initially in the form of RAW data with a resolution of 2048 $\times$ 2448 pixels. This data is subdivided into 16 distinct segments. The subdivision process involves categorizing each pixel based on its row and column indices $(n, m)$. Here, pixels are systematically assigned to specific segments using the formula $(n \bmod 4) \times 4 + (m \bmod 4)$, effectively distributing them across these segments. Each of these segments possesses a resolution of 512 $\times$ 612 pixels. We can see 16 segments in the right-most of Figure~\ref{fig:setup}(b). Each partition is assigned to $K$-th segments, where $K=(n \bmod 4) \times 4 + (m \bmod 4)$.

We subsequently demosaic the 512 $\times$ 612 image into a 2048 $\times$ 2448 image, resulting in each spatial pixel of the 2048 $\times$ 2448 image having 16 intensity values. However, due to minor artifacts observed at the borders of the captured intensity, we crop the image to a resolution of 1900 $\times$ 2100 to ensure image quality. Subsequently, utilizing a pre-calibrated 16 $\times$ 12 reconstruction matrix, we derive four Stokes vectors for each of the three RGB components, total 12 values for each spatial pixel.

{For the hyperspectral Stokes vector, we acquire 84 measurements with 21 wavelength and four QWP angles for 512$\times$612 pixels. Using these captured images, we can reconstruct $\mathbf{s}(\lambda, p)$, the Stokes vector at a wavelength $\lambda$ for a pixel $p$, by solving the least-square problem with $\mathbf{M}(\theta_k, \lambda, p)$, a per-pixel Mueller matrix of the entire system including a QWP oriented at $\theta_k$:

\newcommand{\argmin}{\mathop{\mathrm{argmin}}}

\begin{equation}
\begin{aligned}
\argmin_{\mathbf{s}(\lambda,p)} \, \sum_{k=1}^{4} \, ({I^{\lambda,\theta_k}_\mathrm{meas}(p) - \mathbf{M}(\theta_k, \lambda,p;\lambda)\mathbf{s}(\lambda,p)})^2,
\end{aligned}
\label{eq:hyp_form2}
\end{equation}
where $I^{\lambda,\theta_k}_{meas}(p)$ is an intensity measurement for the pixel $p$ at a wavelength $\lambda$ with QWP of fast-axis angle $\theta_k$.
}


\newpage
\section{Noise and Denoising}
\begin{figure*}
    \centering
    \includegraphics[width=\linewidth]{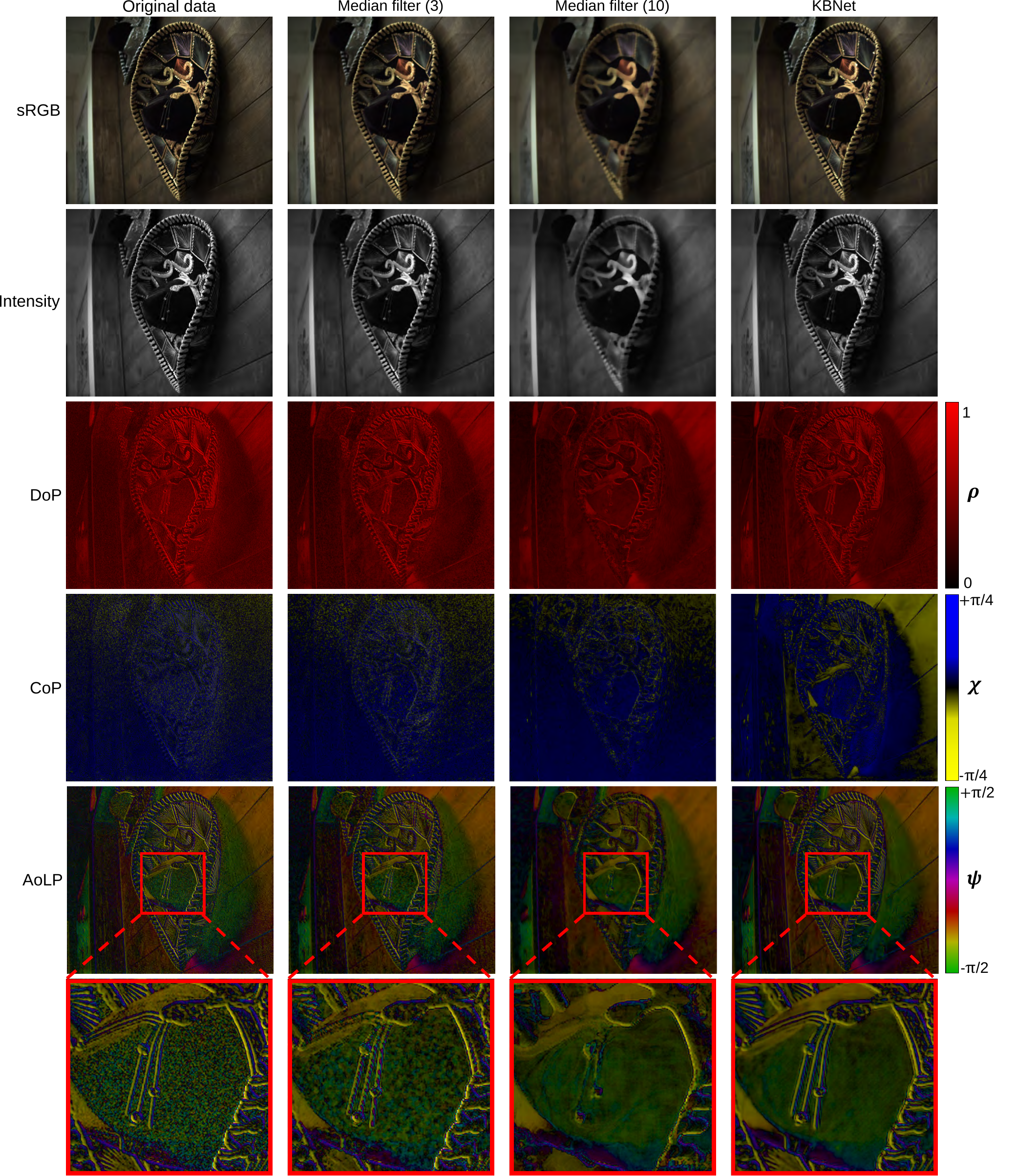}
        \caption{\textbf{Polarization Features of Denoised Stokes Vectors.} The first column shows noisy input data. The second and the third column represent results after applying median filter of size 3$\times$3 and 10$\times$10 respectively. The last column illustrates result after employing single-shot denoising method KBNet \cite{zhang2023kbnet}. Each row presents the polarization characteristics of the red channel from a trichromatic Stokes image, encompassing sRGB, intensity, DoP, CoP, and AoLP. The last row emphasizes a severely noisy portion of AoLP visualization of the original Stokes vector and that of denoised Stokes vectors.}
    \label{fig:denoising_compare}
\end{figure*}

Images from the full-stokes camera can contain severe noise, especially for low intensity scenes. To suppress noise, we applied several approaches such as median filter and deep learning-based denoising network. 
We first employed a median filter to captured raw intensities, which is widely recognized for its efficacy in reducing noise while preserving edges in images. By moving through pixels in an image, it replaces with the median value of neighboring pixels. It reduces more noise compared to original data while losing details as the size of the median filter increases, as shown in Figure~\ref{fig:denoising_compare}. 

Similarly, we applied the state-of-the-art single-shot denoising method, KBNet \cite{zhang2023kbnet}, on the captured RAW intensities. The outcomes, including the visualized polarization characteristics, are illustrated in the last column of the Figure~\ref{fig:denoising_compare}. Comparing to the median filters, KBNet preserves internal details while suppressing severe noise.

\clearpage

\section{Spatio-spectral-polarimetric Reprsentation Methods}
\subsection{PCA}
We conducted principal component analysis (PCA) on both trichromatic and hyperspectral Stokes dataset. We first divided the images into patches of size $P\times P$ patches, $P=10$ in this time, resulting in 190 $\times$ 210 patches for one trichromatic Stokes image and 51 $\times$ 61 patches for one hyperspectral Stokes image, to efficiently extract a spatial basis.

After splitting, we flatten the patches and transform into 200 coefficients.
We then derive the basis vectors for these Stokes patches:
\begin{equation}
\mathbf{p}=\mathbf{c} \cdot \mathbf{b} + \mu,
\end{equation}
with $\mathbf{b}$ as the basis, $\mathbf{c}$ the corresponding coefficient, and $\mu$ the mean of incoming data.
To store 10 $\times$ 10 size of Stokes parameters, we need 10 $\times$ 10 $\times$ $t$ $\times$ $l$ parameters, where $t$ denotes the number of Stokes parameters for one pixel, four, and $l$ represents the number of spectral channels, which is 21 for the hyperspectral Stokes dataset and 3 for the trichromatic Stokes dataset. However, to represent with coefficients and bases, we need only 93 parameters for coefficients and one additional parameter for $\mu$ to achieve bits-per-pixel (BPP) 60 with reasonable results.
Figure \ref{fig:pca_hyp} illustrates the obtained bases of hyperspectral dataset. Each row matches the wavelength across spectral axis, and we can see that bases for each wavelength exhibit similar spatial characteristics with variations in scale.

\subsection{INR Network Architecture}
\begin{figure}[h]
    \centering
    \includegraphics[width=0.7\linewidth]{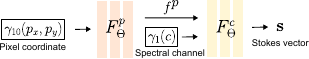}
    \caption{\textbf{Network architecture.}}
    \label{fig:network_architecture}
\end{figure}

\begin{figure*}
    \centering
    \includegraphics[width=\linewidth]{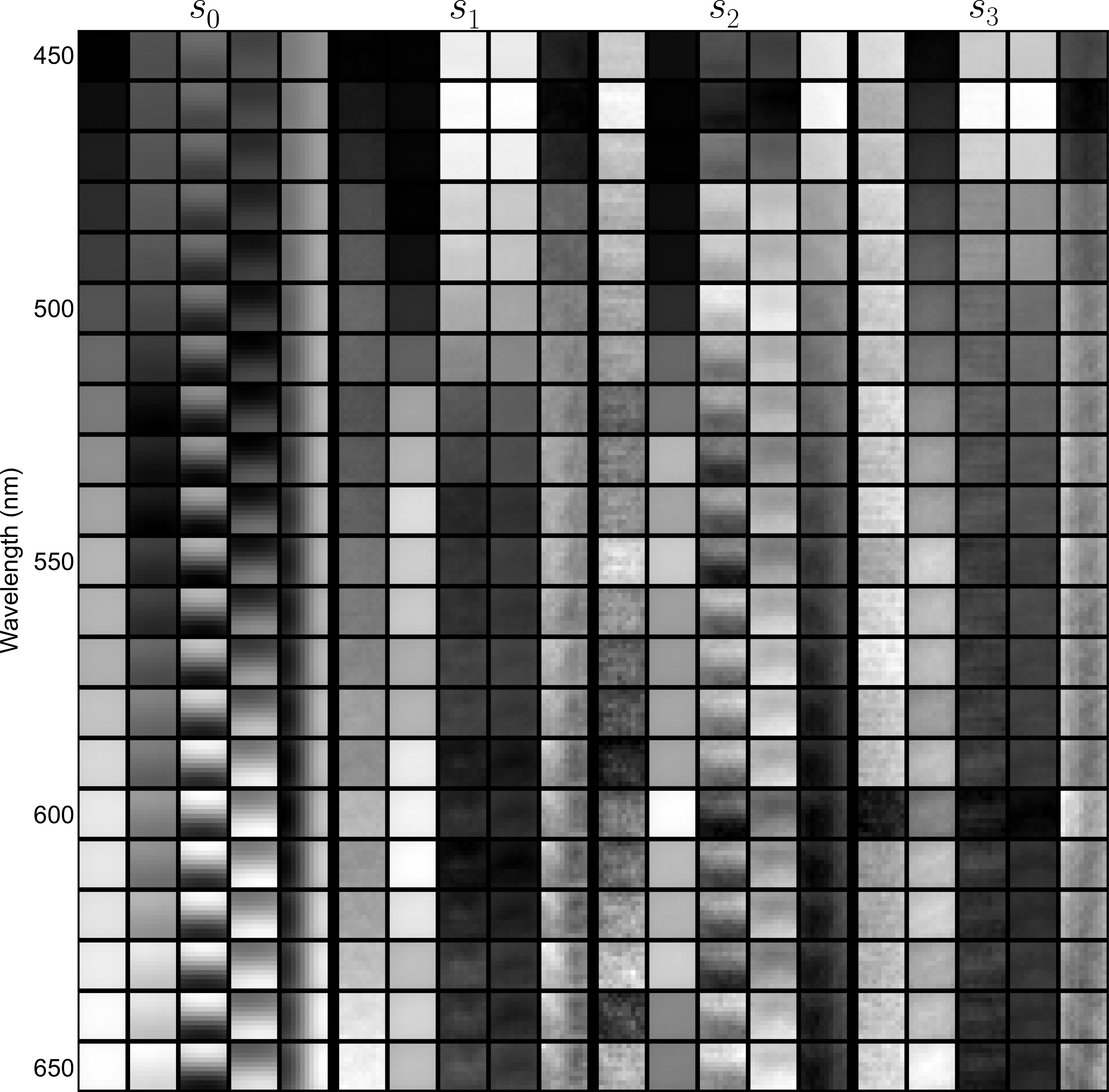}
    \caption{\textbf{Bases obtained via PCA.} each row corresponds to a specific wavelength along the spectrum axis starting from 450\,nm and increasing in increments of 10\,nm, culminating at 650\,nm. The columns are broadly categorized into $s_0$ to $s_3$, and within each group, there are five columns that represent five most significant bases.}
    \label{fig:pca_hyp}
\end{figure*}

We modify the NeSpoF~\cite{kim2023neural} to efficiently represent spectro-polarimetric information of a particular scene. We model the per-pixel MLP ($F_\Theta^p$) and spectral MLP ($F_\Theta^c$) without using intermediate polarimetric features, $s_0$, DoP, $\chi, \psi$,  as shown in Figure~\ref{fig:network_architecture}.
The per-pixel MLP extracts the per-pixel polarimetric feature $f^p$ and the spectral MLP outputs the Stokes vectors from the per-pixel feature $f^p$ and spectral channel $c$.
We set the hyperparameter $k$ of positional encoding~\cite{tancik2020fourier} to 10 and 1 for the pixel coordinate $(p_x, p_y)$ and spectral channel $c$, respectively:
\begin{equation}
\gamma_k(x) = [x, \sin(\omega_0 x), \cos(\omega_0 x), ..., \sin(\omega_k x), \cos(\omega_k x)],
\end{equation}
where $\omega_k = 2^k\pi$. To achieve BPP 60, we set the number of layers 8, which consumes 2.22\,MB to store 100\,MB Stokes data.

\subsection{Additional Results}
\begin{figure*}
    \centering
    \includegraphics[width=\linewidth]{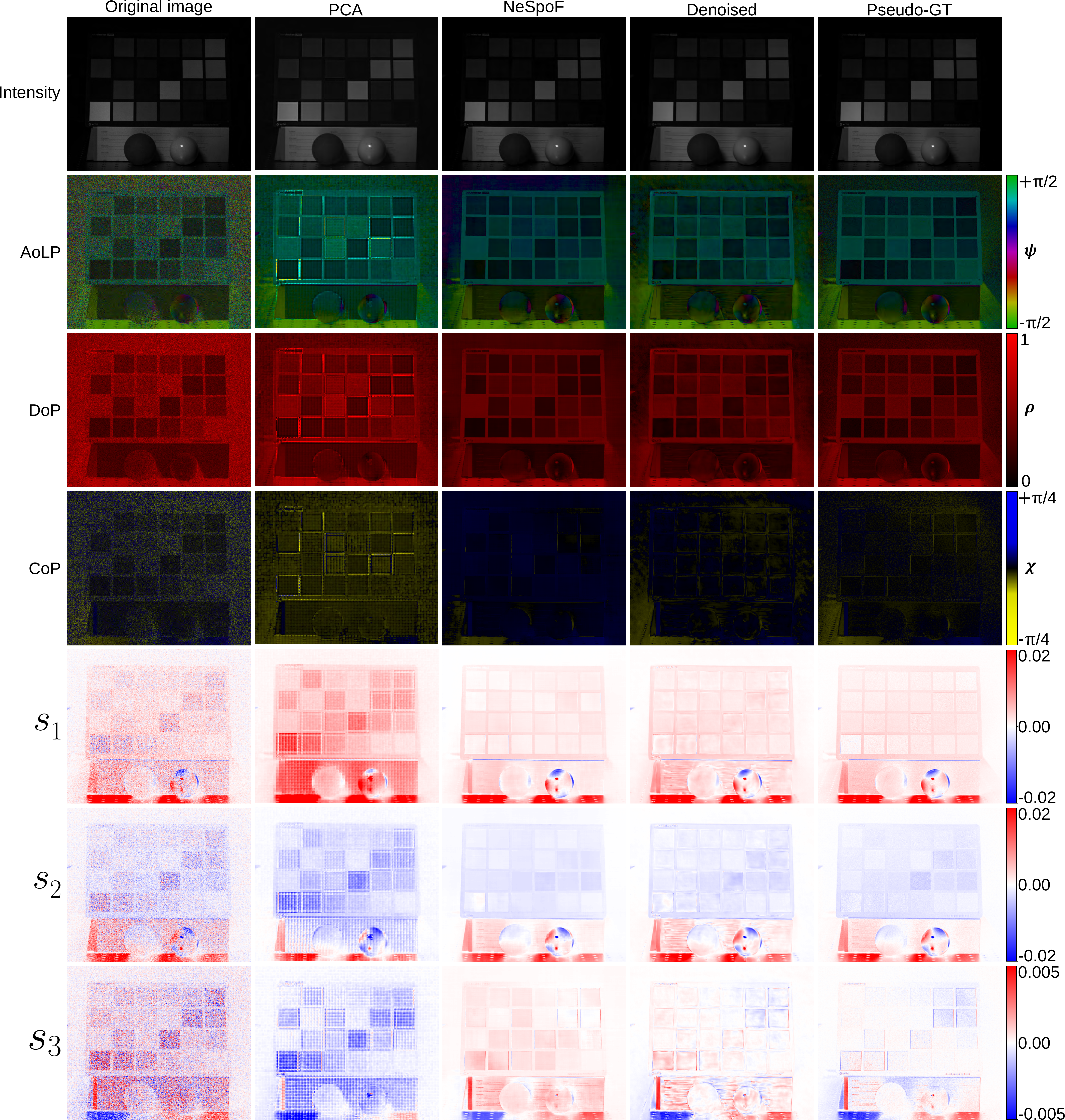}
    \caption{\textbf{Results after Compressive Polarization Representation.} Each column shows original Stokes image, Stokes vectors reconstructed from bases and coefficients by PCA, results after NeSpoF, Stokes vectors reconstructed from denoising methods by \cite{zhang2023kbnet} and pseudo ground truth Stokes parameters obtained by burst-imaging 100 images. Each row illustrates Stokes vectors and polarization characteristics of 550\,nm, encompassing sRGB, intensity and AoLP, DoP, CoP visualization, and $s_1$, $s_2$, $s_3$ information.}
    \label{fig:nespof}
\end{figure*}

Above, we show that employing PCA and INR can reduce the memory footprint of data. Consequently, this approach also facilitates effective denoising of the data. As shown in Figure 4(d) and (f) of the main text, a comparison with pseudo ground-truth(pseudo-GT) Stokes vectors, generated through burst imaging of a hundred images, reveals that the PCA method yields a mean squared error (MSE) of 2.69$\times 10^{-5}$, while the INR approach exhibits an MSE of 1.90$\times 10^{-5}$. Figure~\ref{fig:nespof} presents visualizations of various polarization characteristics and the components $s_1$, $s_2$, and $s_3$ of the Stokes vectors of the original scene. It also showcases the outcomes produced by PCA, NeSpoF, the denoising method, and burst imaging. The denoised Stokes vectors exhibit the lowest error, however, unlike PCA and INR, they do not facilitate data size compaction.

\clearpage

\section{Polarized and Unpolarized Intensity}
This section presents additional results from the decomposition of our datasets into polarized and unpolarized light. Figure~\ref{fig:suppl_polunpol} shows the separated polarized and unpolarized images of 18 different scenes in our trichromatic dataset, while  Figure~\ref{fig:suppl_polunpol2}(a), (b) and (c) shows the same results across the spectrum for one scene in our hyperspectral dataset. As discussed in the main paper, it is evident from our observations that polarized images contain specular reflections, such as the glow of glass. Notably, outdoor scenes captured under sunlight exhibit a rainbow-colored polarization of the sky. Additionally, we can confirm that the LCD display on a monitor mainly emits polarized light, as shown in Figure~\ref{fig:suppl_polunpol}.

\begin{figure}[h]
    \centering
    \includegraphics[width=\linewidth]{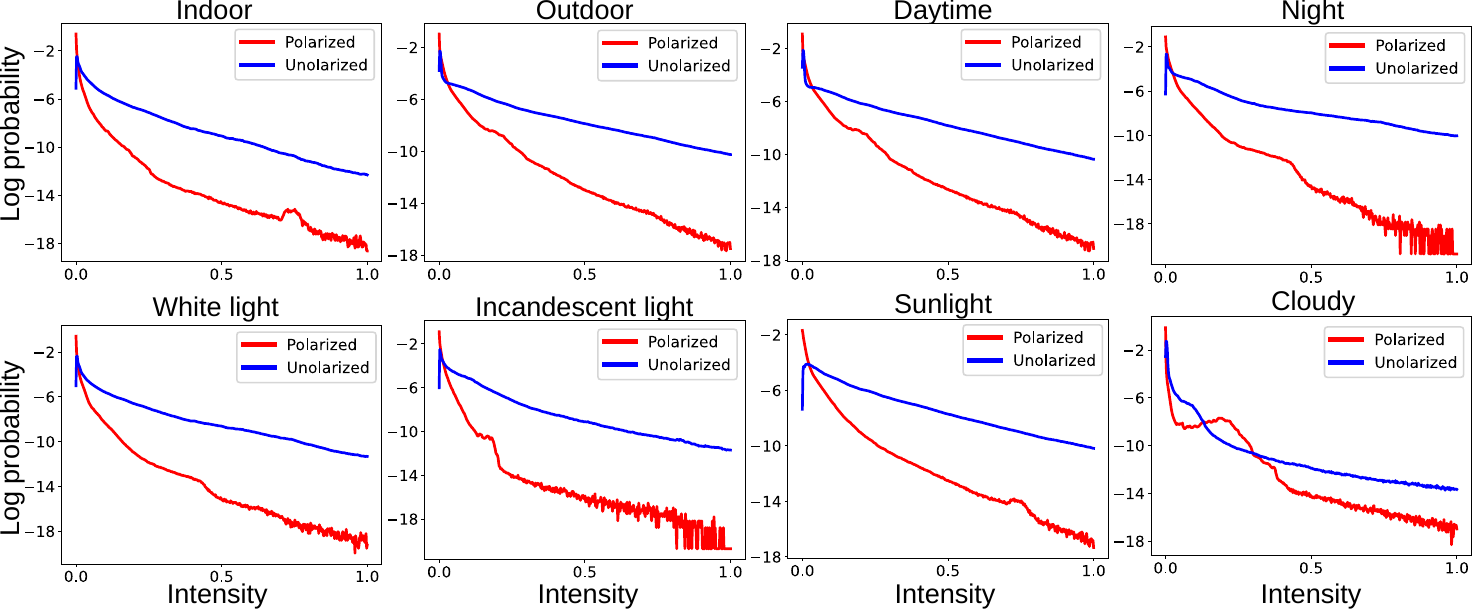}
    \caption{\textbf{Intensity distributions for polarized and unpolarized components with respect to dataset labels.}  }
    \label{fig:suppl_polunpol3}
\end{figure}

\begin{figure*}
    \centering
    \includegraphics[width=0.95\linewidth]{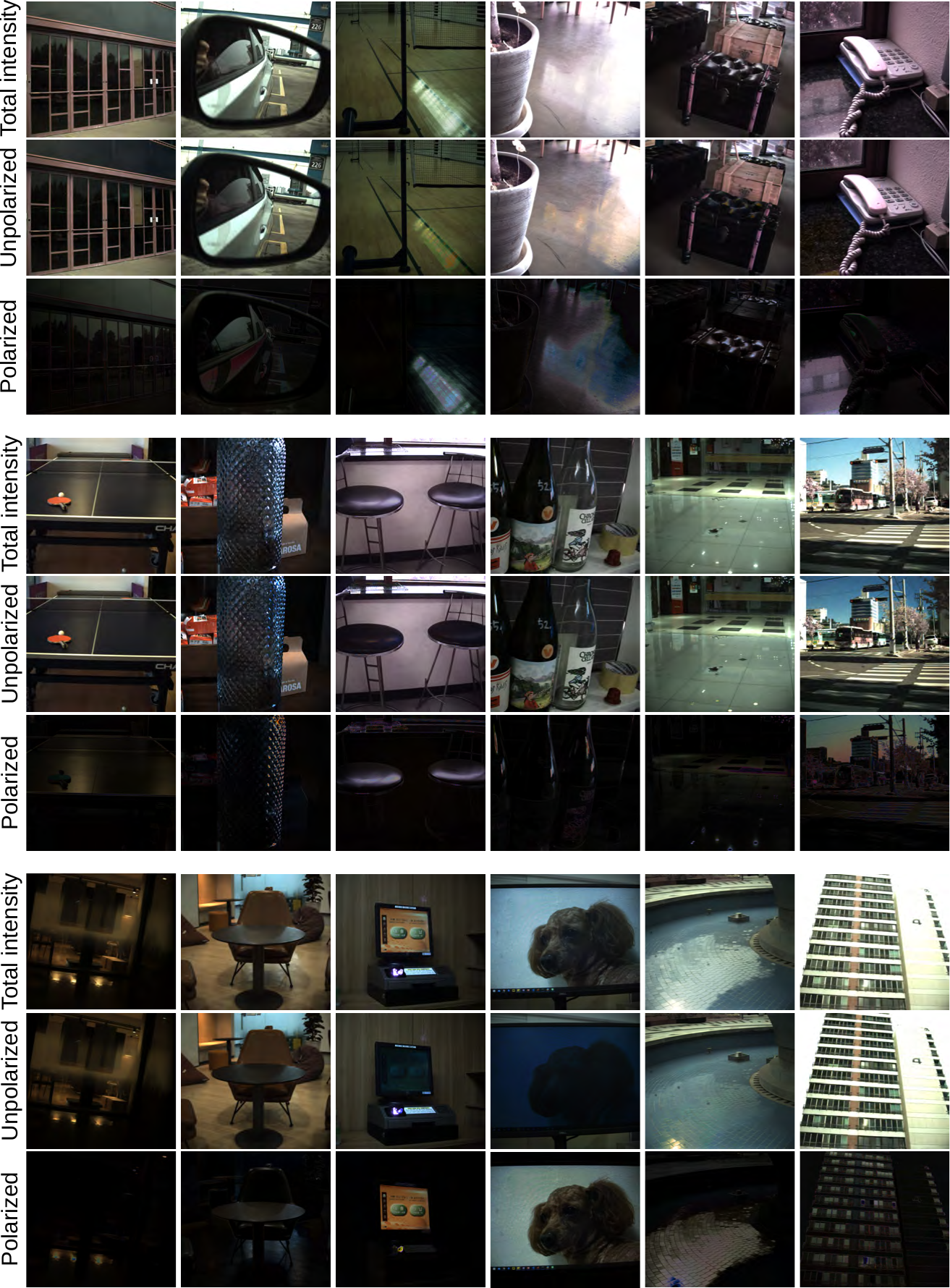}
    \caption{\textbf{Additional results of separation into polarized and unpolarized light in the Trichromatic dataset.} }
    \label{fig:suppl_polunpol}
\end{figure*}

\begin{figure*}
    \centering
    \includegraphics[height=0.95\textheight]{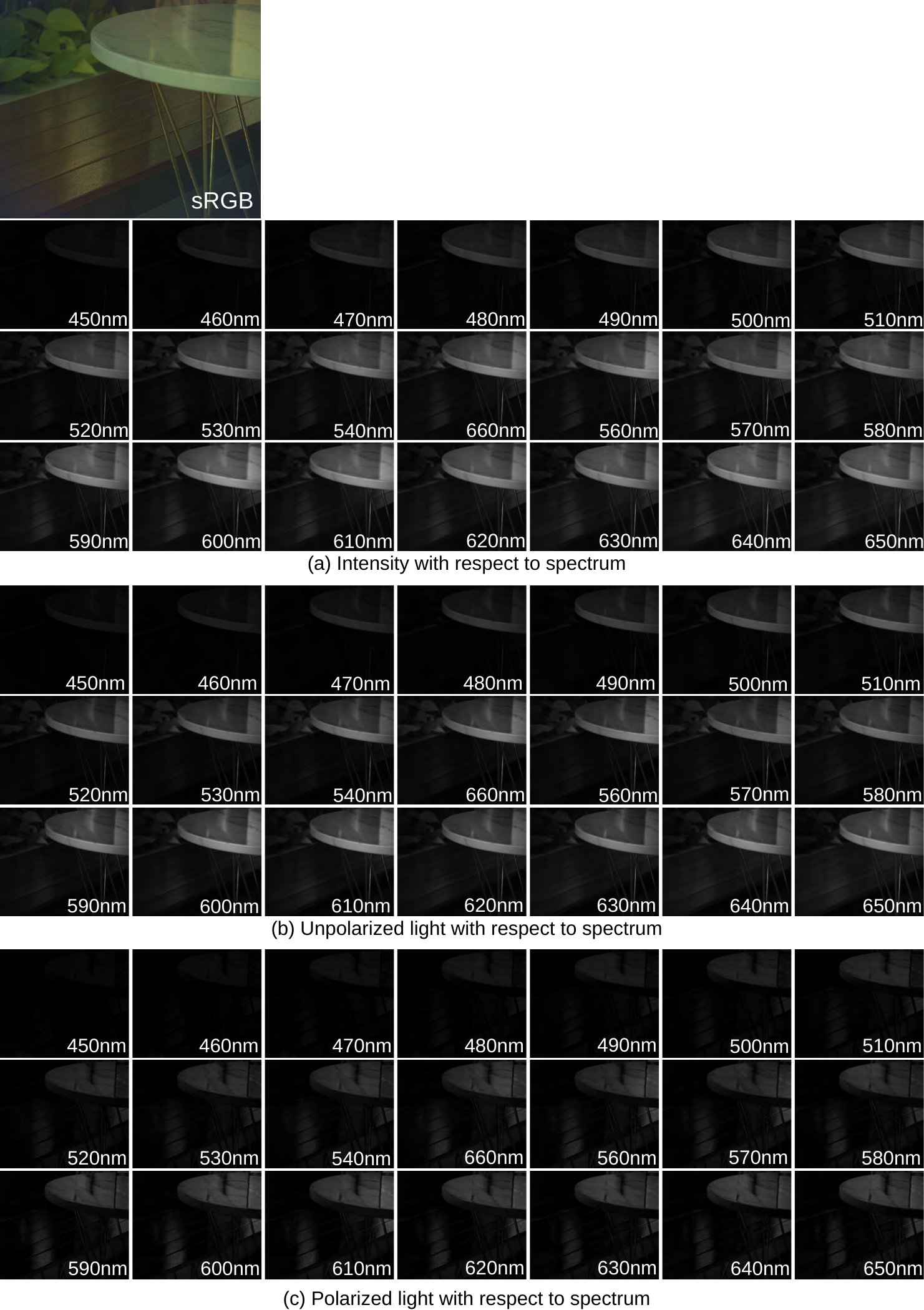}
    \caption{\textbf{Additional results of separation into polarized and unpolarized light in Hyperspectral dataset} Intensities and unpolarized lights are in range 0 to 1, and polarized lights are plotted in range 0 to 0.3 for visualization.}
    \label{fig:suppl_polunpol2}
\end{figure*}

Furthermore, we conducted an analysis of the intensity distributions for polarized and unpolarized components concerning the dataset labels. Figure~\ref{fig:suppl_polunpol3} shows that, irrespective of the label, the intensity of polarized light skews towards low and high-intensity values compared to unpolarized light. However, distinct patterns emerge when comparing intensity distributions captured under sunlight and cloudy conditions. In the cloudy dataset, polarized light is more concentrated near zero compared to the sunlight dataset, and a peak is observed in the middle of the graph. As polarized sunlight passes through clouds, it undergoes scattering events, leading to changes in its polarization states. The distinct polarization properties arising from varying illumination conditions will be further discussed in the subsequent analysis sections.

\clearpage

\section{Stokes-vector Distributions}

\begin{figure*}[t]
	\centering
		\includegraphics[height=\textheight]{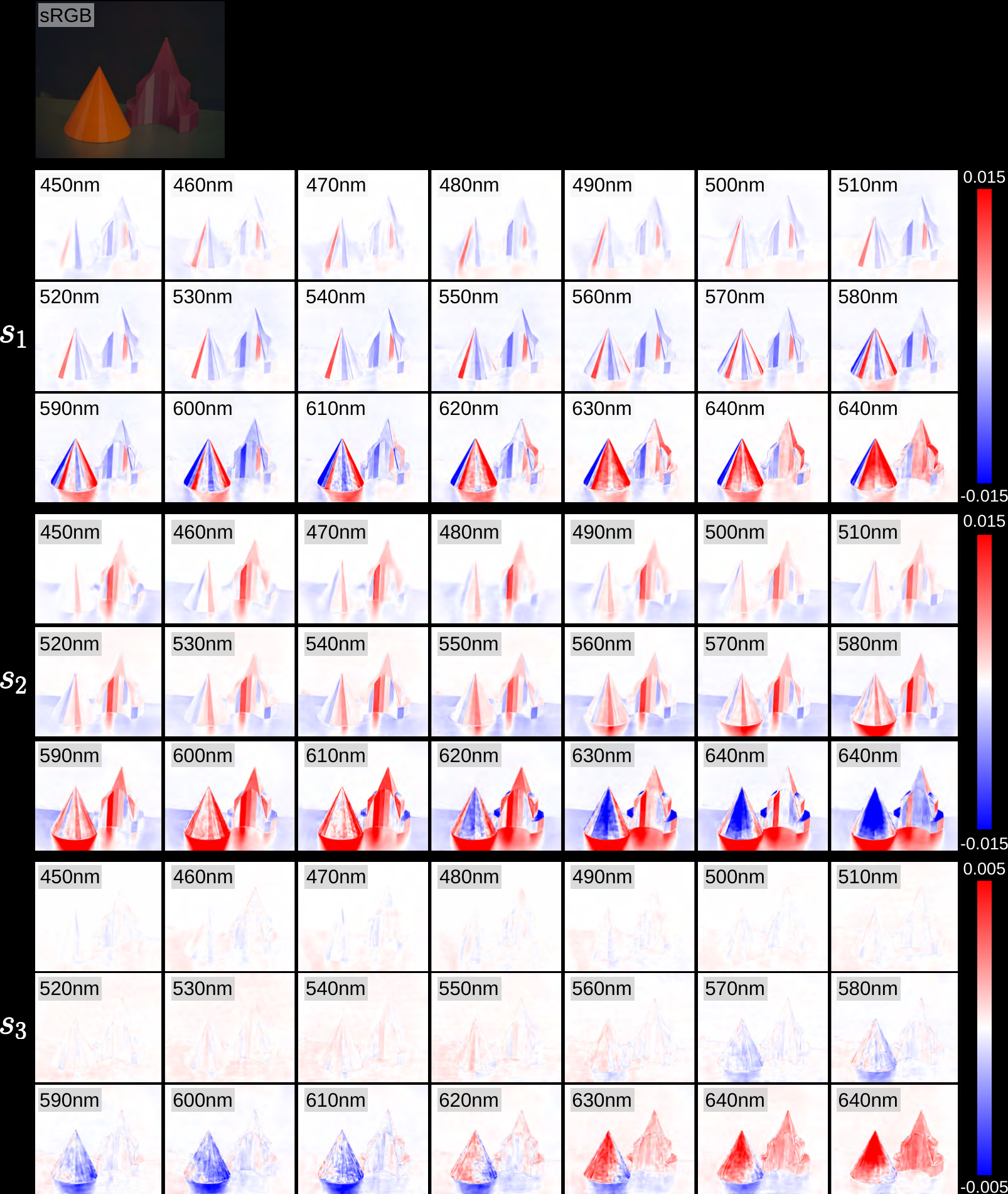}
		\caption{\textbf{Stokes Images over the Spectrum}. Stokes images of $s_1$, $s_2$ and $s_3$ of hyperspectral dataset}
		\label{fig:hyp_stokes_image}
\end{figure*}

\begin{figure*}[t]
	\centering
		\includegraphics[height=\textheight]{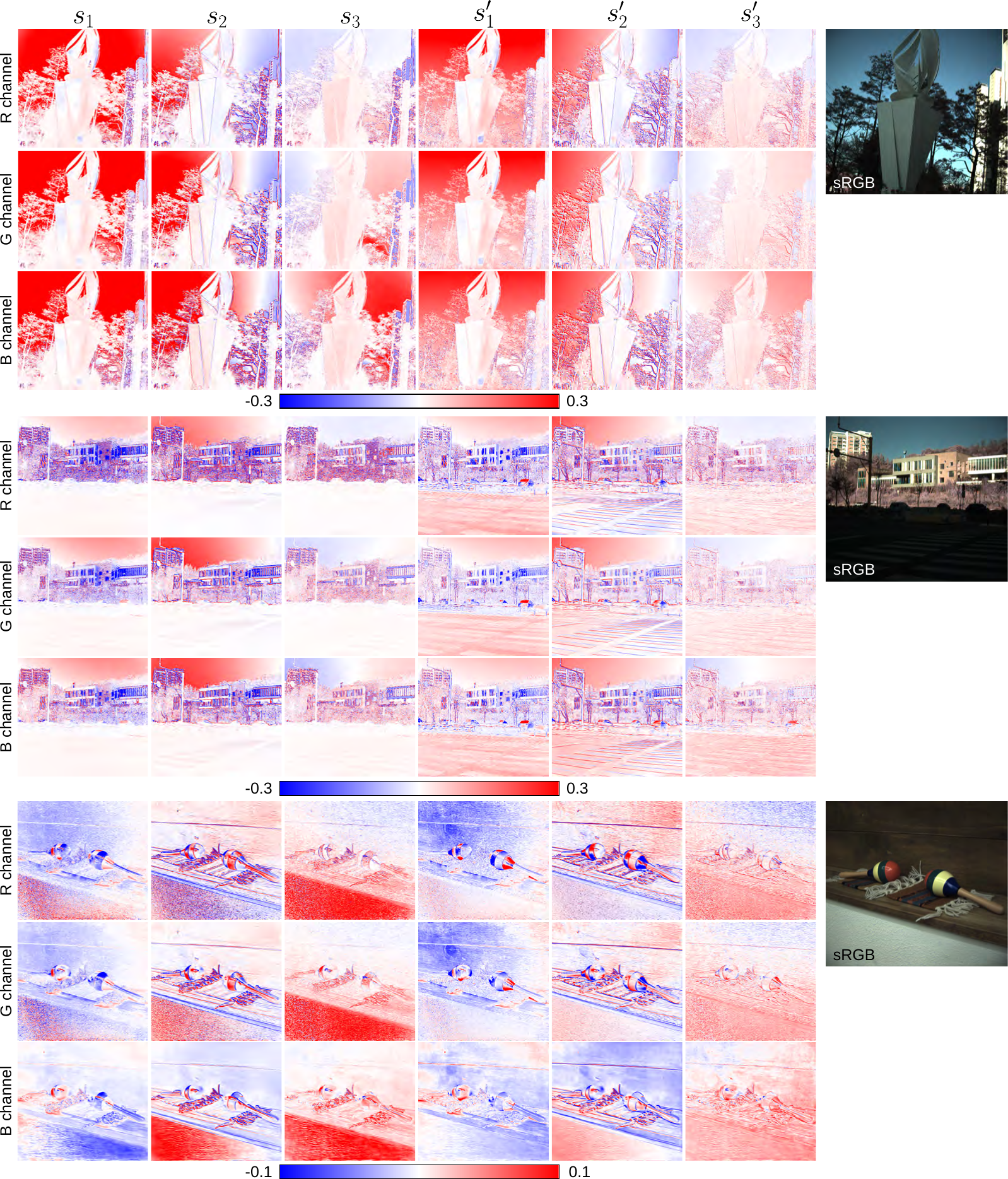}
		\caption{\textbf{Stokes images at R, G and B channels}. Stokes images of $s_1$, $s_2$ and $s_3$ of trichromatic dataset}
		\label{fig:rgb_stokes_image}
\end{figure*}

\begin{figure*}[t]
	\centering
		\includegraphics[width=\linewidth]{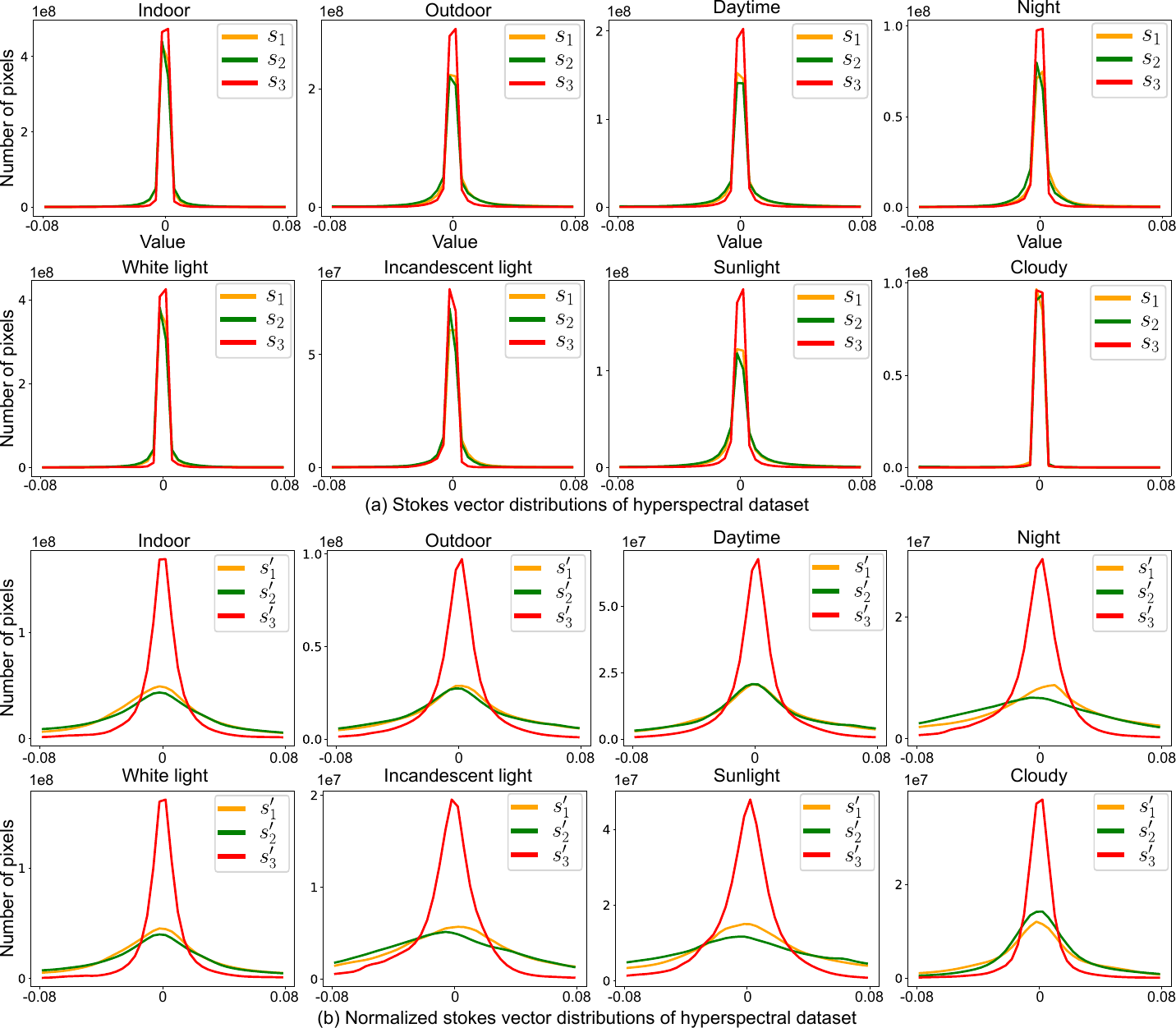}
		\caption{\textbf{Stokes-vector and normalized Stokes-vector distributions of hyperspectral dataset}}
		\label{fig:hyp_stokes_graph}
\end{figure*}

\begin{figure*}[t]
	\centering
		\includegraphics[width=\linewidth]{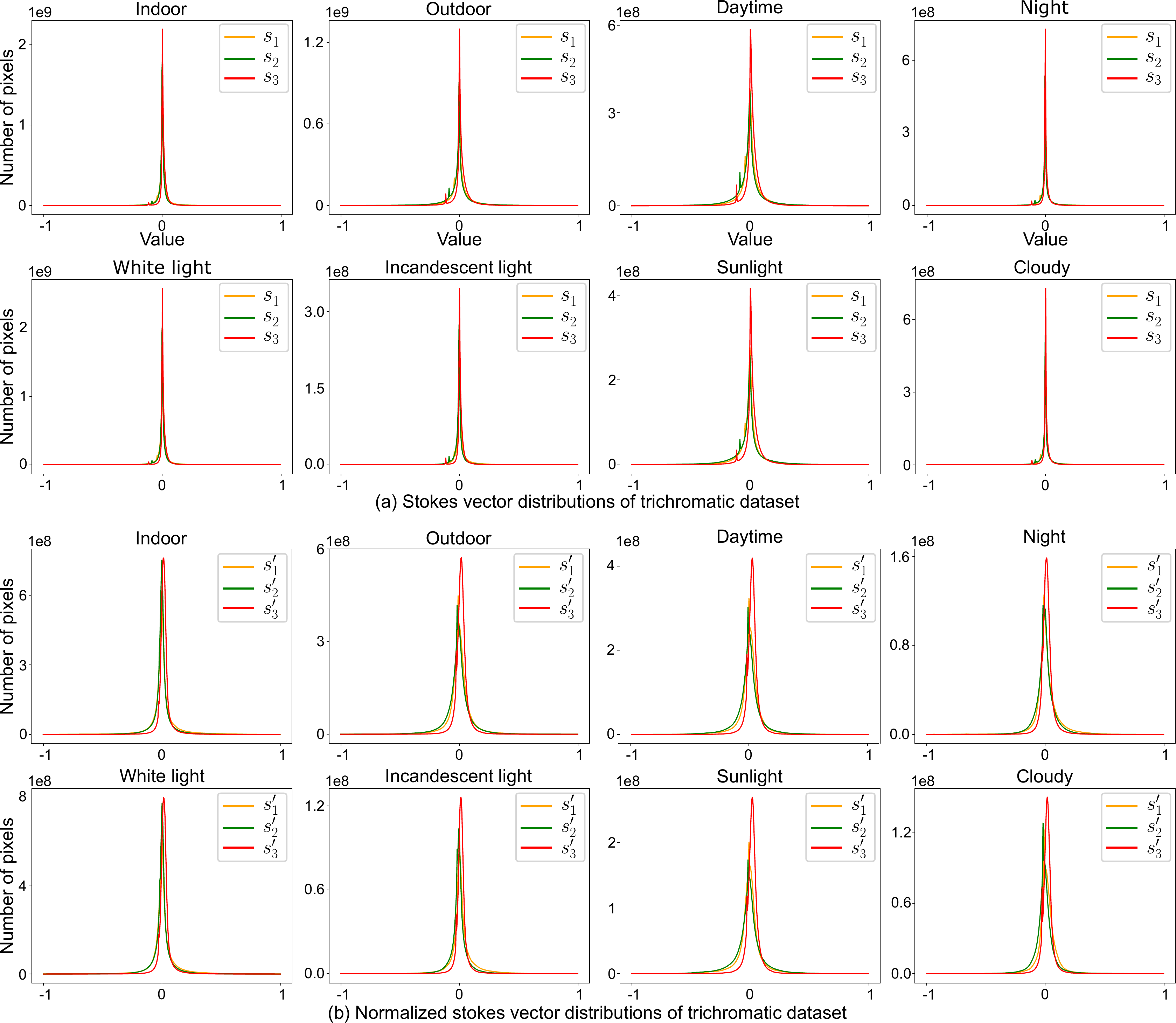}
		\caption{\textbf{Stokes-vector and normalized Stokes-vector distributions of trichromatic dataset}}
		\label{fig:rgb_stokes_graph}
\end{figure*}

We analyze the distributions of Stokes-vectors concerning the spectrum and various capture environments based on our dataset labels.

Figure~\ref{fig:hyp_stokes_image} highlights the diversity of the Stokes vectors across the spectrum. It visualizes the Stokes vector components $s_1$, $s_2$ and $s_3$ for a scene in the hyperspectral dataset. As shown in Figure 8, the values of $s_1$, $s_2$ and $s_3$ on the surface of the cone shows distinct distributions for each wavelength. Particularly, noteworthy is the significant difference in $s_2$ and $s_3$ values at 610nm and 620nm, despite only a 10nm separation. The hyperspectral polarization dataset proves instrumental in providing detailed insights into the analysis and decomposition of polarization information, in contrast to trichromatic polarization data.

Figure~\ref{fig:rgb_stokes_image} highlights the impact of the capture environment on polarization states. The upper two scenes were captured outdoors, and the bottom scene was captured indoors. As shown in Figure~\ref{fig:rgb_stokes_image}, the intensity of Stokes vectors in outdoor scenes tends to be higher than in indoor scenes. Moreover, the outdoor scene captured under sunlight, which includes more circularly polarized light, exhibits greater variability in $s_3$ values compared to $s_1$ and $s_2$ across the R, G, and B channels. Conversely, for the indoor scene, $s_1$ and $s_2$ are widely distributed across color channels rather than $s_3$.

\begin{figure}[h]
    \centering
    \includegraphics[width=0.7\linewidth]{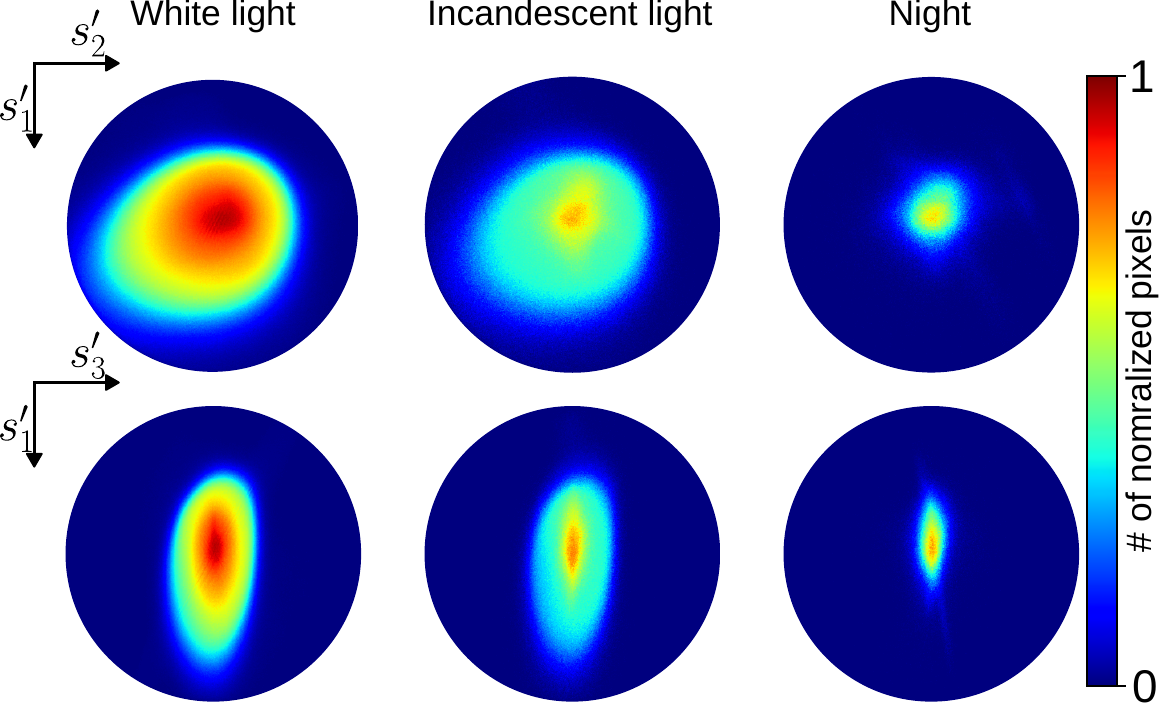}
    \caption{\textbf{Poincaré sphere 
  of white light, incandescent light and night scene}  }
    \label{fig:environment}
\end{figure}

Moreover, we conduct an analysis based on the dataset labels by plotting the distributions of Stokes vectors in both the hyperspectral and trichromatic datasets (Figure~\ref{fig:hyp_stokes_graph},~\ref{fig:rgb_stokes_graph}). We observe that the components of the Stokes vector in the cloudy scene are concentrated near zero compared to the sunlight scene. This observation aligns with the understanding that polarized sunlight undergoes depolarization due to scattering events on the cloud, a phenomenon further illustrated in the Poincaré sphere visualization presented in the main paper.
For the night scene, characterized by predominantly low-intensity pixels, Stokes vectors are distributed unstably and lack a symmetrical shape. Furthermore, the maximum number of pixels is significantly lower than that in the other scenes, indicating a dispersion of these values. Although the data captured under white light and incandescent light exhibit similar patterns in Stokes vector distributions, the maximum number of pixels in the incandescent light scene is lower than that in the white light scene. This indicates that scenes under the white light have a higher slope in its Stokes vector distributions compared to scenes under the incandescent light. This observation is further supported by the Poincaré sphere visualization (see Figure~\ref{fig:environment}).

\clearpage

\section{Gradient Analysis}
\begin{figure*}[h]
    \centering
    \includegraphics[width=\linewidth]{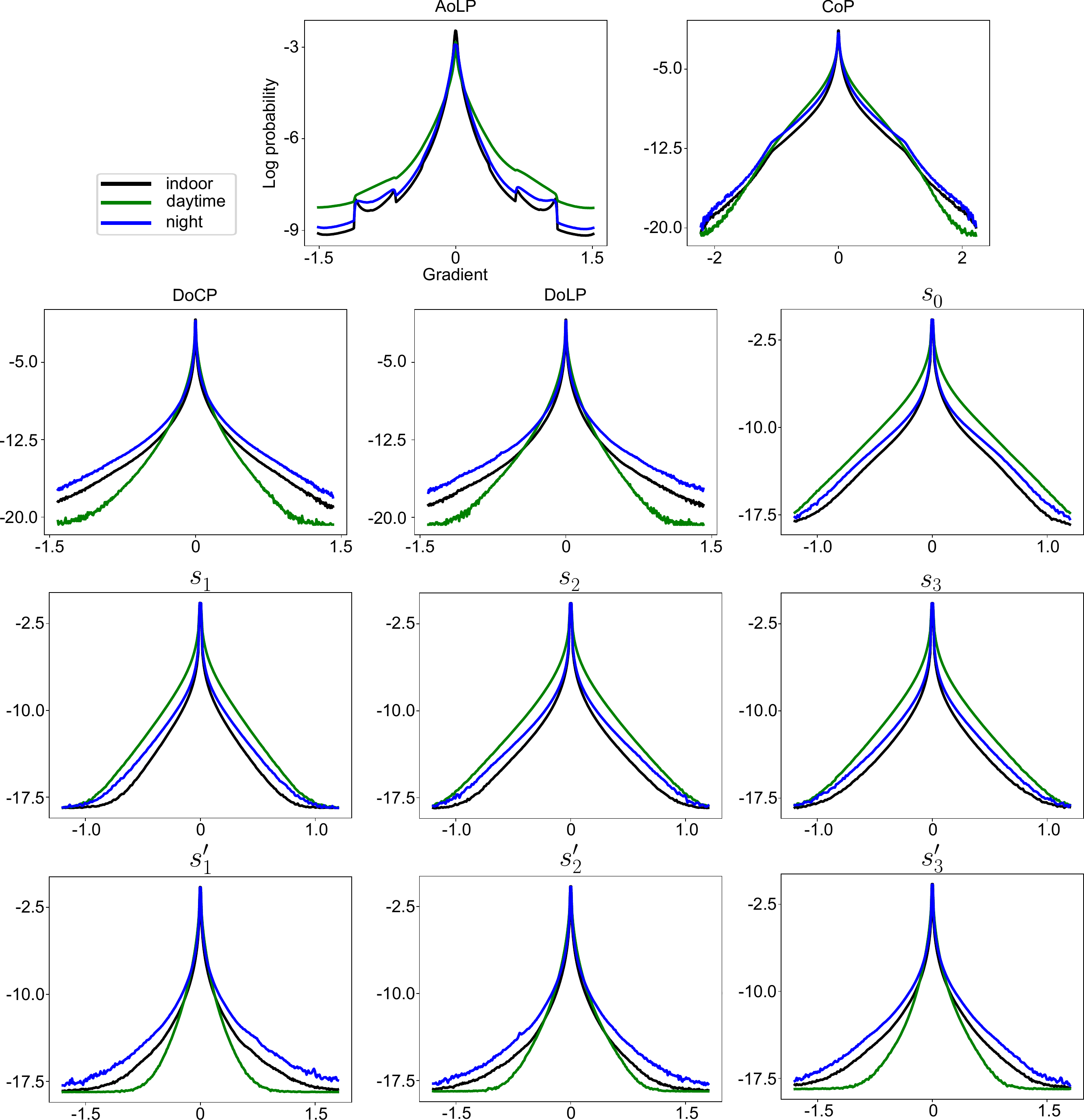}
    \caption{\textbf{Additional analysis of gradient distributions with respect to dataset labels.} }
    \label{fig:gradient}
\end{figure*}

In this section, we present additional gradient distributions based on the location and time of data capture (indoor, daytime, and night). As shown in Figure~\ref{fig:gradient}, we plot the gradient distributions of Stokes vectors ($s_0$, $s_1$, $s_2$, and $s_3$), normalized Stokes vectors ($s_1'$, $s_2'$, and $s_3'$), Angle of Linear Polarization (AoLP), Chirality of Polarization (CoP), Degree of Circular Polarization (DoCP), and Degree of Linear Polarization (DoLP). Scenes captured at different locations and times exhibit distinct patterns in their gradient distributions, while still retaining similar shapes to the hyper-Laplacian prior.
Interestingly, data captured during the daytime shows that the gradient of Stokes vectors ($s_0$, $s_1$, $s_2$, and $s_3$) is distributed more widely compared to indoor and night scenes. However, for the normalized Stokes vectors ,$s_1'$, $s_2'$, and $s_3'$, daytime scenes are more concentrated near zero than others.

\clearpage

\section{Shape from Polarization}
\begin{figure*}
    \centering
    \includegraphics[width=\linewidth]{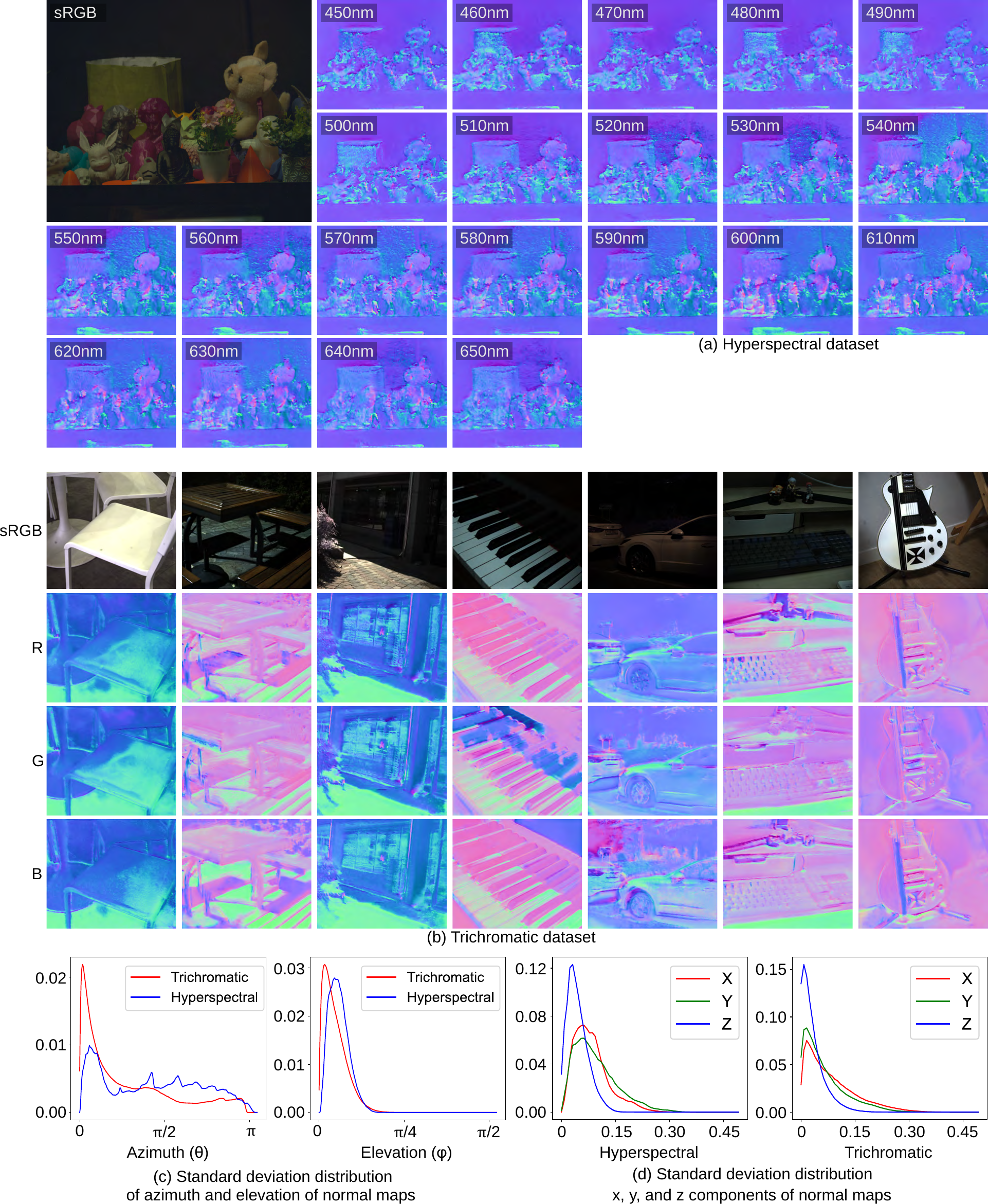}
    \caption{\textbf{Normal maps obtained via SfP by Lei at el. \cite{lei2022wild} and their statistics.} (a) Normal maps acquired from the same scene in hyperspectral Stokes dataset but with different wavelength. (b) Normal maps obtained from various illumination condition from the trichromatic Stokes dataset. (c) Standard deviation probability of azimuth and elevation of normal maps categorized by dataset. (d) Standard deviation probability of $x$, $y$, and $z$ components of normal maps.}
    \label{fig:sfp_hyp}
\end{figure*}

Through shape-from-polarization (SfP) methods, we can reconstruct 3D normal map from polarization data such as Stokes parameters, DoLP and AoLP information. However, as they utilize features from linear polarization state and monochromatic data, they lose consistency of normal maps along the spectral axis. 
Divergent normal maps recovered across different wavelengths using SfP by Lei at el. \cite{lei2022wild} from hyperspectral Stokes dataset are shown in Figure~\ref{fig:sfp_hyp}(a) and trichromatic Stokes dataset in Figure~\ref{fig:sfp_hyp}(b). In Figure~\ref{fig:sfp_hyp}(a), although Stokes parameters are shown to have different values for each spectrum, normal map of the scene should not be distinct, which is not accomplished by existing SfP methods. Those distinctions are observed regardless of the illumination condition and the number of spectral channels. Figure~\ref{fig:sfp_hyp}(b) shows various scenes from the trichromatic Stokes dataset, including sunlight, night time, white light, and yellow light. 
Figure~\ref{fig:sfp_hyp}(c) plots the standard deviation probability of azimuth $\theta$ and elevation $\phi$ while Figure~\ref{fig:sfp_hyp}(d) shows the standard deviation probability of $x$, $y$, and $z$ components of normal maps across the spectral bands for each dataset. We can see that $x$ and $y$ vary more than $z$ components leading to fluctuating azimuth, which indicates that reconstructed normal maps do not guarantee reliable $x$ and $y$ components. Utilizing circular polarization and hyperspectral information, SfP yields more accurate normal maps.

\clearpage

\section{Dataset Examples}
\begin{figure*}
    \centering
    \includegraphics[height=\textheight]{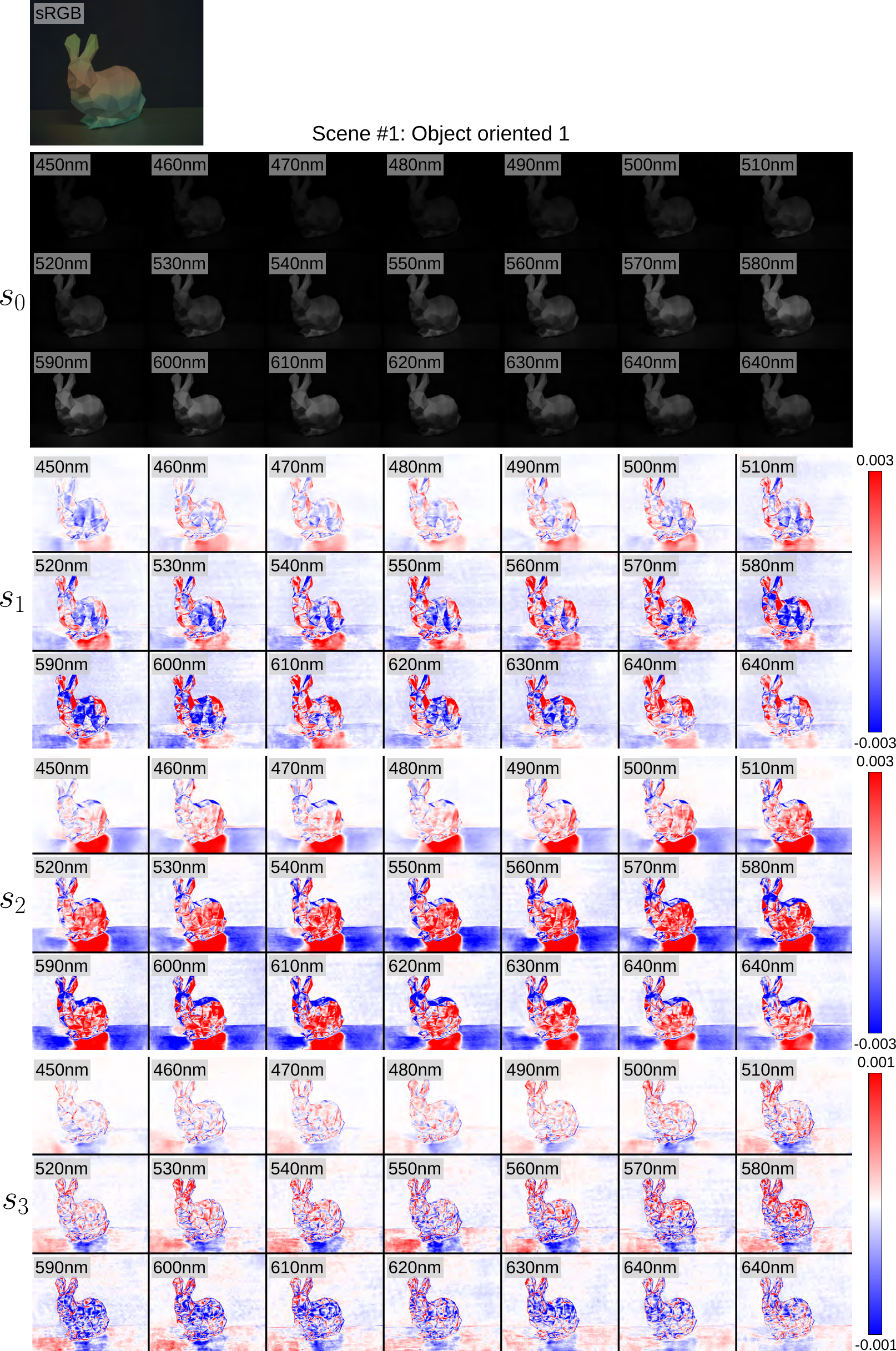}
    \caption{\textbf{Dataset example of object-centric data in the hyperspectral dataset.} }
    \label{fig:suppl_hyp_obj1}
\end{figure*}

\begin{figure*}
    \centering
    \includegraphics[height=\textheight]{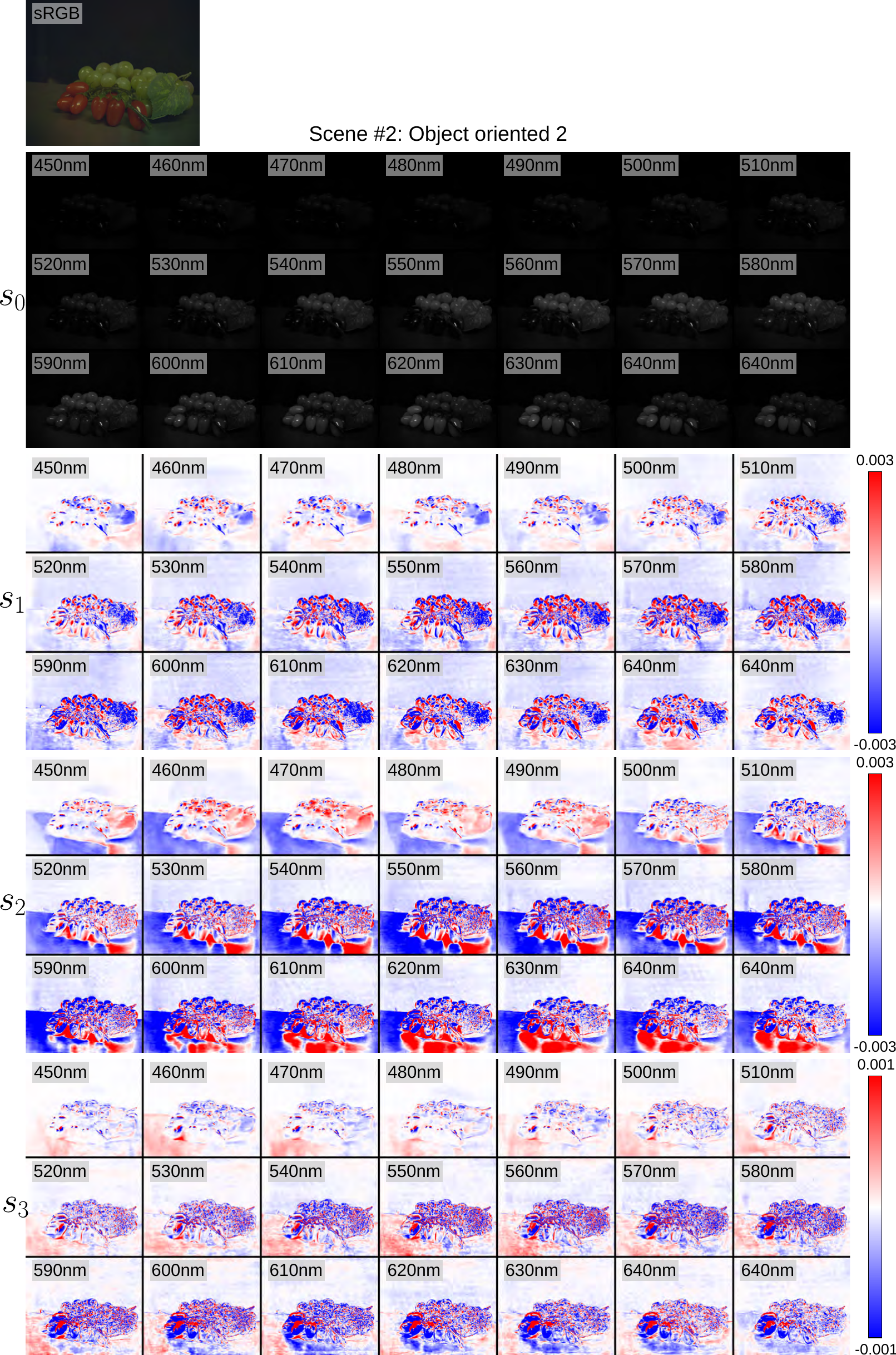}
    \caption{\textbf{Dataset example of object-centric data in the hyperspectral dataset.} }
    \label{fig:suppl_hyp_obj2}
\end{figure*}

\begin{figure*}
    \centering
    \includegraphics[height=\textheight]{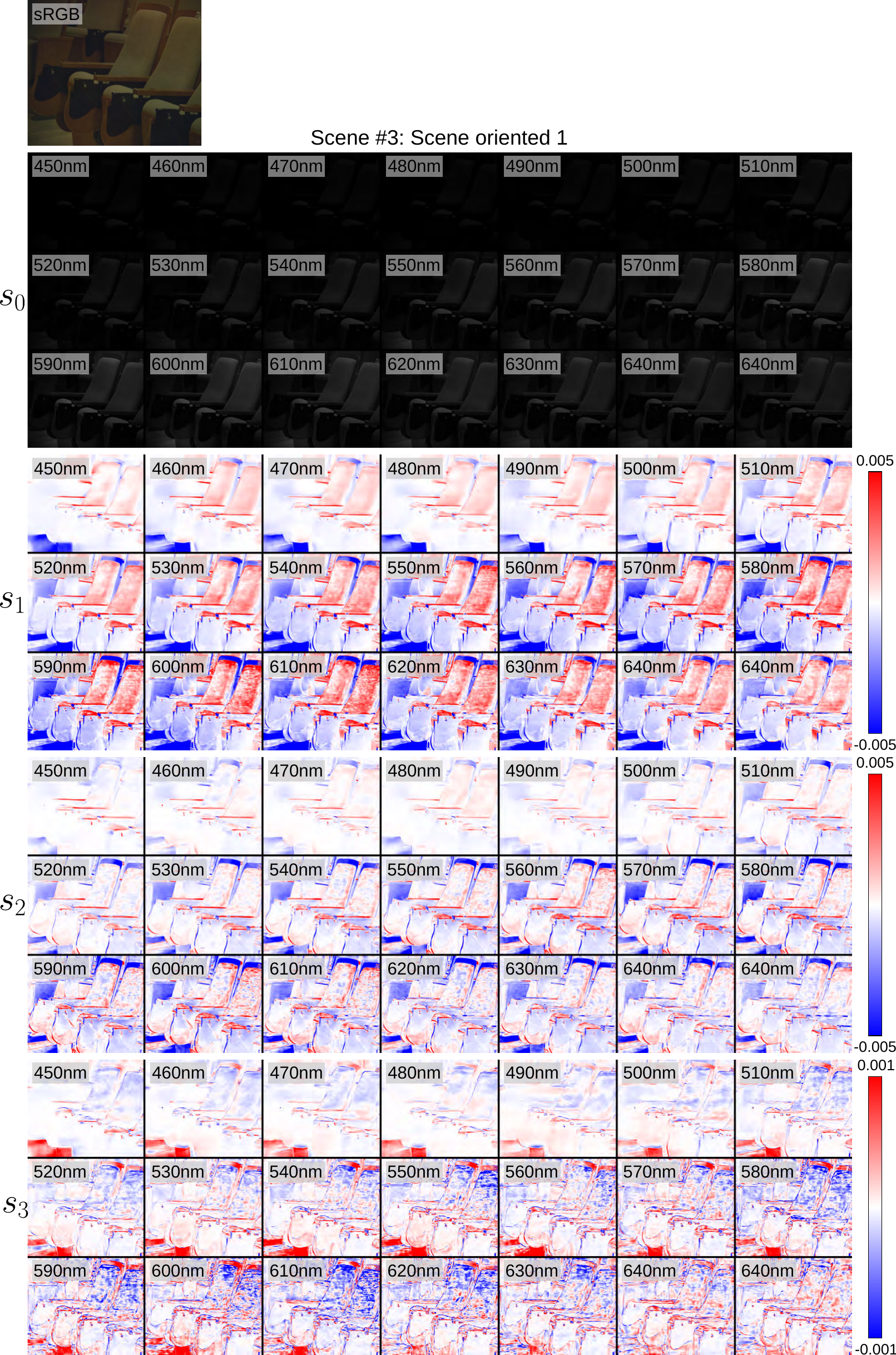}
    \caption{\textbf{Dataset example of scene-centric data in the hyperspectral dataset.} }
    \label{fig:suppl_hyp_scene1}
\end{figure*}

\begin{figure*}
    \centering
    \includegraphics[height=\textheight]{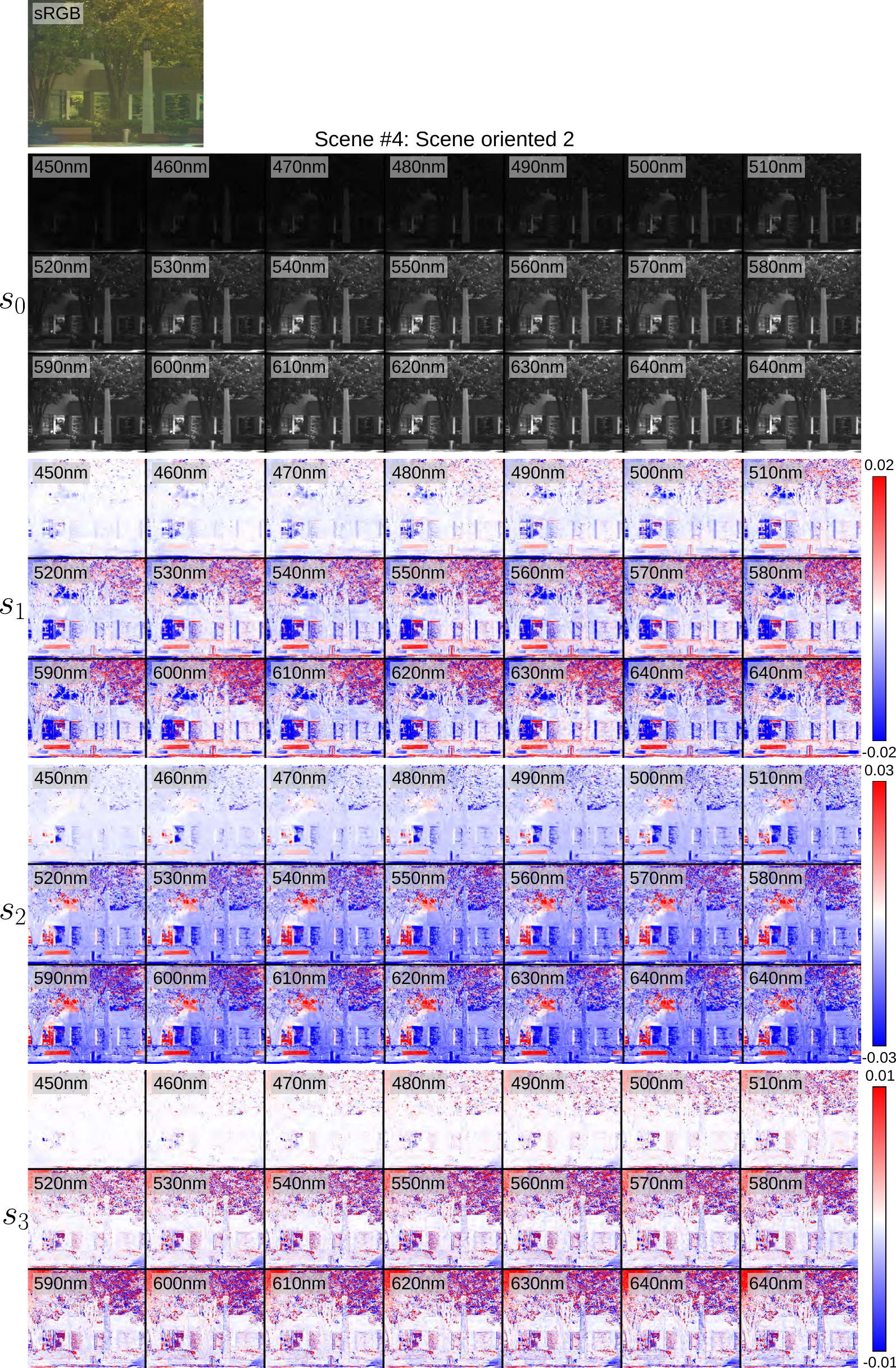}
    \caption{\textbf{Dataset example of scene-centric data in the hyperspectral dataset.} }
    \label{fig:suppl_hyp_scene2}
\end{figure*}

\begin{figure*}
    \centering
    \includegraphics[height=\textheight]{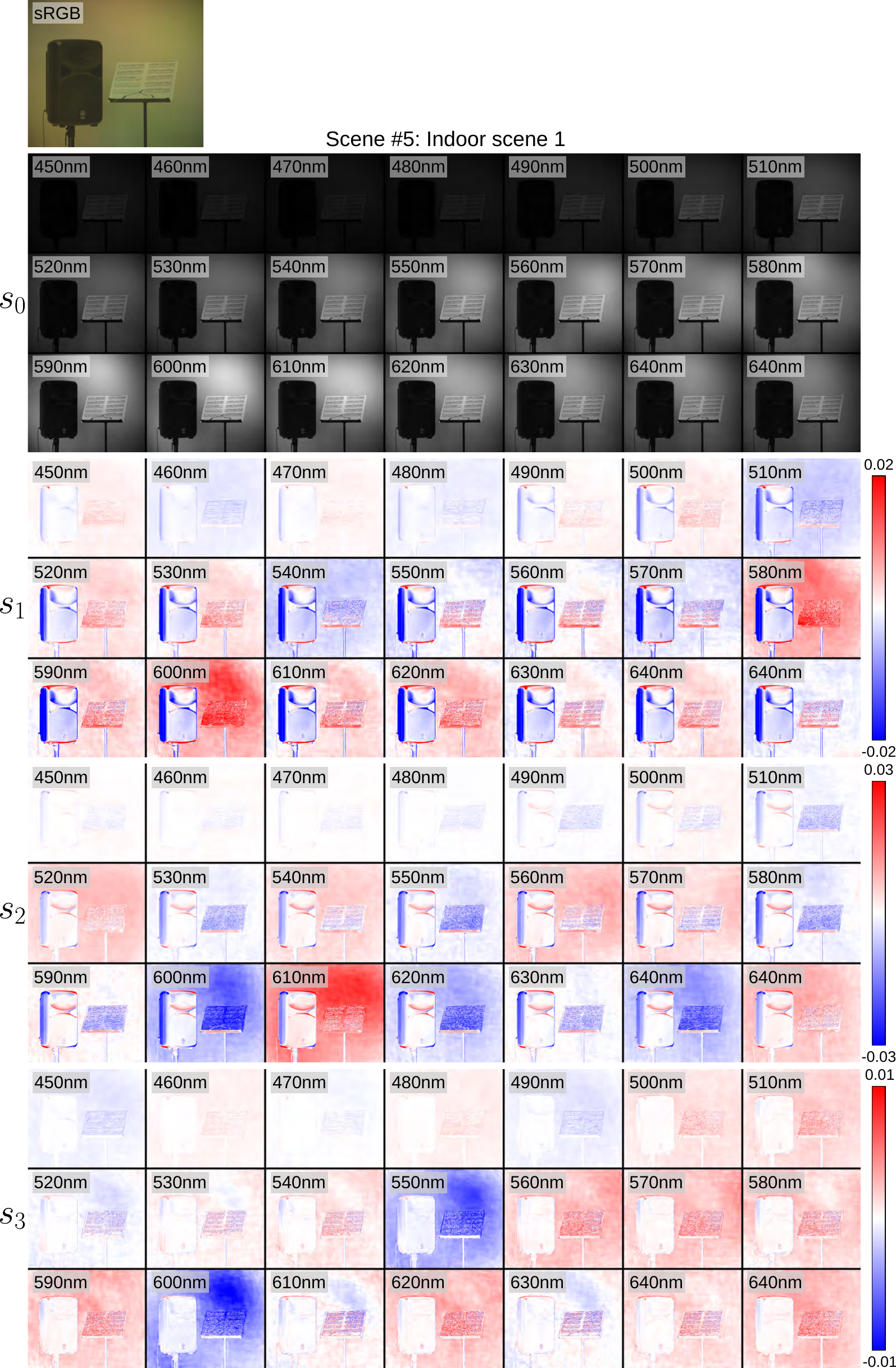}
    \caption{\textbf{Dataset example of indoor scene data in the hyperspectral dataset.} }
    \label{fig:suppl_hyp_indoor1}
\end{figure*}

\begin{figure*}
    \centering
    \includegraphics[height=\textheight]{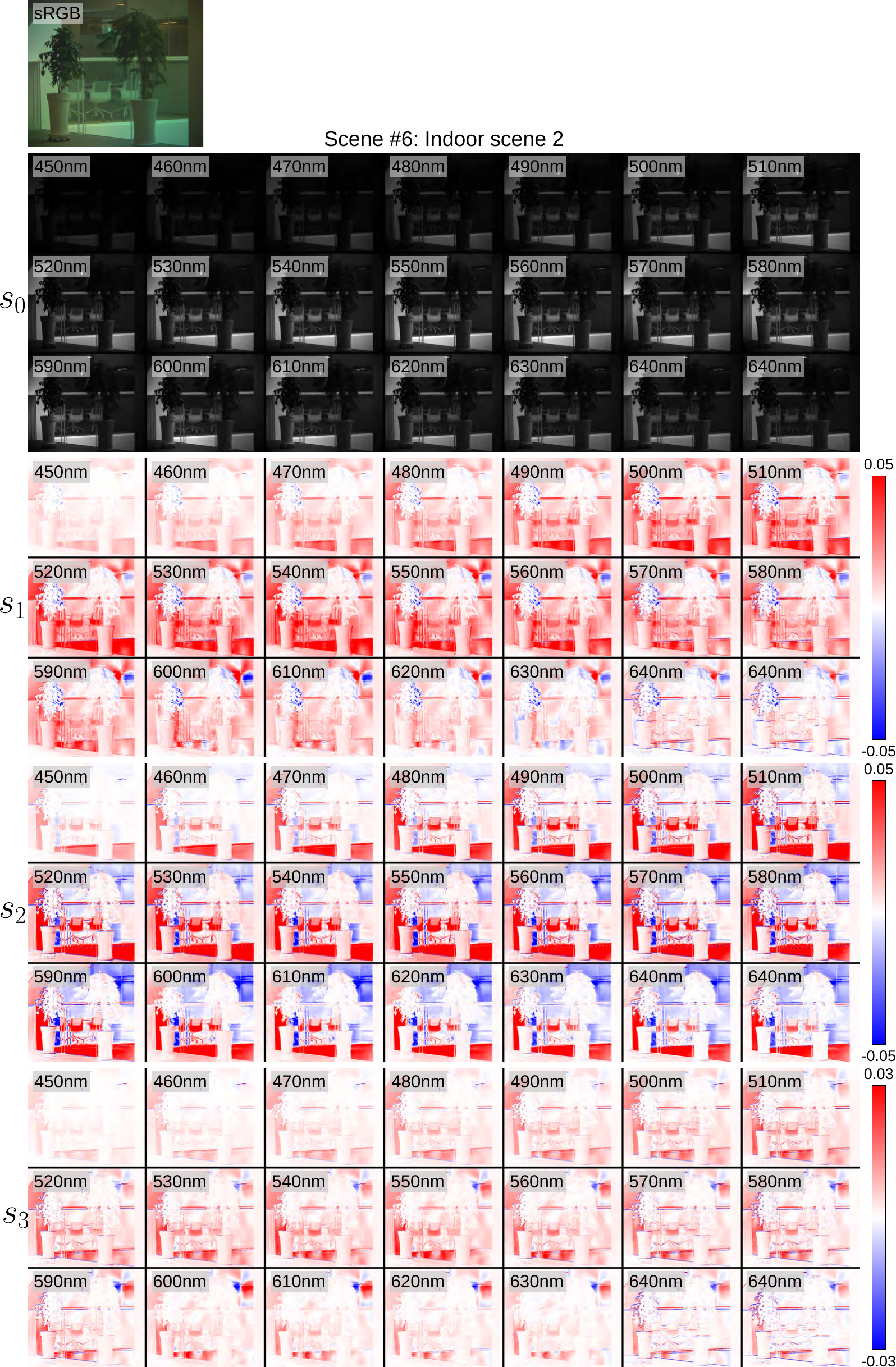}
    \caption{\textbf{Dataset example of indoor scene data in the hyperspectral dataset.} }
    \label{fig:suppl_hyp_indoor2}
\end{figure*}

\begin{figure*}
    \centering
    \includegraphics[height=\textheight]{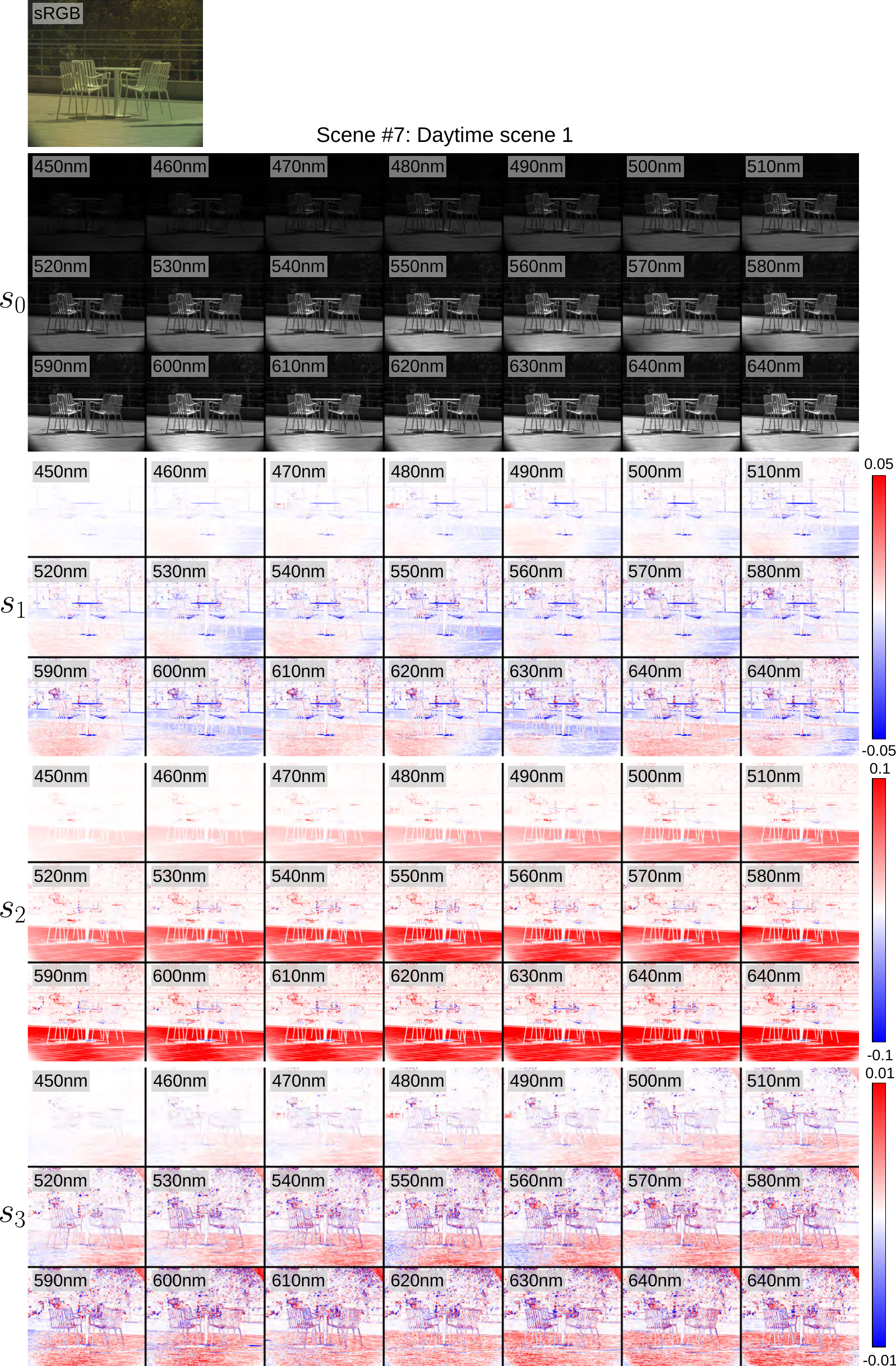}
    \caption{\textbf{Dataset example of daytime scene data in the hyperspectral dataset.} }
    \label{fig:suppl_hyp_day1}
\end{figure*}

\begin{figure*}
    \centering
    \includegraphics[height=\textheight]{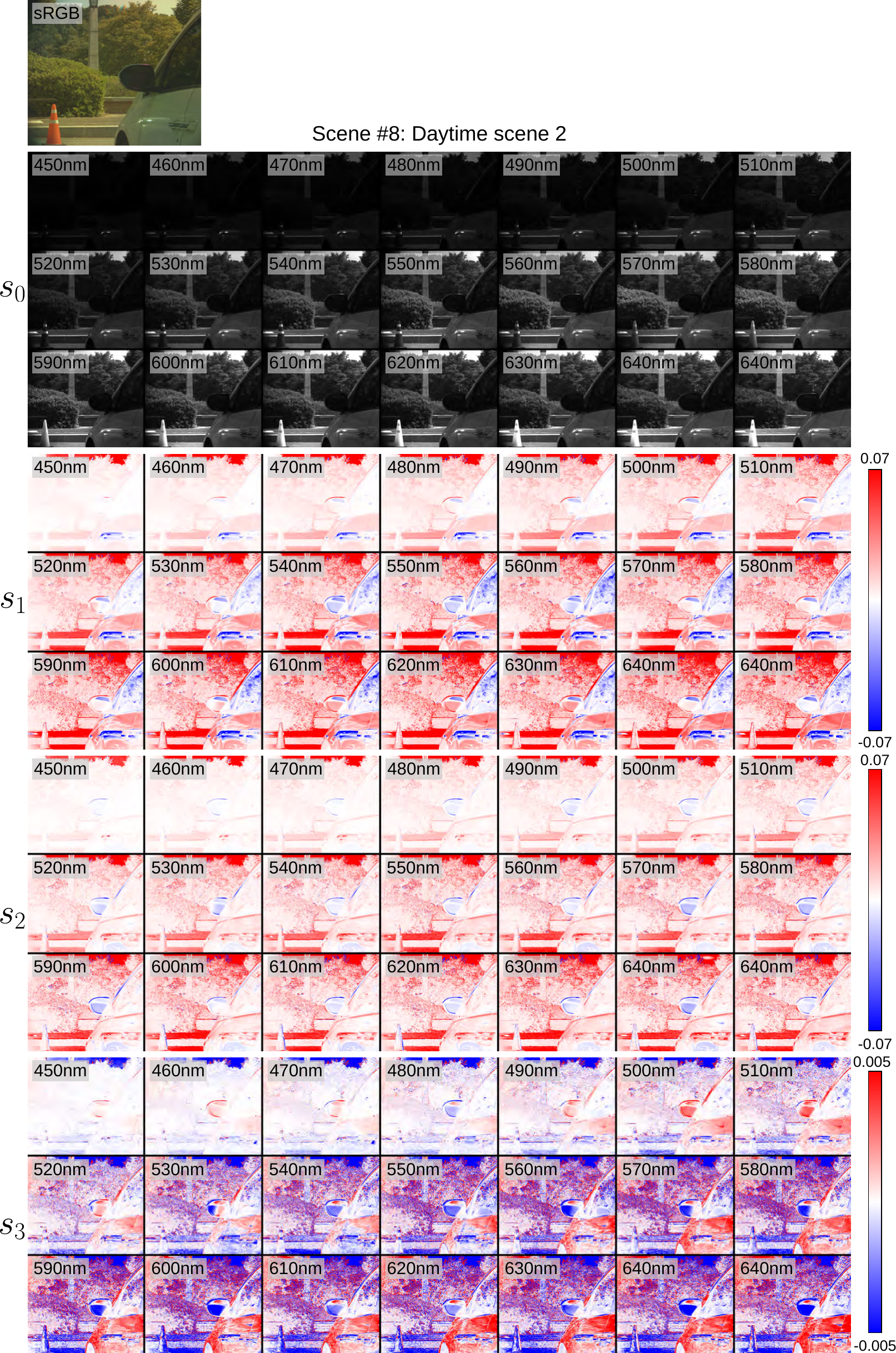}
    \caption{\textbf{Dataset example of daytime scene data in the hyperspectral dataset.} }
    \label{fig:suppl_hyp_day2}
\end{figure*}

\begin{figure*}
    \centering
    \includegraphics[height=\textheight]{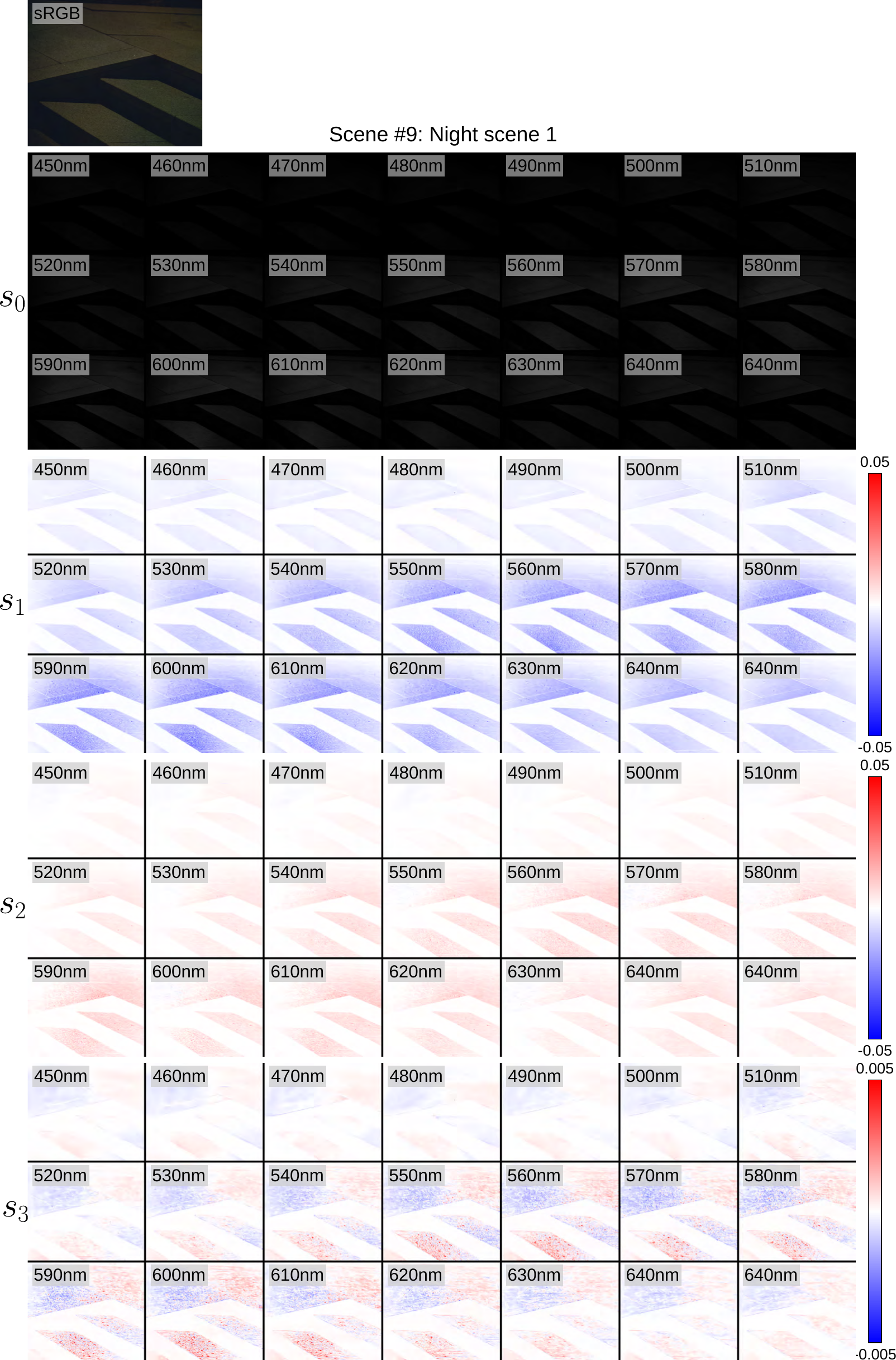}
    \caption{\textbf{Dataset example of night scene data in the hyperspectral dataset.} }
    \label{fig:suppl_hyp_night1}
\end{figure*}

\begin{figure*}
    \centering
    \includegraphics[height=\textheight]{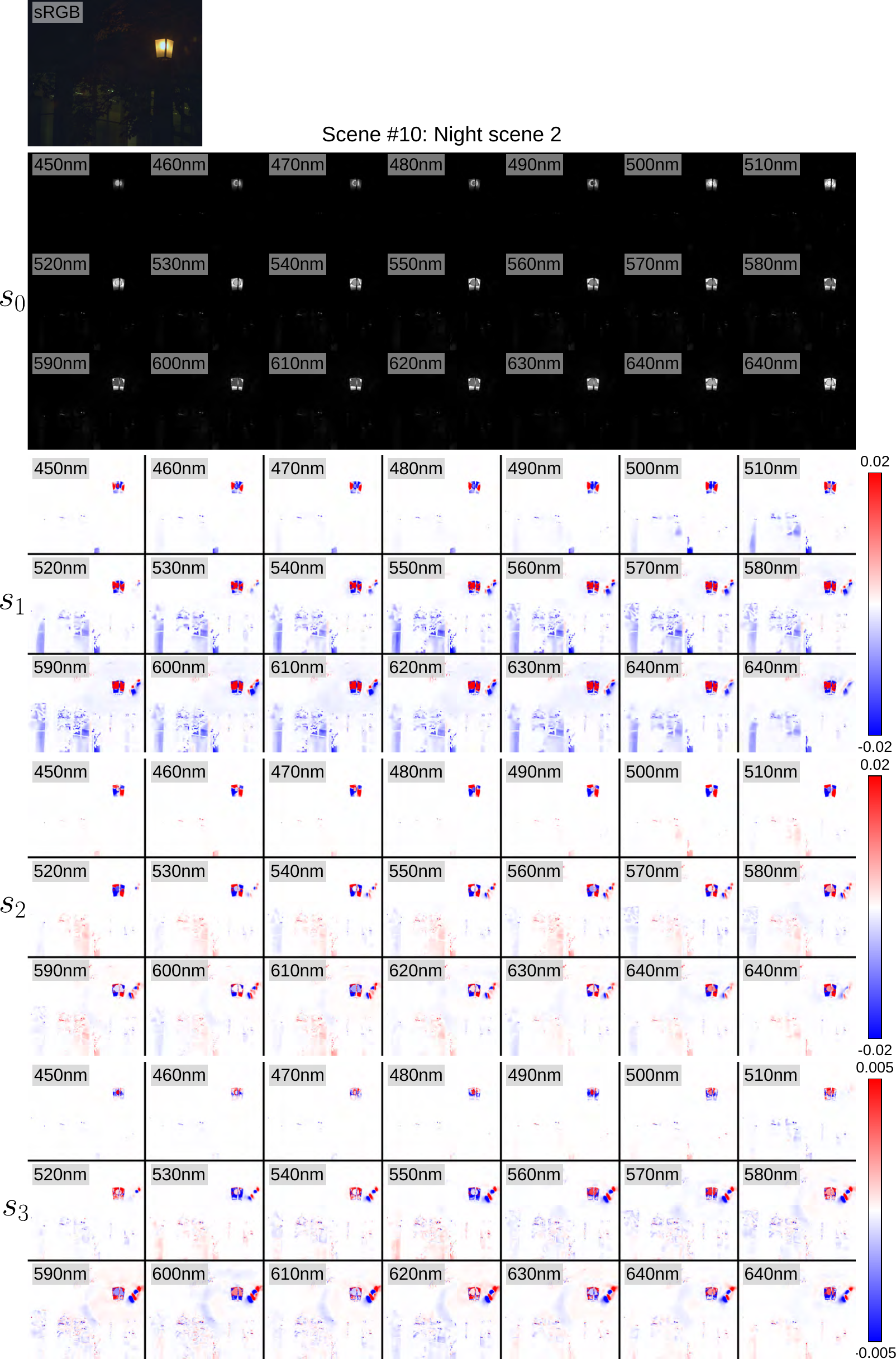}
    \caption{\textbf{Dataset example of night scene data in the hyperspectral dataset.} }
    \label{fig:suppl_hyp_night2}
\end{figure*}

\begin{figure*}
    \centering
    \includegraphics[height=\textheight]{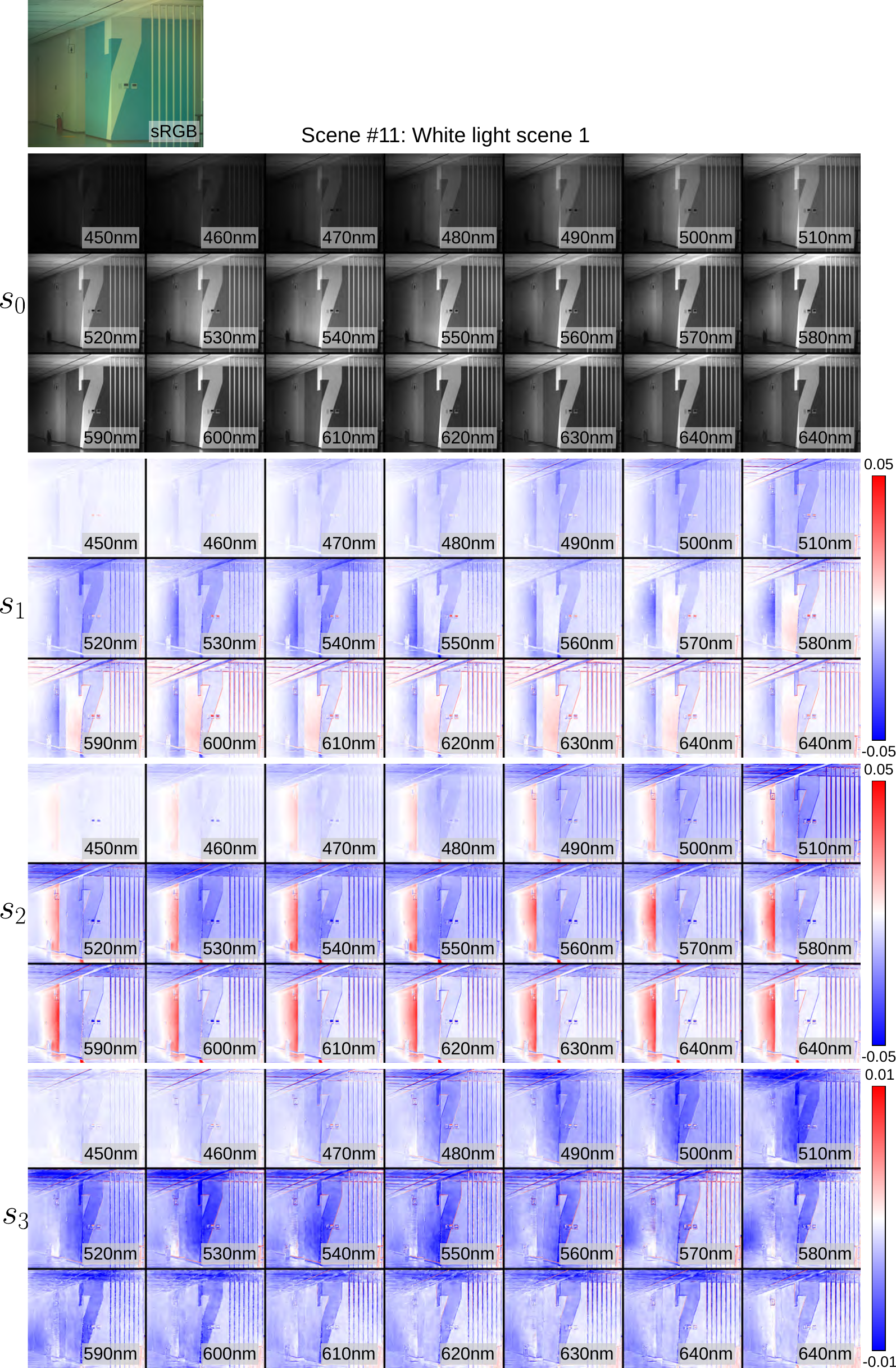}
    \caption{\textbf{Dataset example of scene data captured under white light in the hyperspectral dataset.} }
    \label{fig:suppl_hyp_white1}
\end{figure*}

\begin{figure*}
    \centering
    \includegraphics[height=\textheight]{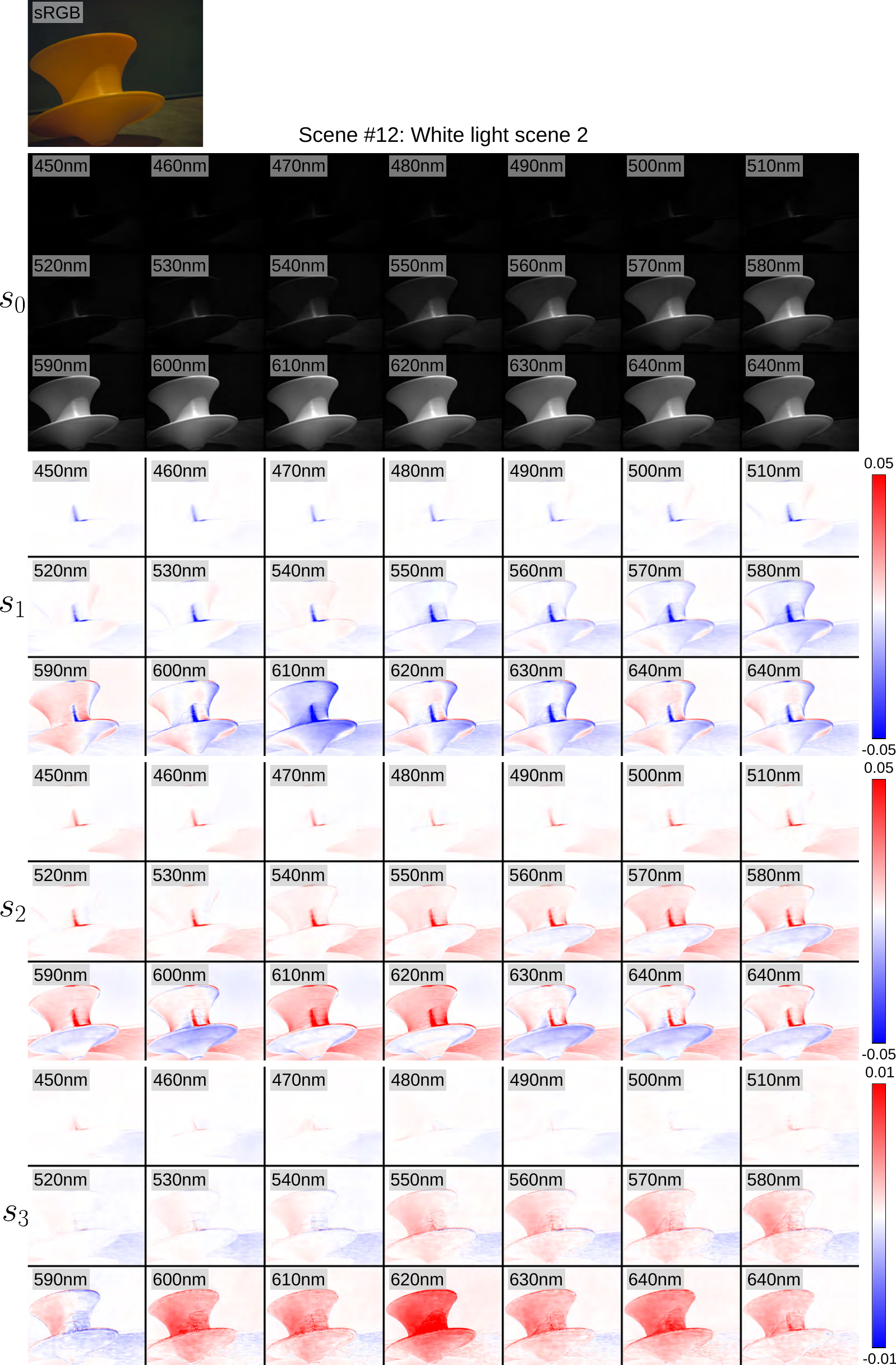}
    \caption{\textbf{Dataset example of scene data captured under white light in the hyperspectral dataset.} }
    \label{fig:suppl_hyp_white2}
\end{figure*}

\begin{figure*}
    \centering
    \includegraphics[height=\textheight]{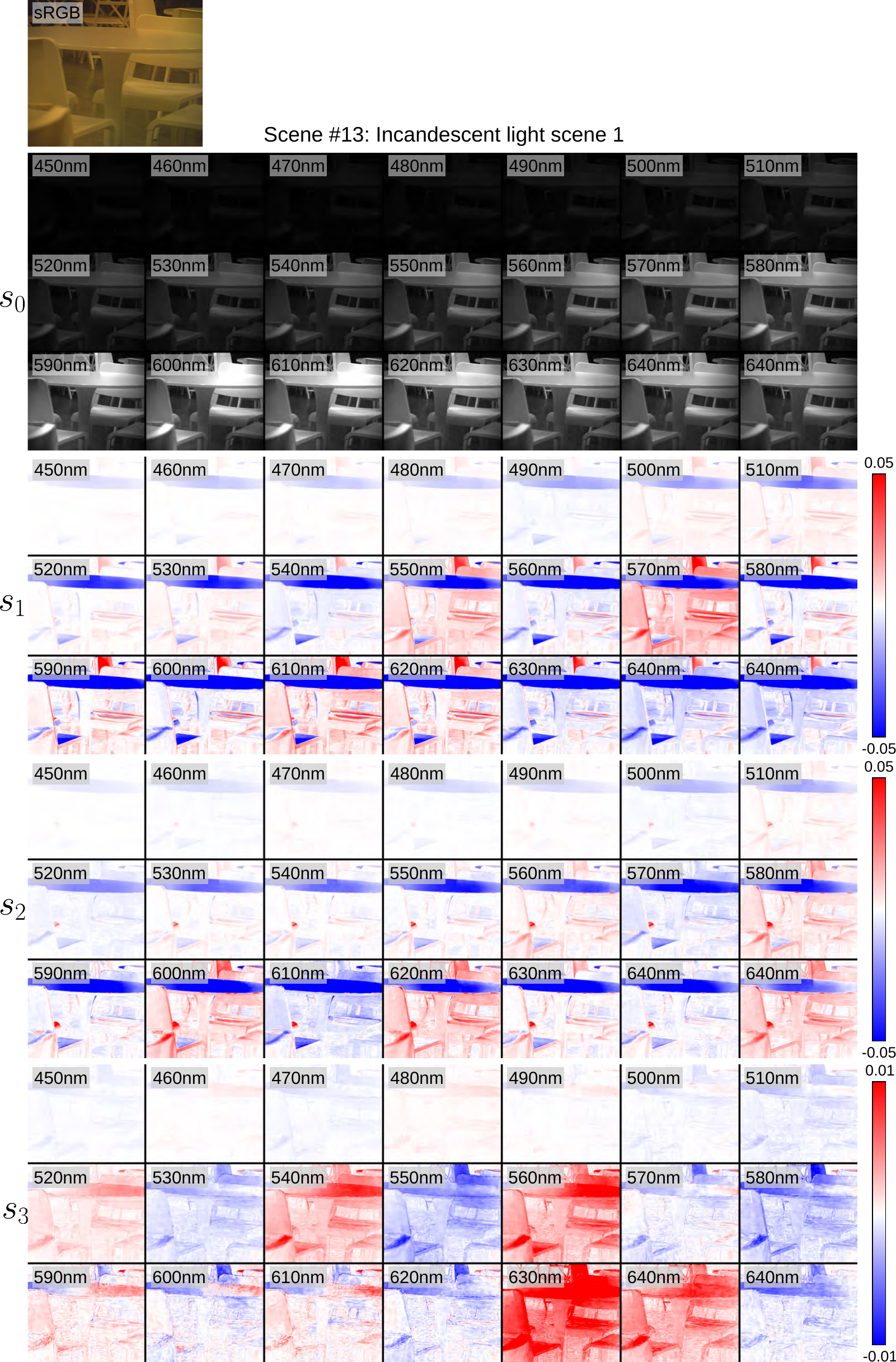}
    \caption{\textbf{Dataset example of scene data captured under incandescent light in the hyperspectral dataset.} }
    \label{fig:suppl_hyp_yellow1}
\end{figure*}

\begin{figure*}
    \centering
    \includegraphics[height=\textheight]{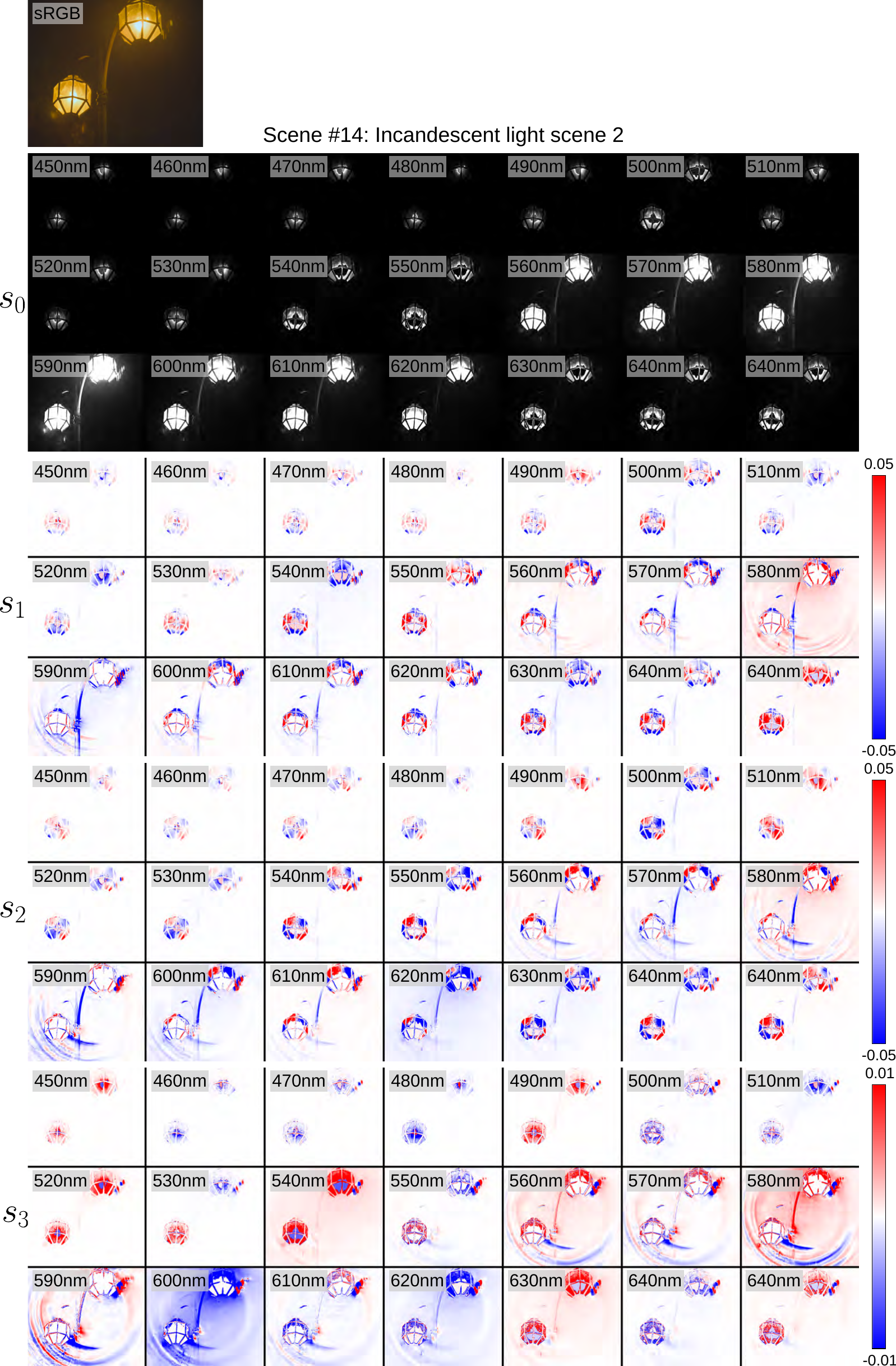}
    \caption{\textbf{Dataset example of scene data captured under incandescent light in the hyperspectral dataset.} }
    \label{fig:suppl_hyp_yellow2}
\end{figure*}

\begin{figure*}
    \centering
    \includegraphics[height=\textheight]{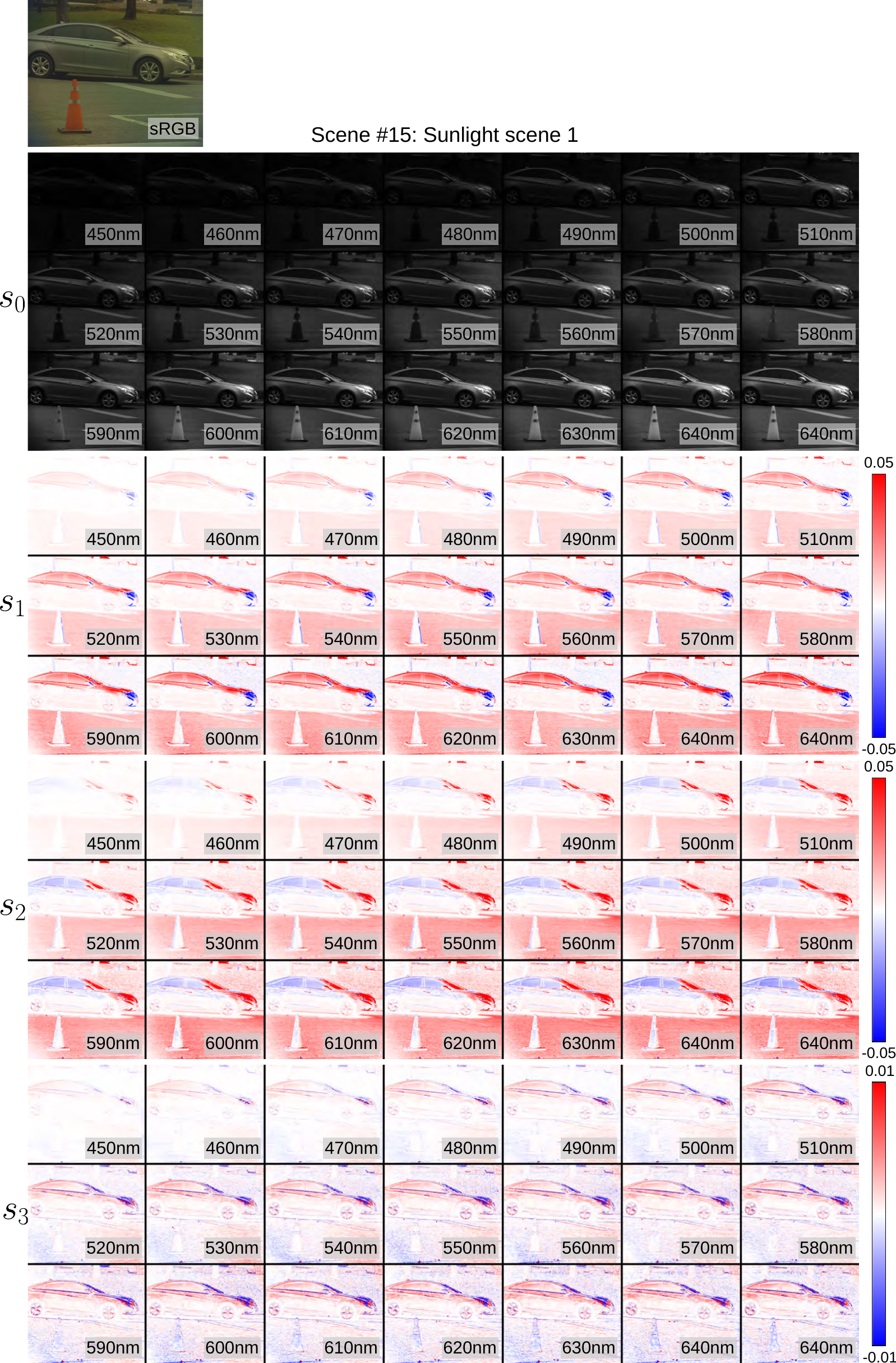}
    \caption{\textbf{Dataset example of scene data captured under sunlight in the hyperspectral dataset.} }
    \label{fig:suppl_hyp_sunny1}
\end{figure*}

\begin{figure*}
    \centering
    \includegraphics[height=\textheight]{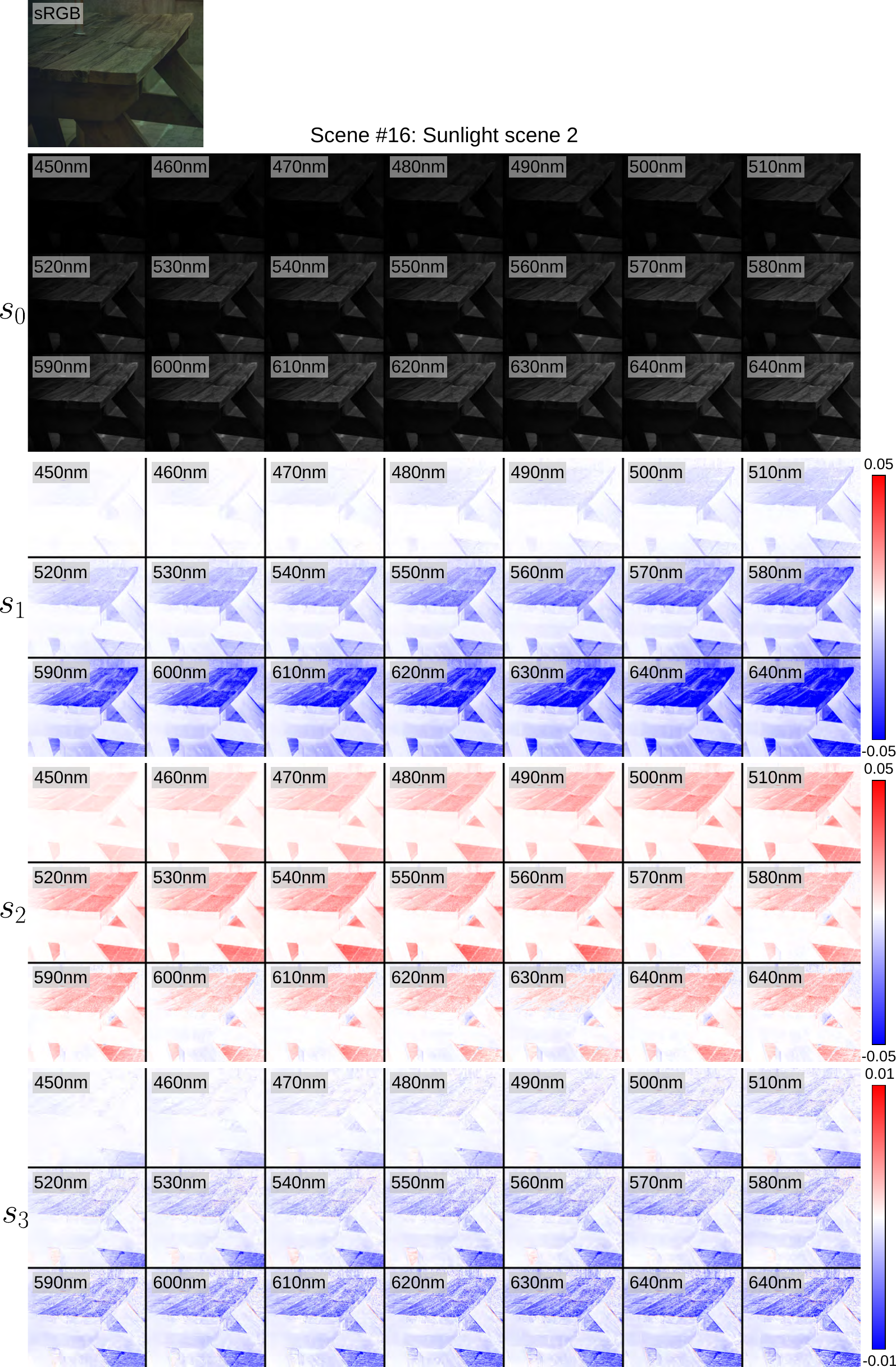}
    \caption{\textbf{Dataset example of scene data captured under sunlight in the hyperspectral dataset.} }
    \label{fig:suppl_hyp_sunny2}
\end{figure*}

\begin{figure*}
    \centering
    \includegraphics[height=\textheight]{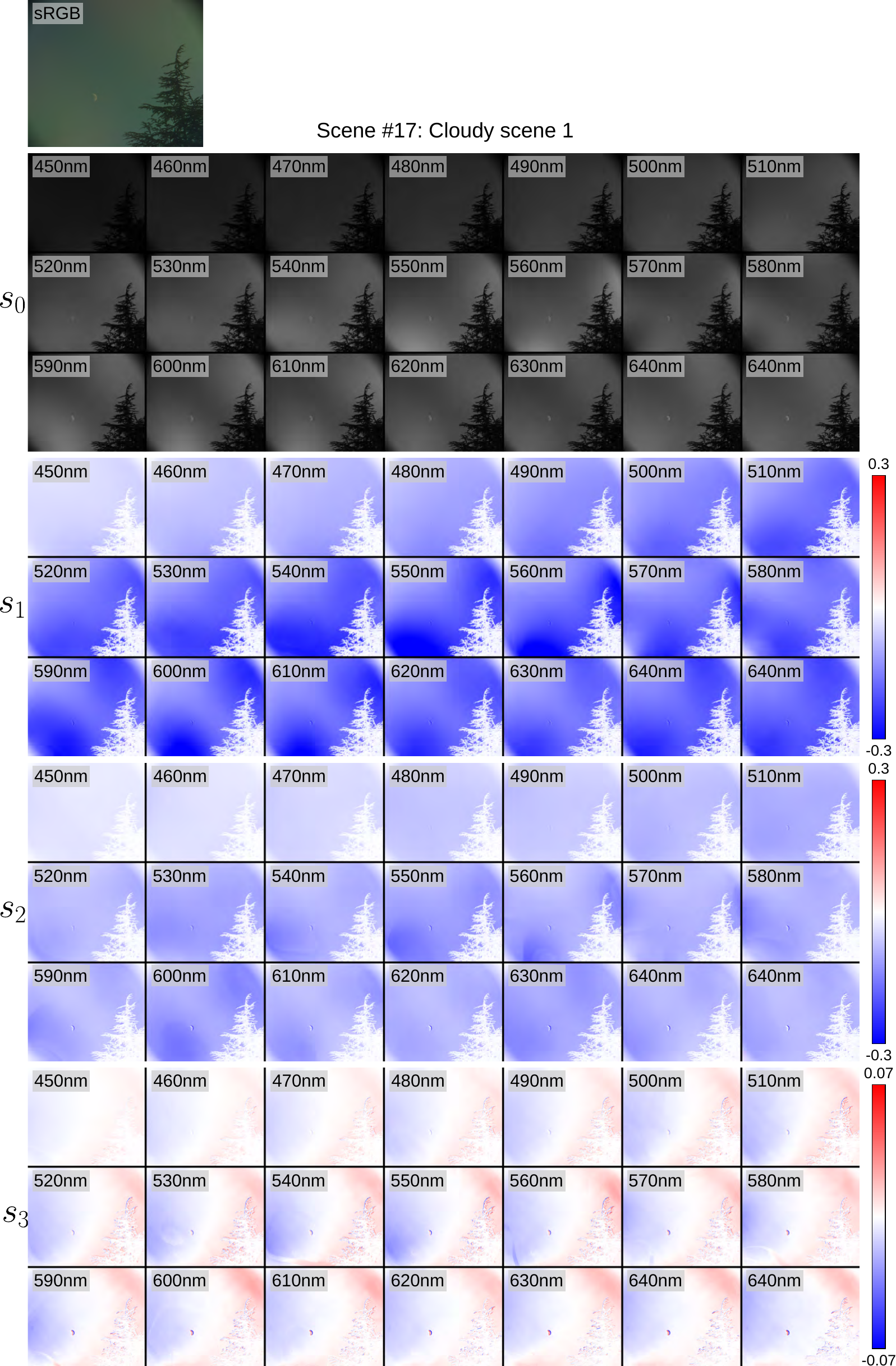}
    \caption{\textbf{Dataset example of scene data captured under cloudy condition in the hyperspectral dataset.} }
    \label{fig:suppl_hyp_cloudy1}
\end{figure*}

\begin{figure*}
    \centering
    \includegraphics[height=\textheight]{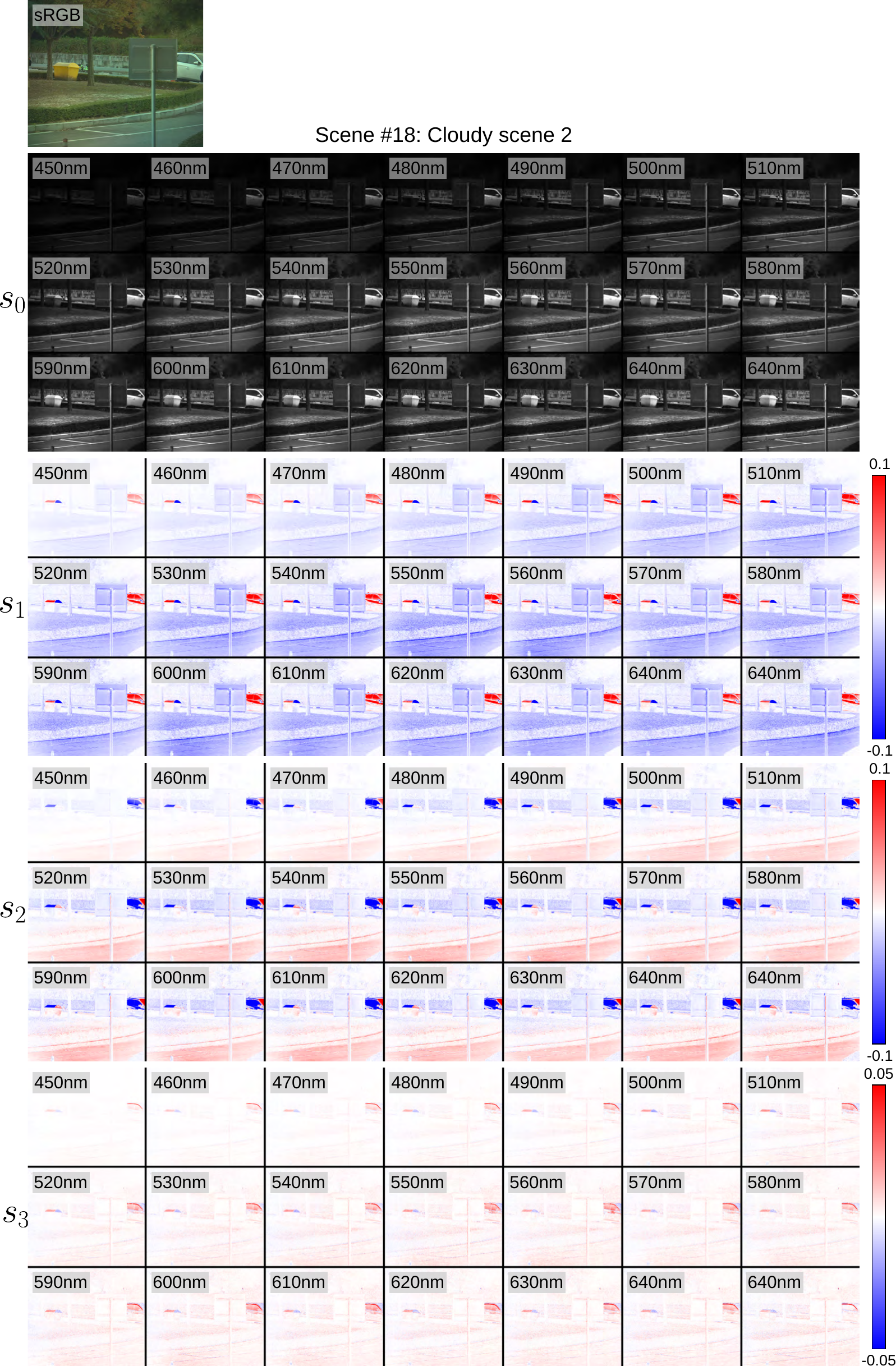}
    \caption{\textbf{Dataset example of scene data captured under cloudy condition in the hyperspectral dataset.} }
    \label{fig:suppl_hyp_cloudy2}
\end{figure*}

In this section, we present examples of our datasets in terms of sRGB, intensity and Stokes vectors ($s_1$, $s_2$ and $s_3$) across the spectrum. We collected the data based on nine different labels, encompassing location and time (indoor, outdoor daytime, outdoor night), scene types (object-oriented and scene-oriented), illumination sources (white light, incandescent light, clear sunlight and cloudy conditions). Figure~\ref{fig:suppl_hyp_obj1},~\ref{fig:suppl_hyp_obj2} shows object-oriented scenes, characterized by one or two single objects, while Figure~\ref{fig:suppl_hyp_scene1},~\ref{fig:suppl_hyp_scene2} shows scene-oriented scenes. Indoor scenes are showcased in Figure~\ref{fig:suppl_hyp_indoor1},~\ref{fig:suppl_hyp_indoor2}, outdoor daytime scenes in Figure~\ref{fig:suppl_hyp_day1},~\ref{fig:suppl_hyp_day2}, and outdoor night scenes with low intensities in Figure~\ref{fig:suppl_hyp_night1},~\ref{fig:suppl_hyp_night2}. Figure~\ref{fig:suppl_hyp_white1},~\ref{fig:suppl_hyp_white2} display data captured under white light, commonly found in indoor settings, and Figure~\ref{fig:suppl_hyp_yellow1},~\ref{fig:suppl_hyp_yellow2} shows data captured under incandescent light, known for emitting a yellowish hue. Figure~\ref{fig:suppl_hyp_sunny1},~\ref{fig:suppl_hyp_sunny2} show data captured under clear sunlight, and Figure~\ref{fig:suppl_hyp_cloudy1},~\ref{fig:suppl_hyp_cloudy2} display data captured under cloudy conditions. As shown in these figures, our datasets spans diverse scene types, objects times, and illumination conditions.


\clearpage

{\small
\bibliographystyle{template/ieee_fullname}
\bibliography{reference_supp}
}

\end{CJK}